\documentclass[10pt,twocolumn,letterpaper]{article}

\usepackage[pagenumbers]{cvpr}
\usepackage{times}
\usepackage{epsfig}
\usepackage{graphicx}
\usepackage{amsmath}
\usepackage{amssymb}

\usepackage{bbm}
\usepackage{bm}
\usepackage{xspace}
\usepackage[utf8]{inputenc} 
\usepackage{url}            
\usepackage{booktabs}       
\usepackage{amsfonts}       
\usepackage{nicefrac}       
\usepackage{microtype}      
\usepackage{color}
\usepackage{algorithm}
\usepackage{algorithmic}
\usepackage{array}
\usepackage{multirow}
\usepackage{enumitem}
\usepackage{makecell}
\usepackage{gensymb}         
\usepackage{mathtools}
\usepackage{dblfloatfix}
\usepackage{pifont}
\usepackage[permil]{overpic}

\usepackage{titletoc,tocloft}

\makeatletter
\DeclareRobustCommand\onedot{\futurelet\@let@token\@onedot}
\def\@onedot{\ifx\@let@token.\else.\null\fi\xspace}

\def\eg{\emph{e.g}\onedot} 
\def\ie{\emph{i.e}\onedot}

\makeatother

\def\mathunderline#1#2{\color{#1}\underline{{\color{black}#2}}\color{black}}
\def\colorinterp{\mathunderline{blue}{\operatorname{interp}}}
\def\colorsoftplus{\mathunderline{green}{\operatorname{softplus}}}
\def\coloralpha{\mathunderline{red}{\operatorname{alpha}}}

\usepackage{xcolor}
\usepackage{ifthen}
\usepackage{textcomp}
\definecolor{red}{rgb}{1,0,0}
\definecolor{slateblue}{rgb}{0.7,0.35,0.9}
\definecolor{green}{rgb}{0,1,0}
\definecolor{mahogany}{rgb}{0.75, 0.25, 0.0}
\definecolor{purple}{rgb}{0.6, 0, 0.6}
\definecolor{darkpurple}{rgb}{0.3, 0, 0.3}
\definecolor{darkgreen}{rgb}{0, 0.4, 0}
\definecolor{frenchblue}{rgb}{0.0, 0.45, 0.73}
\definecolor{blue}{rgb}{0,0,1}
\definecolor{goldenrod}{rgb}{0.65, 0.45, 0.03}
\definecolor{gray}{rgb}{0.5,0.5,0.5}
\definecolor{gold}{rgb}{1.0, 0.874, 0}
\definecolor{silver}{rgb}{0.67,0.67,0.67}
\definecolor{brown}{rgb}{0.8, 0.678, 0.4}

\newcommand{\gold}[1]{\colorbox{gold}{\textbf{#1}}}
\newcommand{\silver}[1]{\colorbox{silver}{\textbf{#1}}}
\newcommand{\brown}[1]{\colorbox{brown}{\textbf{#1}}}

\newboolean{revising}
\setboolean{revising}{true}
\ifthenelse{\boolean{revising}}
{
    \newcommand{\ignore}[1]{}

} {
    \newcommand{\ignore}[1]{}

}

\makeatletter
\renewcommand{\paragraph}{%
  \@startsection{paragraph}{4}%
  {\z@}{0.5\baselineskip \@plus 0ex \@minus 0ex}{-1em}%
  {\normalfont\normalsize\bfseries}%
}
\makeatother

\makeatletter
\newcommand\footnoteref[1]{\protected@xdef\@thefnmark{\ref{#1}}\@footnotemark}
\makeatother

\usepackage[pagebackref,breaklinks,colorlinks]{hyperref}

\usepackage[capitalize]{cleveref}
\crefname{section}{Sec.}{Secs.}
\Crefname{section}{Section}{Sections}
\Crefname{table}{Table}{Tables}
\crefname{table}{Tab.}{Tabs.}

\def\cvprPaperID{} 
\def\confName{}
\def\confYear{}

\begin{document}

\title{Direct Voxel Grid Optimization:\\Super-fast Convergence for Radiance Fields Reconstruction}

\author{
Cheng Sun\thanks{National Tsing Hua University}~$^{,}$\thanks{ASUS AICS Department} \\ \href{mailto:chengsun@gapp.nthu.edu.tw}{\small\tt \textcolor{black}{chengsun@gapp.nthu.edu.tw}}
\and
Min Sun\footnotemark[1]~$^{,}$\thanks{Joint Research Center for AI Technology and All Vista Healthcare} \\ \href{mailto:sunmin@ee.nthu.edu.tw}{\small\tt \textcolor{black}{sunmin@ee.nthu.edu.tw}}
\and
Hwann-Tzong Chen\footnotemark[1]~$^{,}$\thanks{Aeolus Robotics} \\ \href{mailto:htchen@cs.nthu.edu.tw}{\small\tt \textcolor{black}{htchen@cs.nthu.edu.tw}}
}

\maketitle

\begin{abstract}

We present a super-fast convergence approach to reconstructing the per-scene radiance field from a set of images that capture the scene with known poses. This task, which is often applied to novel view synthesis, is recently revolutionized by Neural Radiance Field (NeRF) for its state-of-the-art quality and flexibility.
However, NeRF and its variants require a lengthy training time ranging from hours to days for a single scene.
In contrast, our approach achieves NeRF-comparable quality and converges rapidly from scratch in less than 15 minutes with a single GPU.
We adopt a representation consisting of a density voxel grid for scene geometry and a feature voxel grid with a shallow network for complex view-dependent appearance.
Modeling with explicit and discretized volume representations is not new, but we propose two simple yet non-trivial techniques that contribute to fast convergence speed and high-quality output.
First, we introduce the post-activation interpolation on voxel density, which is capable of producing sharp surfaces in lower grid resolution.
Second, direct voxel density optimization is prone to suboptimal geometry solutions, so we robustify the optimization process by imposing several priors.
Finally, evaluation on five inward-facing benchmarks shows that our method matches, if not surpasses, NeRF's quality, yet it only takes about 15 minutes to train from scratch for a new scene.
Code: \url{ https://github.com/sunset1995/DirectVoxGO}.
\end{abstract}

\begin{figure}
    \centering
    \bgroup
    \def\arraystretch{0.5}
    \begin{subfigure}[t]{\linewidth}
        \centering
        \begin{tabular}{@{}ccc@{}}
            \includegraphics[width=0.28\linewidth]{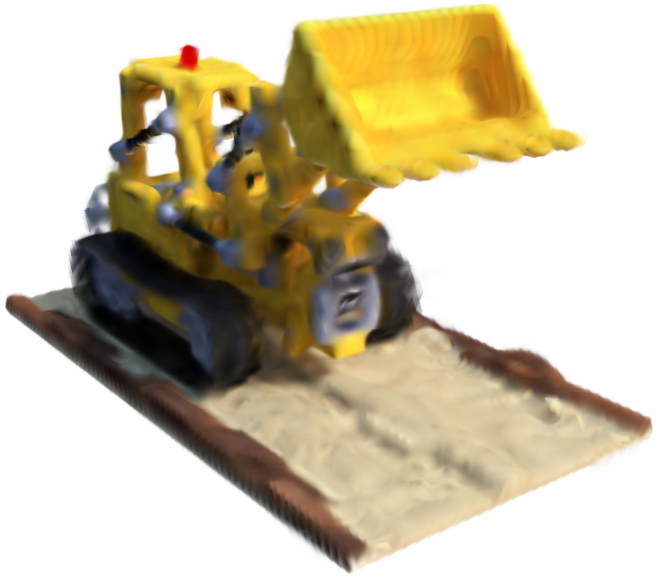} &
            \includegraphics[width=0.28\linewidth]{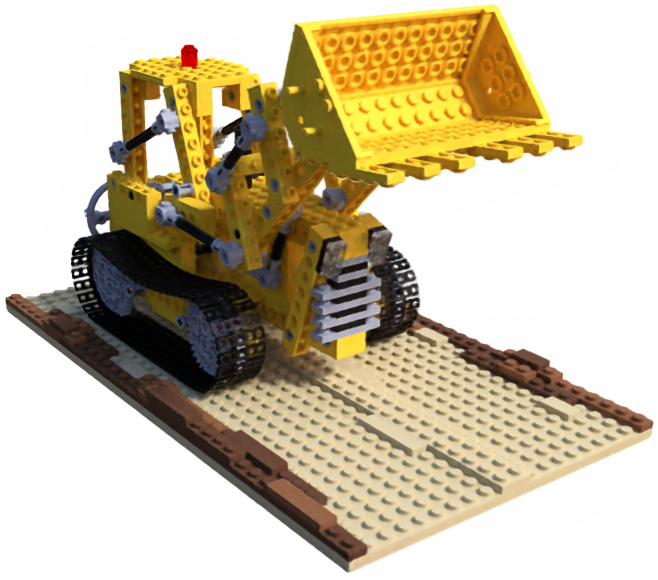} & 
            \includegraphics[width=0.28\linewidth]{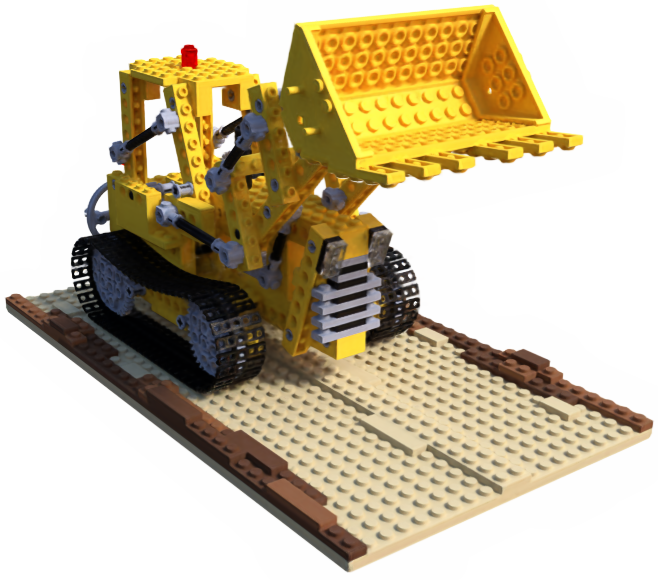} \\
            {\scriptsize 23.64 PSNR.} &
            {\scriptsize 32.72 PSNR.} &
            {\scriptsize 34.22 PSNR.} \\
            {\scriptsize Ours at 2.33 mins.} &
            {\scriptsize Ours at 5.07 mins.} &
            {\scriptsize Ours at 13.72 mins.}
        \end{tabular}
        \caption{
        The synthesized novel view by our method at three training checkpoints.
        }
    \end{subfigure}
    \egroup
    \par\medskip
    \begin{subfigure}[t]{\linewidth}
        \centering
        \includegraphics[width=\linewidth]{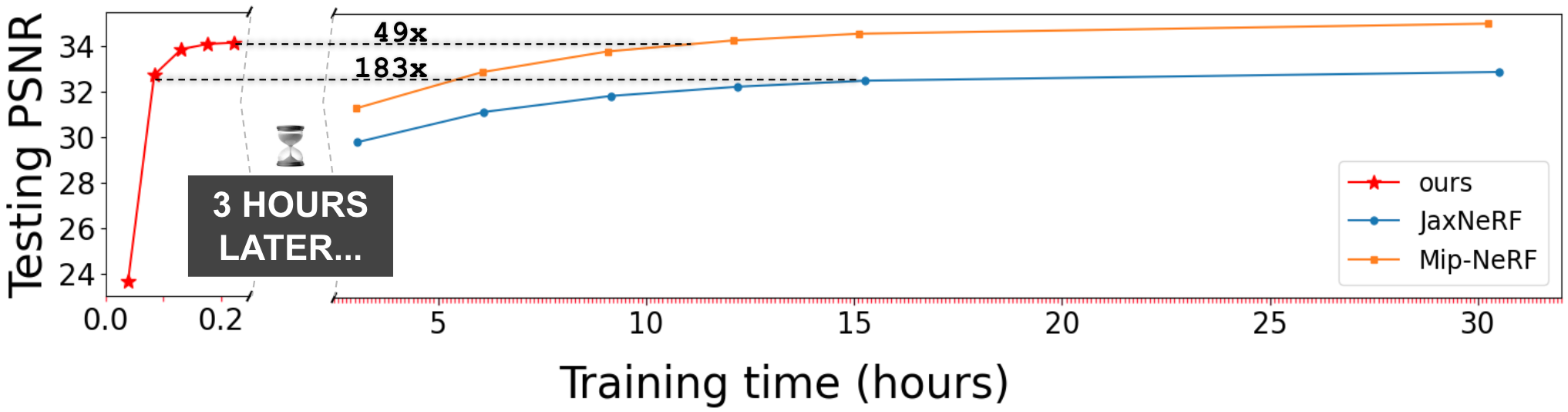}
        \caption{
        The training curves of different methods on {\it Lego} scene.
        The training time of each method is measured on our machine with a single NVIDIA RTX 2080 Ti GPU.
        }
    \end{subfigure}
    \caption{
    {\bf Super-fast convergence by our method.}
    The key to our speedup is to optimize the volume density modeled in a dense voxel grid directly.
    Note that our method needs neither a conversion step from any trained implicit model (\eg, NeRF) nor a cross-scene pretraining, \ie, our voxel grid representation is directly and efficiently trained from scratch for each scene.
    }
    \label{fig:exdv_teaser}
    \vspace{-1em}
\end{figure}

\section{Introduction}
Achieving free-viewpoint navigation of 3D objects or scenes from only a set of calibrated images as input is a demanding task.
For instance, it enables online product showcase to provide an immersive user experience comparing to static image demonstration.
Recently, Neural Radiance Fields (NeRFs)~\cite{MildenhallSTBRN20} have emerged as powerful representations yielding state-of-the-art quality on this task.

Despite its effectiveness in representing scenes, NeRF is known to be hampered by the need of lengthy training time and the inefficiency in rendering new views. This makes NeRF infeasible for many application scenarios.
Several follow-up methods~\cite{YuLTLNK2021,GarbinKJSV2021,HedmanSMBD2021,ReiserPLG2021,LiuGLCT20,RebainJYLYT21,LindellMW2021} have shown significant speedup of FPS in testing phase, some of which even achieve real-time rendering.
However, only few methods show training times speedup, and the improvements are not comparable to ours~\cite{BarronMTHMS21,DengLZR21,LiuPLWWTZW21} or lead to worse quality~\cite{ChenXZZXYS21,WangWGSZBMSF21}.
On a single GPU machine, several hours of per scene optimization or a day of pretraining is typically required.

To reconstruct a volumetric scene representation from a set of images, NeRF uses multilayer perceptron (MLP) to implicitly learn the mapping from a queried 3D point (with a viewing direction) to its colors and densities.
The queried properties along a camera ray can then be accumulated into a pixel color by volume rendering techniques.
Our work takes inspiration from the recent success~\cite{YuLTLNK2021,HedmanSMBD2021,GarbinKJSV2021} that uses classic voxel grid to explicitly store the scene properties, which enables real-time rendering and shows good quality.
However, their methods can not train from scratch and need a conversion step from the trained implicit model, which causes a bottleneck to the training time.

The key to our speedup is to use a dense voxel grid to directly model the 3D geometry (volume density).
Developing an elaborate strategy for view-dependent colors is not in the main scope of this paper, and we simply use a hybrid representation (feature grid with shallow MLP) for colors.

Directly optimizing the density voxel grid leads to super-fast converges but is prone to suboptimal solutions, where our method allocates ``cloud" at free space and tries to fit the photometric loss with the cloud instead of searching a geometry with better multi-view consistency.
Our solution to this problem is simple and effective.
First, we initialize the density voxel grid to yield opacities very close to zero everywhere to avoid the geometry solutions being biased toward the cameras' near planes.
Second, we give a lower learning rate to voxels visible to fewer views, which can avoid redundant voxels that are allocated just for explaining the observations from a small number of views.
We show that the proposed solutions can successfully avoid the suboptimal geometry and work well on the five datasets.

Using the voxel grid to model volume density still faces a challenge in scalability.
For parsimony, our approach automatically finds a BBox tightly encloses the volume of interest to allocate the voxel grids.
Besides, we propose post-activation---applying all the activation functions after trilinearly interpolating the density voxel grid.
Previous work either interpolates the voxel grid for the activated opacity or uses nearest-neighbor interpolation, which results in a smooth surface in each grid cell.
Conversely, we prove mathematically and empirically that the proposed post-activation can model (beyond) a sharp linear surface within a single grid cell.
As a result, we can use fewer voxels to achieve better qualities---our method with $160^3$ dense voxels already outperforms NeRF in most cases.

In summary, we have two main technical contributions.
First, we implement two priors to avoid suboptimal geometry in direct voxel density optimization.
Second, we propose the {\em post-activated voxel-grid interpolation}, which enables sharp boundary modeling in lower grid resolution.
The resulting key merits of this work are highlighted as follows:
\begin{itemize}
    \setlength\itemsep{0em}
    \item Our convergence speed is about two orders of magnitude faster than NeRF---reducing training time from $10\mathrm{-}20$ hours to $15$ minutes on our machine with a single NVIDIA RTX 2080 Ti GPU, as shown in Fig.~\ref{fig:exdv_teaser}.
    \item We achieve visual quality comparable to NeRF at a rendering speed that is about $45\times$ faster.
    \item Our method does not need cross-scene pretraining.
    \item Our grid resolution is about $160^3$, while the grid resolution in previous work~\cite{GarbinKJSV2021,YuLTLNK2021,HedmanSMBD2021} ranges from $512^3$ to $1300^3$ to achieve NeRF-comparable quality.
\end{itemize}

\section{Related work}

\paragraph{Representations for novel view synthesis.}
Images synthesis from novel viewpoints given a set of images capturing the scene is a long-standing task with rich studies.
Previous work has presented several scene representations reconstructed from the input images to synthesize the unobserved viewpoints.
Lumigraph~\cite{GortlerGSC96,BuehlerBMGC01} and light field representation~\cite{LevoyH96,LevinD10,DavisLD12,ShiHDKD14} directly synthesize novel views by interpolating the input images but require very dense scene capture.
Layered depth images~\cite{ShadeGHS98,TulsianiTS18,DhamoTLNT19,ShihSKH20} work for sparse input views but rely on depth maps or estimated depth with sacrificed quality.
Mesh-based representations~\cite{DebevecTM96,WoodAACDSS00,WaechterMG14,ThiesZN19} can run in real-time but have a hard time with gradient-based optimization without template meshes provided.
Recent approaches employ 2D/3D Convolutional Neural Network (CNNs) to estimate multiplane images (MPIs)~\cite{ZhouTFFS18,FlynnBDDFOST19,MildenhallSCKRN19,SrinivasanTBRNS19,LiXDS20,TuckerS20} for forward-facing captures; estimate voxel grid~\cite{LombardiSSSLS19,SitzmannTHNWZ19,HeCJS20} for inward-facing captures.
Our method uses gradient-descent to optimize voxel grids directly and does not rely on neural networks to predict the grid values, and we still outperform the previous works~\cite{LombardiSSSLS19,SitzmannTHNWZ19,HeCJS20} with CNNs by a large margin.

\paragraph{Neural radiance fields.}
Recently, NeRF~\cite{MildenhallSTBRN20} stands out to be a prevalent method for novel view synthesis with rapid progress, which takes a moderate number of input images with known camera poses.
Unlike traditional explicit and discretized volumetric representations (\eg, voxel grids and MPIs), NeRF uses coordinate-based multilayer perceptrons (MLP) as an implicit and continuous volumetric representation.
NeRF achieves appealing quality and has good flexibility with many follow-up extensions to various setups, \eg, relighting~\cite{BiXSMSHHKR20,SrinivasanDZTMB21,ZhangSDDFB21,BossBJBLL21}, deformation~\cite{ParkSBBGSM21,TretschkTGZLT21,GafniTZN21,NoguchiSLH21,ParkSHBBGMS21}, self-calibration~\cite{LinFBRIL21,WangWXCP21,MengCLWSXHY21,JeongACACP21,LinCTL21}, meta-learning~\cite{TancikMWSSBN21}, dynamic scene modeling~\cite{Martin-BruallaR21,PumarolaCPM21,LiNSW21,XianHK021,GaoSKH21}, and generative modeling~\cite{SchwarzLN020,ChanMK0W21,KosiorekSZMSMR21}.
Nevertheless, NeRF has unfavorable limitations of lengthy training progress and slow rendering speed.
In this work, we mainly follow NeRF's original setup, while our method can optimize the volume density explicitly encoded in a voxel grid to speed up both training and testing by a large margin with comparable quality.

\paragraph{Hybrid volumetric representations.}
To combine NeRF's implicit representation and traditional grid representations, the coordinate-based MLP is extended to also conditioning on the local feature in the grid.
Recently, hybrid voxels~\cite{HedmanSMBD2021,LiuGLCT20} and MPIs~\cite{WizadwongsaPYS21} representations have shown success in fast rendering speed and result quality.
We use hybrid representation to model view-dependent color as well.

\begin{figure*}
    \centering
    \begin{overpic}[width=.9\linewidth, trim=0 263 497 327, clip]{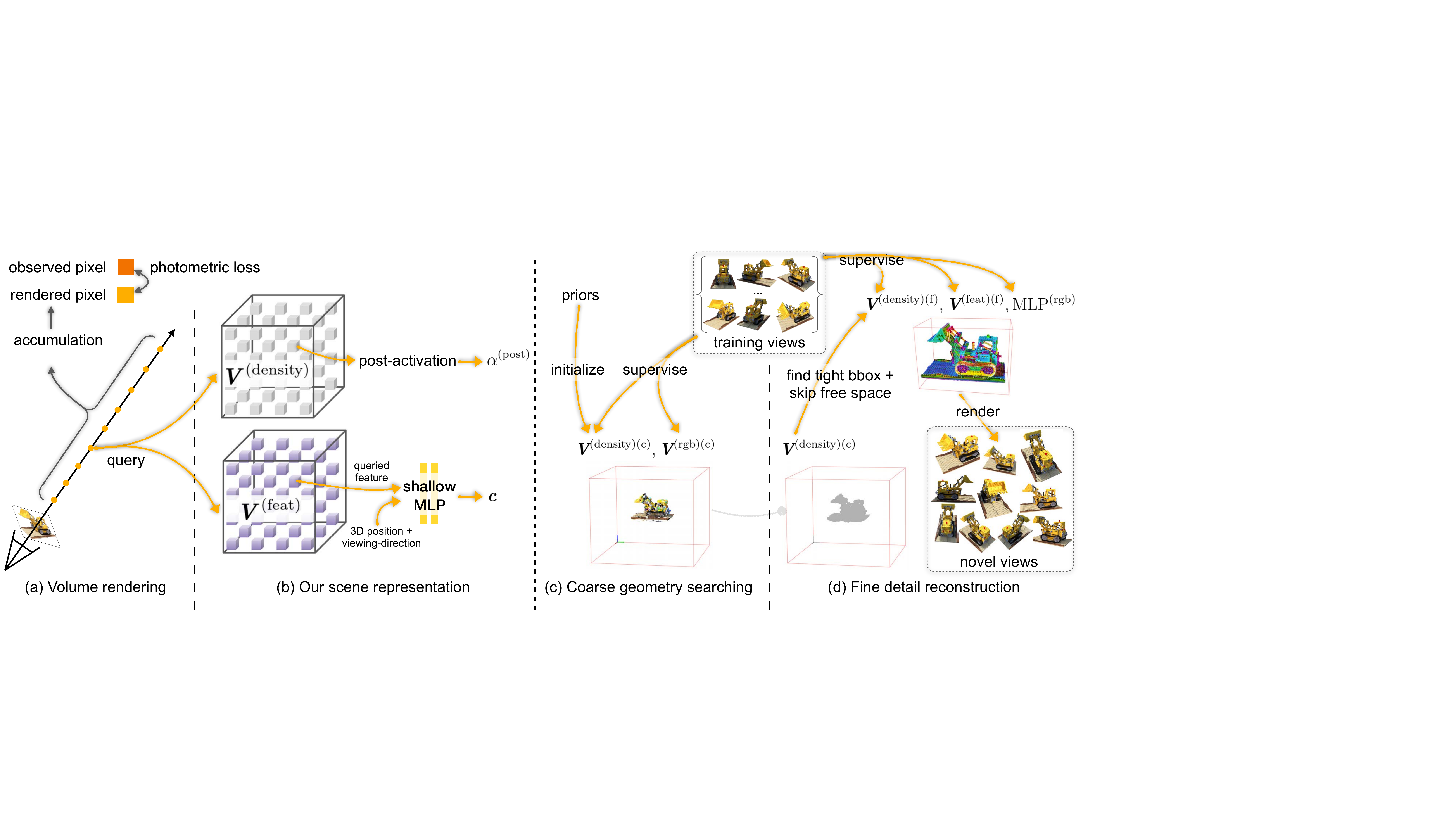}
    \put(60,5){\scriptsize (\cref{sec:pre_nerf})}
    \put(315,5){\scriptsize (\cref{ssec:fine_learn})}
    \put(575,5){\scriptsize (\cref{ssec:coarse_learn})}
    \put(830,5){\scriptsize (\cref{ssec:fine_learn})}
    \put(24,237){\scriptsize (\cref{eq:volume_rendering})}
    \put(160,305){\scriptsize (\cref{eq:photo_loss})}
    \put(355,218){\scriptsize (\cref{sec:post_act_vox})}
    \end{overpic}
    \caption{
        {\bf Approach overview.}
        We first review NeRF in \cref{sec:pre_nerf}.
        In \cref{sec:post_act_vox}, we present a novel post-activated density voxel grid to support sharp surface modeling in lower grid resolutions.
        In \cref{sec:approach}, we show our approach to the reconstruction of radiance field with super-fast convergence, where we first find a coarse geometry in \cref{ssec:coarse_learn} and then reconstruct the fine details and view-dependent effects in \cref{ssec:fine_learn}.
    }
    \label{fig:overview}
    \vspace{-1em}
\end{figure*}
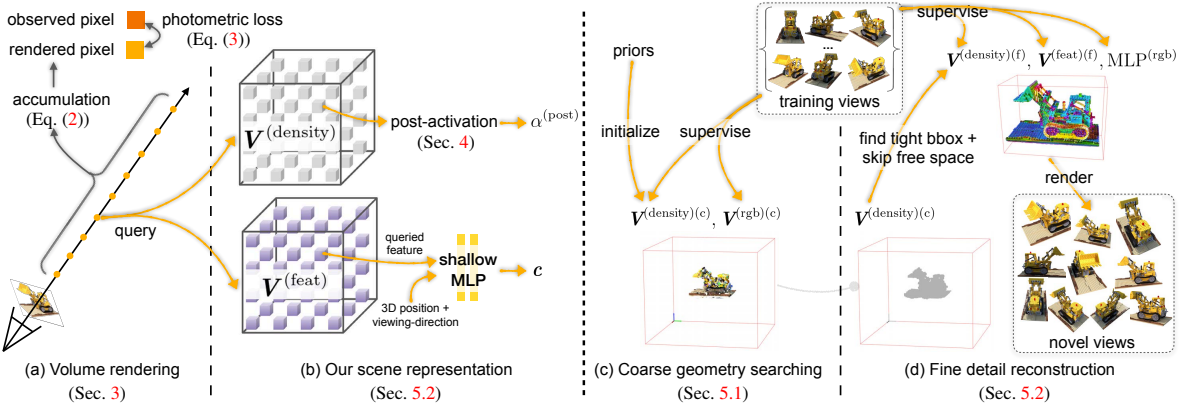

\paragraph{Fast NeRF rendering.}
NSVF~\cite{LiuGLCT20} uses octree in its hybrid representation to avoid redundant MLP queries in free space.
However, NSVF still needs many training hours due to the deep MLP in its representation.
Recent methods further use thousands of tiny MLPs~\cite{ReiserPLG2021} or explicit volumetric representations~\cite{WizadwongsaPYS21,YuLTLNK2021,GarbinKJSV2021,HedmanSMBD2021} to achieve real-time rendering.
Unfortunately, gradient-based optimization is not directly applicable to their methods due to their topological data structures or the lack of priors.
As a result, these methods~\cite{ReiserPLG2021,WizadwongsaPYS21,GarbinKJSV2021,YuLTLNK2021,HedmanSMBD2021} still need a conversion step from a trained implicit model (\eg, NeRF) to their final representation that supports real-time rendering.
Their training time is still burdened by the lengthy implicit model optimization.

\paragraph{Fast NeRF convergence.}
Recent works that focus on fewer input views setup also bring faster convergence as a side benefit.
These methods rely on generalizable pre-training~\cite{YuYTK21,ChenXZZXYS21,WangWGSZBMSF21} or external MVS depth information~\cite{LiuPLWWTZW21,DengLZR21}, while ours does not.
Further, they still require several per-scene fine-tuning hours~\cite{DengLZR21} or fail to achieve NeRF quality in the full input-view setup~\cite{YuYTK21,ChenXZZXYS21,WangWGSZBMSF21}.
Most recently, NeuRay~\cite{LiuPLWWTZW21} shows NeRF's quality with 40 minutes per-scene training time in the lower-resolution setup.
Under the same GPU spec, our method achieves NeRF's quality in 15 minutes per scene on the high-resolution setup and does not require depth guidance and cross-scene pre-training.

\section{Preliminaries} \label{sec:pre_nerf}
To represent a 3D scene for novel view synthesis, Neural Radiance Fields (NeRFs)~\cite{MildenhallSTBRN20} employ multilayer perceptron (MLP) networks to map a 3D position $\bm{x}$ and a viewing direction $\bm{d}$ to the corresponding density $\sigma$ and view-dependent color emission $\bm{c}$:
\begin{subequations}
\label{eq:nerf_mlp}
\begin{align}
    (\sigma, \bm{e}) &= \operatorname{MLP}^{\mathrm{(pos)}}(\bm{x})~,\\
    \bm{c} &= \operatorname{MLP}^{\mathrm{(rgb)}}(\bm{e}, \bm{d})~,
\end{align}
\end{subequations}
where the learnable MLP parameters are omitted, and $\bm{e}$ is an intermediate embedding to help the much shallower $\operatorname{MLP}^{\mathrm{(rgb)}}$ to learn $\bm{c}$ (see NeRF++~\cite{ZhangRSK20} for more discussions on the architecture design).
In practice, positional encoding is applied to $\bm{x}$ and $\bm{d}$, which enables the MLPs to learn the high-frequency details from low-dimensional input~\cite{TancikSMFRSRBN20}.
For output activation, $\operatorname{Sigmoid}$ is applied on $\bm{c}$; $\operatorname{ReLU}$ or $\operatorname{Softplus}$ is applied on $\sigma$ (see Mip-NeRF~\cite{BarronMTHMS21} for more discussion on output activation).

To render the color of a pixel $\hat{C}(\bm{r})$, we cast the ray $\bm{r}$ from the camera center through the pixel; $K$ points are then sampled on $\bm{r}$ between the pre-defined near and far planes; the $K$ ordered sampled points are then used to query for their densities and colors $\{(\sigma_i, \bm{c}_i)\}_{i=1}^K$ (MLPs are queried in NeRF).
Finally, the $K$ queried results are accumulated into a single color with the volume rendering quadrature in accordance with the optical model given by Max~\cite{Max95a}:
\begin{subequations} \label{eq:volume_rendering}
\begin{align}
    \hat{C}(\bm{r}) &= \left( \sum_{i=1}^{K} T_i \alpha_i \bm{c}_i \right) + T_{\scriptscriptstyle K+1} \bm{c}_{\mathrm{bg}}  ~, \\
    \alpha_i &= \operatorname{alpha}(\sigma_i, \delta_i) = 1 - \exp(-\sigma_i \delta_i) ~, \label{eq:density_2_alpha} \\
    T_i &= \prod_{j=1}^{i-1} (1 - \alpha_j) ~, \label{eq:acc_trans}
\end{align}
\end{subequations}
where $\alpha_i$ is the probability of termination at the point $i$; $T_i$ is the accumulated transmittance from the near plane to point $i$; $\delta_i$ is the distance to the adjacent sampled point, and $\bm{c}_{bg}$ is a pre-defined background color.

Given the training images with known poses, NeRF model is trained by minimizing the photometric MSE between the observed pixel color $C(\bm{r})$ and the rendered color $\hat{C}(\bm{r})$:
\begin{equation} \label{eq:photo_loss}
    \mathcal{L}_{\mathrm{photo}} = \frac{1}{|\mathcal{R}|} \sum_{r\in\mathcal{R}} \left\|\hat{C}(\bm{r}) - C(\bm{r})\right\|_2^2 ~ ,
\end{equation}
where $\mathcal{R}$ is the set of rays in a sampled mini-batch.

\section{Post-activated density voxel grid} \label{sec:post_act_vox}

\begin{figure}
    \centering
    \includegraphics[width=\linewidth]{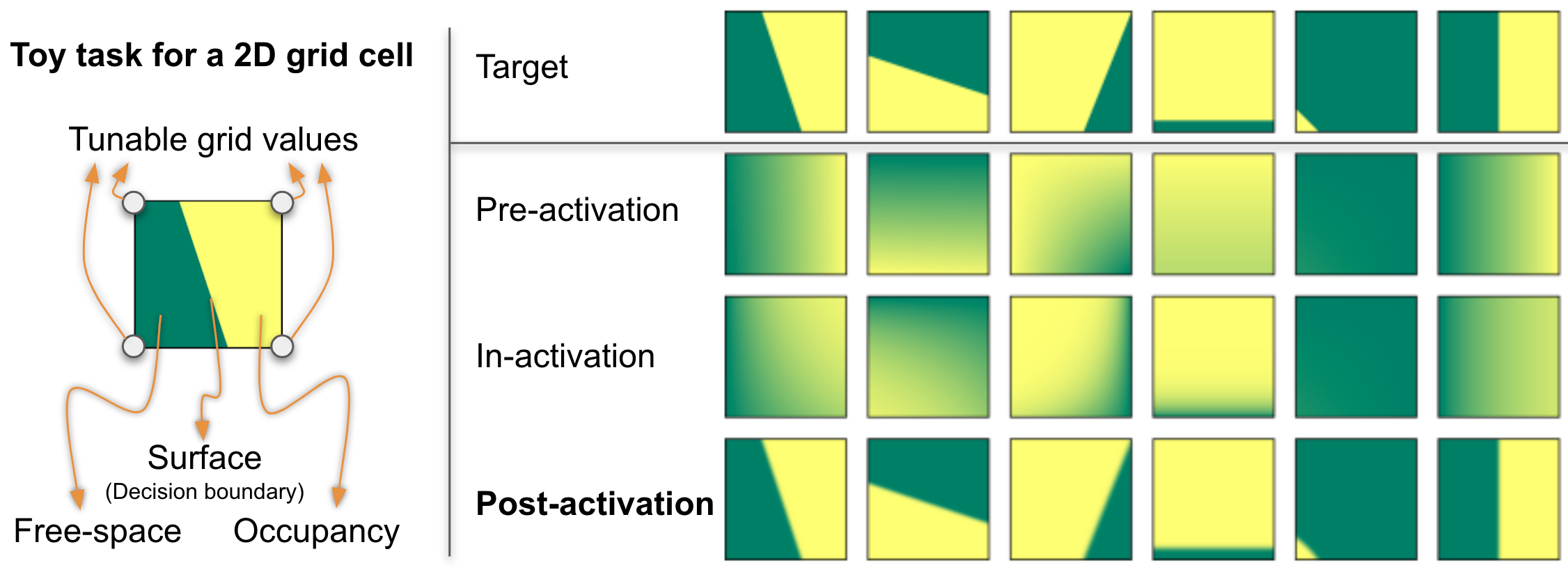}
    \caption{
    {\bf A single grid cell with post-activation is capable of modeling sharp linear surfaces.}
    {\it Left}: We depict the toy task for a 2D grid cell, where a grid cell is optimized for the linear surface (decision boundary) across it.
    {\it Right}: Each column shows an example task for three different methods.
    The results show that a single grid cell with post-activation (\cref{eq:post_activation}) is adequate to recover faithfully the linear surface.
    Conversely, pre-activation (\cref{eq:pre_activation}) and in-activation (\cref{eq:in_activation}) fail to accomplish the tasks as they can only fit into smooth results, and thus would require more grid cells to recover the surface detail.
    See supplementary material for the mathematical proof.
    }
    \label{fig:pre_in_post_patch}
\end{figure}

\paragraph{Voxel-grid representation.}
A voxel-grid representation models the modalities of interest (\eg, density, color, or feature) explicitly in its grid cells.
Such an explicit scene representation is efficient to query for any 3D positions via interpolation:
\begin{equation} \label{eq:interp}
\operatorname{interp}(\bm{x}, \bm{V}): \left(\mathbb{R}^3, \mathbb{R}^{C \times N_x \times N_y \times N_z}\right) \to \mathbb{R}^C ~,
\end{equation}
where $\bm{x}$ is the queried 3D point, $\bm{V}$ is the voxel grid, $C$ is the dimension of the modality, and $N_x\cdot N_y \cdot N_z$ is the total number of voxels.
Trilinear interpolation is applied if not specified otherwise.

\paragraph{Density voxel grid for volume rendering.}
Density voxel grid, $\bm{V}^{\text{(density)}}$, is a special case with $C=1$, which stores the density values for volume rendering (\cref{eq:volume_rendering}).
We use $\ddot{\sigma} \in \mathbb{R}$ to denote the raw voxel density before applying the density activation (\ie, a mapping of $\mathbb{R} \to \mathbb{R}_{\geq 0}$).
In this work, we use the shifted $\operatorname{softplus}$ mentioned in Mip-NeRF~\cite{BarronMTHMS21} as the density activation:
\begin{equation} \label{eq:density_activation}
    \sigma = \operatorname{softplus}(\ddot{\sigma}) = \log(1+\exp(\ddot{\sigma}+b)) ~ ,
\end{equation}
where the shift $b$ is a hyperparameter.
Using $\operatorname{softplus}$ instead of $\operatorname{ReLU}$ is crucial to optimize voxel density directly, as it is irreparable when a voxel is falsely set to a negative value with $\operatorname{ReLU}$ as the density activation.
Conversely, $\operatorname{softplus}$ allows us to explore density very close to $0$.

\begin{figure}
    \centering
    \begin{subfigure}[t]{\linewidth}
        \centering
        \includegraphics[width=\linewidth]{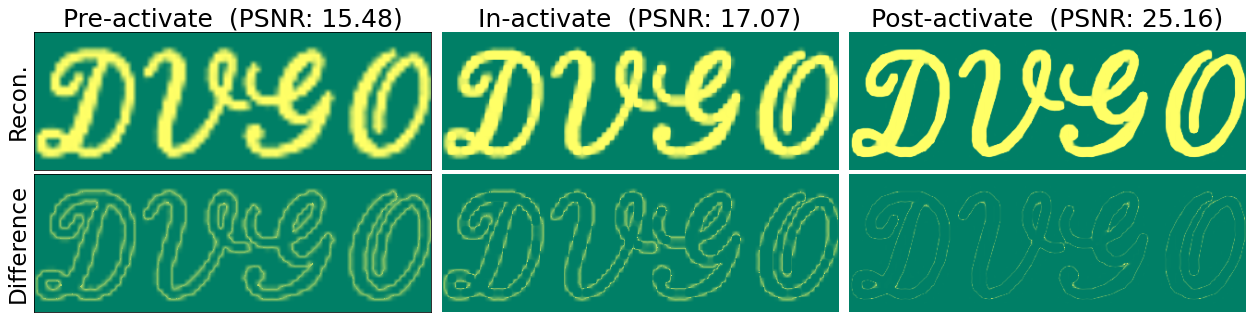}
        \caption{
        Visual comparison of image fitting results under grid resolution $(H/5) \times (W/5)$.
        The first row is the results of pre-, in-, and post-activation.
        The second row is their per-pixel absolute difference to the target image.
        }
        \label{fig:exdv_poc_a}
    \end{subfigure}
    \par\medskip
    \begin{subfigure}[t]{\linewidth}
        \centering
        \includegraphics[width=\linewidth]{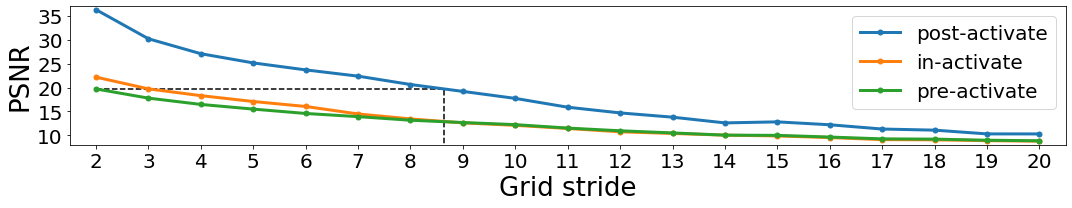}
        \caption{
        PSNRs achieved by pre-, in- and post-activation under different grid strides. A grid stride $s$ means that the grid resolution is $(H/s) \times (W/s)$.
        The black dashed line highlights that post-activation with stride $\approx 8.5$ can achieve the same PSNR as pre-activation with stride $2$ in this example.
        }
        \label{fig:exdv_poc_b}
    \end{subfigure}
    \caption{
    {\bf Toy example on image fitting.}
    The target 2D image is binary to imitate the scenario that most of the 3D space is either occupied or free.
    The objective is to reconstruct the target image by a low-resolution 2D grid.
    In each optimization step, the tunable 2D grid is queried by interpolation with pre-activation (\cref{eq:pre_activation}), in-activation (\cref{eq:in_activation}), or post-activation (\cref{eq:post_activation}) to minimize the mean squared error to the target image.
    The result reveals that the {\it post-activation} can produce sharp boundaries even with low grid resolution (\cref{fig:exdv_poc_a}) and is much better than the other two under various grid resolutions (\cref{fig:exdv_poc_b}).
    This motivates us to model the 3D geometry directly via voxel grids with {\it post-activation}.
    }
    \label{fig:exdv_poc}
\end{figure}

\paragraph{Sharp decision boundary via post-activation.}
The interpolated voxel density is processed by $\operatorname{softplus}$ (\cref{eq:density_activation}) and $\operatorname{alpha}$ (\cref{eq:density_2_alpha}) functions sequentially for volume rendering.
We consider three different orderings---pre-activation, in-activation, and post-activation---of plugging in the trilinear interpolation and performing the activation, given a queried 3D point $\bm{x}$:
\begin{subequations} \label{eq:pre_in_post}
\begin{align}
    \alpha^{\text{(pre)}} &= \colorinterp\left(\bm{x}, \coloralpha\left(\colorsoftplus\left(\bm{V}^{\text{(density)}}\right)\right)\right) ~ , \label{eq:pre_activation} \\
    \alpha^{\text{(in)}} &= \coloralpha\left(\colorinterp\left(\bm{x}, \colorsoftplus\left(\bm{V}^{\text{(density)}}\right)\right)\right) ~ , \label{eq:in_activation} \\
    \alpha^{\text{(post)}} &= \coloralpha\left(\colorsoftplus\left(\colorinterp\left(\bm{x}, \bm{V}^{\text{(density)}}\right)\right)\right) ~. \label{eq:post_activation}
\end{align}
\end{subequations}
The input $\delta$ to the function $\operatorname{alpha}$ (\cref{eq:density_2_alpha}) is omitted for simplicity.
We show that the post-activation, \ie, applying all the non-linear activation after the trilinear interpolation, is capable of producing sharp surfaces (decision boundaries) with much fewer grid cells.
In \cref{fig:pre_in_post_patch}, we use a 2D grid cell as an example to show that a grid cell with post-activation can produce a sharp linear boundary, while pre- and in-activation can only produce smooth results and thus require more cells for the surface detail.
In \cref{fig:exdv_poc}, we further use binary image regression as a toy example to compare their capability, which also shows that post-activation can achieve a much better efficiency in grid cell usage.

\section{Fast and direct voxel grid optimization} \label{sec:approach}
We depict an overview of our approach in \cref{fig:overview}.
In \cref{ssec:coarse_learn}, we first search the coarse geometry of a scene.
In \cref{ssec:fine_learn}, we then reconstruct the fine detail including view-dependent effects.
Hereinafter we use superscripts $^{\text{(c)}}$ and $^{\text{(f)}}$ to denote variables in the coarse and fine stages.

\subsection{Coarse geometry searching} \label{ssec:coarse_learn}
Typically, a scene is dominated by free space (\ie, unoccupied space).
Motivated by this fact, we aim to efficiently find the coarse 3D areas of interest before reconstructing the fine detail and view-dependent effect that require more computation resources.
We can thus greatly reduce the number of queried points on each ray in the later fine stage.

\paragraph{Coarse scene representation.}
We use a coarse density voxel grid  $\bm{V}^{\text{(density)(c)}} \in \mathbb{R}^{1 \times N_x^{\text{(c)}} \times N_y^{\text{(c)}} \times N_z^{\text{(c)}}}$ with post-activation (\cref{eq:post_activation}) to model scene geometry.
We only model view-invariant color emissions by $\bm{V}^{\text{(rgb)(c)}} \in \mathbb{R}^{3 \times N_x^{\text{(c)}} \times N_y^{\text{(c)}} \times N_z^{\text{(c)}}}$ in the coarse stage.
A query of any 3D point $\bm{x}$ is efficient with interpolation:
\begin{subequations} \label{eq:coarse_queries}
\begin{align}
    \ddot{\sigma}^{\text{(c)}} &= \operatorname{interp}\left(\bm{x}, \bm{V}^{\text{(density)(c)}}\right)~, \label{eq:coarse_queries_densities} \\
    \bm{c}^{\text{(c)}} &= \operatorname{interp}\left(\bm{x}, \bm{V}^{\text{(rgb)(c)}}\right)~,
\end{align}
\end{subequations}
where $\bm{c}^{\text{(c)}} \in \mathbb{R}^3$ is the view-invariant color and $\ddot{\sigma}^{\text{(c)}} \in \mathbb{R}$ is the raw volume density.

\paragraph{Coarse voxels allocation.}
We first find a bounding box (BBox) tightly enclosing the camera frustums of training views (See the red BBox in \cref{fig:overview}c for an example).
Our voxel grids are aligned with the BBox.
Let $L_x^{\text{(c)}}, L_y^{\text{(c)}}, L_z^{\text{(c)}}$ be the lengths of the BBox and $M^{\text{(c)}}$ be the hyperparameter for the expected total number of voxels in the coarse stage.
The voxel size is $s^{\text{(c)}} = \sqrt[3]{L_x^{\text{(c)}}\cdot L_y^{\text{(c)}}\cdot L_z^{\text{(c)}} / M^{\text{(c)}}}$, so there are $N_x^{\text{(c)}}, N_y^{\text{(c)}}, N_z^{\text{(c)}} = \lfloor L_x^{\text{(c)}} / s^{\text{(c)}} \rfloor, \lfloor L_y^{\text{(c)}} / s^{\text{(c)}} \rfloor, \lfloor L_z^{\text{(c)}} / s^{\text{(c)}} \rfloor$ voxels on each side of the BBox.

\paragraph{Coarse-stage points sampling.}
On a pixel-rendering ray, we sample query points as
\begin{subequations} \label{eq:coarse_sample}
\begin{align} 
    \bm{x}_0 &= \bm{o} + t^{\text{(near)}} \bm{d} ~,\\
    \bm{x}_i &= \bm{x}_0 + i \cdot \delta^{\text{(c)}} \cdot \frac{\bm{d}}{\|\bm{d}\|^2} ~,
\end{align}
\end{subequations}
where $\bm{o}$ is the camera center, $\bm{d}$ is the ray-casting direction, $t^{\text{(near)}}$ is the camera near bound, and $\delta^{\text{(c)}}$ is a hyperparameter for the step size that can be adaptively chosen according to the voxel size $s^{(c)}$.
The query index $i$ ranges from $1$ to $\lceil t^{\text{(far)}} \cdot \|\bm{d}\|^2 / \delta^{\text{(c)}} \rceil$, where $t^{\text{(far)}}$ is the camera far bound, so the last sampled point stops nearby the far plane.

\paragraph{Prior 1: low-density initialization.}
At the start of training, the importance of points far from a camera is down-weighted due to the accumulated transmittance term in \cref{eq:acc_trans}.
As a result, the coarse density voxel grid $\bm{V}^{\text{(density)(c)}}$ could be accidentally trapped into a suboptimal ``cloudy" geometry with higher densities at camera near planes.
We thus have to initialize $\bm{V}^{\text{(density)(c)}}$ more carefully to ensure that all sampled points on rays are visible to the cameras at the beginning, \ie, the accumulated transmittance rates $T_i$s in \cref{eq:acc_trans} are close to $1$.

In practice, we initialize all grid values in $\bm{V}^{\text{(density)(c)}}$ to $0$ and set the bias term in \cref{eq:density_activation} to
\begin{equation} \label{eq:low_density_init}
    b = \log\left({\left(1 - \alpha^{\text{(init)(c)}}\right)^{-\frac{1}{s^{\text{(c)}}}}} - 1\right) ~,
\end{equation}
where $\alpha^{\text{(init)(c)}}$ is a hyperparameter.
Thereby, the accumulated transmittance $T_i$ is decayed by $1 - \alpha^{\text{(init)(c)}} \approx 1$ for a ray that traces forward a distance of a voxel size $s^{\text{(c)}}$.
See supplementary material for the derivation and proof.

\paragraph{Prior 2: view-count-based learning rate.}
There could be some voxels visible to too few training views in real-world capturing, while we prefer a surface with consistency in many views instead of a surface that can only explain few views.
In practice, we set different learning rates for different grid points in $\bm{V}^{\text{(density)(c)}}$.
For each grid point indexed by $j$, we count the number of training views $n_j$ to which point $j$ is visible, and then scale its base learning rate by $n_j / n_{\text{max}}$, where $n_{\text{max}}$ is the maximum view count over all grid points.

\paragraph{Training objective for coarse representation.}
The scene representation is reconstructed by minimizing the mean square error between the rendered and observed colors.
To regularize the reconstruction, we mainly use background entropy loss to encourage the accumulated alpha values to concentrate on background or foreground.
Please refer to supplementary material for more detail.


\subsection{Fine detail reconstruction} \label{ssec:fine_learn}
Given the optimized coarse geometry $\bm{V}^{\text{(density)(c)}}$ in \cref{ssec:coarse_learn}, we now can focus on a smaller subspace to reconstruct the surface details and view-dependent effects.
The optimized $\bm{V}^{\text{(density)(c)}}$ is frozen in this stage.

\paragraph{Fine scene representation.}
In the fine stage, we use a higher-resolution density voxel grid $\bm{V}^{\text{(density)(f)}} \in \mathbb{R}^{1 \times N_x^{\text{(f)}} \times N_y^{\text{(f)}} \times N_z^{\text{(f)}}}$ with post-activated interpolation (\cref{eq:post_activation}).
Note that, alternatively, it is also possible to use a more advanced data structure~\cite{YuLTLNK2021,LiuGLCT20,HedmanSMBD2021} to refine the voxel grid based on the current $\bm{V}^{\text{(density)(c)}}$ but we leave that for future work.
To model view-dependent color emission, we opt to use an explicit-implicit hybrid representation as we find in our prior experiments that an explicit representation tends to produce worse results, and an implicit representation entails a slower training speed.
Our hybrid representation comprises \emph{i}) a feature voxel grid $\bm{V}^{\text{(feat)(f)}} \in \mathbb{R}^{D \times N_x^{\text{(f)}} \times N_y^{\text{(f)}} \times N_z^{\text{(f)}}}$, where $D$ is a hyperparameter for feature-space dimension, and \emph{ii}) a shallow MLP parameteriszed by $\Theta$.
Finally, queries of 3D points $\bm{x}$ and viewing-direction $\bm{d}$ are performed by
\begin{subequations} \label{eq:fine_queries}
\begin{align}
    \ddot{\sigma}^{\text{(f)}} &= \operatorname{interp}\left(\bm{x}, \bm{V}^{\text{(density)(f)}}\right)~, \label{eq:fine_queries_densities} \\
    \bm{c}^{\text{(f)}} &= \operatorname{MLP}_{\Theta}^{\text{(rgb)}}\left(\operatorname{interp}(\bm{x}, \bm{V}^{\text{(feat)(f)}}), \bm{x}, \bm{d}\right) ~, \label{eq:fine_queries_rgb}
\end{align}
\end{subequations}
where $\bm{c}^{\text{(f)}} \in \mathbb{R}^3$ is the view-dependent color emission and $\ddot{\sigma}^{\text{(f)}} \in \mathbb{R}$ is the raw volume density in the fine stage.
Positional embedding~\cite{MildenhallSTBRN20} is applied on $\bm{x}, \bm{d}$ for the $\operatorname{MLP}_{\Theta}^{\text{(rgb)}}$.

\paragraph{Known free space and unknown space.}
A query point is in the known free space if the post-activated alpha value from the optimized $\bm{V}^{\text{(density)(c)}}$ is less than the threshold $\tau^{\text{(c)}}$.
Otherwise, we say the query point is in the unknown space.

\paragraph{Fine voxels allocation.}
We densely query $\bm{V}^{\text{(density)(c)}}$ to find a BBox tightly enclosing the unknown space, where $L_x^{\text{(f)}}, L_y^{\text{(f)}}, L_z^{\text{(f)}}$ are the lengths of the BBox.
The only hyperparameter is the expected total number of voxels $M^{\text{(f)}}$.
The voxel size $s^{\text{(f)}}$ and the grid dimensions $N_x^{\text{(f)}}, N_y^{\text{(f)}}, N_z^{\text{(f)}}$ can then be derived automatically from $M^{\text{(f)}}$ as per \cref{ssec:coarse_learn}.

\paragraph{Progressive scaling.}
Inspired by NSVF~\cite{LiuGLCT20}, we progressively scale our voxel grid $\bm{V}^{\text{(density)(f)}}$ and $\bm{V}^{\text{(feat)(f)}}$.
Let $\mathrm{pg\_ckpt}$ be the set of checkpoint steps.
The initial number of voxels is set to $\lfloor M^{\text{(f)}} / 2^{|\mathrm{pg\_ckpt}|} \rfloor$.
When reaching the training step in $\mathrm{pg\_ckpt}$, we double the number of voxels such that the number of voxels after the last checkpoint is $M^{\text{(f)}}$; the voxel size $s^{\text{(f)}}$ and the grid dimensions $N_x^{\text{(f)}}, N_y^{\text{(f)}}, N_z^{\text{(f)}}$ are updated accordingly.
Scaling our scene representation is much simpler.
At each checkpoint, we resize our voxel grids, $\bm{V}^{\text{(density)(f)}}$ and $\bm{V}^{\text{(feat)(f)}}$, by trilinear interpolation.

\paragraph{Fine-stage points sampling.}
The points sampling strategy is similar to \cref{eq:coarse_sample} with some modifications.
We first filter out rays that do not intersect with the known free space.
For each ray, we adjust the near- and far-bound, $t^{\text{(near)}}$ and $t^{\text{(far)}}$, to the two endpoints of the ray-box intersection.
We do not adjust $t^{\text{(near)}}$ if $\bm{x}_0$ is already inside the BBox.

\paragraph{Free space skipping.}
Querying $\bm{V}^{\text{(density)(c)}}$ (\cref{eq:coarse_queries_densities}) is faster than querying $\bm{V}^{\text{(density)(f)}}$ (\cref{eq:fine_queries_densities}); querying for view-dependent colors (\cref{eq:fine_queries_rgb}) is the slowest.
We improve fine-stage efficiency by free space skipping in both training and testing.
First, we skip sampled points that are in the known free space by checking the optimized $\bm{V}^{\text{(density)(c)}}$ (\cref{eq:coarse_queries_densities}).
Second, we further skip sampled points in unknown space with low activated alpha value (threshold at $\tau^{\text{(f)}}$) by querying $\bm{V}^{\text{(density)(f)}}$ (\cref{eq:fine_queries_densities}).

\paragraph{Training objective for fine representation.}
We use the same training losses as the coarse stage, but we use a smaller weight for the regularization losses as we find it empirically leads to slightly better quality.

\section{Experiments}

\subsection{Datasets}
We evaluate our approach on five inward-facing datasets.
{\bf Synthetic-NeRF}~\cite{MildenhallSTBRN20} contains eight objects with realistic images synthesized by NeRF.
{\bf Synthetic-NSVF}~\cite{LiuGLCT20} contains another eight objects synthesized by NSVF.
Strictly following NeRF's and NSVF's setups, we set the image resolution to $800 \times 800$ pixels and let each scene have 100 views for training and 200 views for testing.
{\bf BlendedMVS}~\cite{YaoLLZRZFQ20} is a synthetic MVS dataset that has realistic ambient lighting from real image blending.
We use a subset of four objects provided by NSVF.
The image resolution is $768 \times 576$ pixels, and one-eighth of the images are for testing.
{\bf Tanks\&Temples}~\cite{KnapitschPZK17} is a real-world dataset.
We use a subset of five scenes provided by NSVF, each containing views captured by an inward-facing camera circling the scene.
The image resolution is $1920 \times 1080$ pixels, and one-eighth of the images are for testing.
{\bf DeepVoxels}~\cite{SitzmannTHNWZ19} dataset contains four simple Lambertian objects.
The image resolutions are $512 \times 512$, and each scene has 479 views for training and 1000 views for testing.

\subsection{Implementation details} \label{ssec:implementation}
We choose the same hyperparameters generally for all scenes.
The expected numbers of voxels are set to $M^{\text{(c)}}=100^3$ and $M^{\text{(f)}}=160^3$ in coarse and fine stages if not stated otherwise.
The activated alpha values are initialized to be $\alpha^{\text{(init)(c)}}=10^{-6}$ in the coarse stage.
We use a higher $\alpha^{\text{(init)(f)}}=10^{-2}$ as the query points are concentrated on the optimized coarse geometry in the fine stage.
The points sampling step sizes are set to half of the voxel sizes, \ie, $\delta^{\text{(c)}} = 0.5\cdot s^{\text{(c)}}$ and $\delta^{\text{(f)}} = 0.5\cdot s^{\text{(f)}}$.
The shallow MLP layer comprises two hidden layers with $128$ channels.
We use the Adam optimizer~\cite{KingmaB14} with a batch size of $8{,}192$ rays to optimize the coarse and fine scene representations for $10$k and $20$k iterations.
The base learning rates are $0.1$ for all voxel grids and $10^{-3}$ for the shallow MLP.
The exponential learning rate decay is applied.
See supplementary material for detailed hyperparameter setups.

\subsection{Comparisons} \label{ssec:comp_qual}

\begin{table*}[tbp]
    \centering
    \bgroup
    \small
    \begin{tabular}{l@{\hskip 5.5pt}|@{\hskip 5.5pt}c@{\hskip 5.5pt}c@{\hskip 5.5pt}c@{\hskip 5.5pt}|@{\hskip 5.5pt}c@{\hskip 5.5pt}c@{\hskip 5.5pt}c@{\hskip 5.5pt}|@{\hskip 5.5pt}c@{\hskip 5.5pt}c@{\hskip 5.5pt}c@{\hskip 5.5pt}|@{\hskip 5.5pt}c@{\hskip 5.5pt}c@{\hskip 5.5pt}c@{}}
    \hline
    \multirow{2}{*}{Methods} & \multicolumn{3}{c@{\hskip 5.5pt}|@{\hskip 5.5pt}}{Synthetic-NeRF} & \multicolumn{3}{c@{\hskip 5.5pt}|@{\hskip 5.5pt}}{Synthetic-NSVF} & \multicolumn{3}{c@{\hskip 5.5pt}|@{\hskip 5.5pt}}{BlendedMVS} & \multicolumn{3}{c}{Tanks and Temples} \\
     & PSNR$\uparrow$ & SSIM$\uparrow$ & LPIPS$\downarrow$ & PSNR$\uparrow$ & SSIM$\uparrow$ & LPIPS$\downarrow$ & PSNR$\uparrow$ & SSIM$\uparrow$ & LPIPS$\downarrow$ & PSNR$\uparrow$ & SSIM$\uparrow$ & LPIPS$\downarrow$ \\
    \hline
    \hline
    SRN~\cite{SitzmannZW19}                   & 22.26 & 0.846 & 0.170\textsuperscript{vgg}  &  24.33 & 0.882 & 0.141\textsuperscript{alex}  &  20.51 & 0.770 & 0.294\textsuperscript{alex}  &  24.10 & 0.847 & 0.251\textsuperscript{alex} \\
    NV~\cite{LombardiSSSLS19}                 & 26.05 & 0.893 & 0.160\textsuperscript{vgg}  &  25.83 & 0.892 & 0.124\textsuperscript{alex}  &  23.03 & 0.793 & 0.243\textsuperscript{alex}  &  23.70 & 0.834 & 0.260\textsuperscript{alex} \\
    NeRF~\cite{MildenhallSTBRN20}             & 31.01 & 0.947 & 0.081\textsuperscript{vgg}  &  30.81 & 0.952 & 0.043\textsuperscript{alex}  &  24.15 & 0.828 & 0.192\textsuperscript{alex}  &  25.78 & 0.864 & 0.198\textsuperscript{alex} \\
    \hline
    \hline
    \multicolumn{13}{@{}l}{Improved visual quality from NeRF} \\
    \hline
    JaxNeRF~\cite{jaxnerf2020github}          & 31.69 & 0.953 & 0.068\textsuperscript{vgg}  &  -     & -     & -      &  -     & -     & -      &  27.94 & 0.904 & 0.168\textsuperscript{vgg} \\
    JaxNeRF+~\cite{jaxnerf2020github}         & 33.00 & 0.962 & \textcolor{gray}{0.038}  &  -     & -     & -      &  -     & -     & -      &  -     & -     & -     \\
    Mip-NeRF~\cite{BarronMTHMS21}             & 33.09 & 0.961 & 0.043\textsuperscript{vgg}  &  -     & -     & -      &  -     & -     & -      &  -     & -     & -     \\
    \hline
    \hline
    \multicolumn{13}{@{}l}{Improved test-time rendering speed (and visual quality) from NeRF} \\
    \hline
    AutoInt~\cite{LindellMW2021}              & 25.55 & 0.911 & \textcolor{gray}{0.170}  &  -     & -     & -      &  -     & -     & -      &  -     & -     & -     \\
    FastNeRF~\cite{GarbinKJSV2021}            & 29.97 & 0.941 & \textcolor{gray}{0.053}  &  -     & -     & -      &  -     & -     & -      &  -     & -     & -     \\
    SNeRG~\cite{HedmanSMBD2021}               & 30.38 & 0.950 & \textcolor{gray}{0.050}  &  -     & -     & -      &  -     & -     & -      &  -     & -     & -     \\
    KiloNeRF~\cite{ReiserPLG2021}             & 31.00 & 0.95  & \textcolor{gray}{0.03}   &  33.37 & 0.97  & \textcolor{gray}{0.02}   &  27.39 & 0.92  & \textcolor{gray}{0.06}   &  28.41 & 0.91  & \textcolor{gray}{0.09}  \\
    PlenOctrees~\cite{YuLTLNK2021}            & 31.71 & 0.958 & 0.053\textsuperscript{vgg}  &  -     & -     & -      &  -     & -     & -      &  27.99 & 0.917 & 0.131\textsuperscript{vgg} \\
    NSVF~\cite{LiuGLCT20}                     & 31.75 & 0.953 & 0.047\textsuperscript{alex}  &  35.18 & 0.979 & 0.015\textsuperscript{alex}  &  26.89 & 0.898 & 0.114\textsuperscript{alex}  &  28.48 & 0.901 & 0.155\textsuperscript{alex} \\
    \hline
    \hline
    \multicolumn{13}{@{}l}{Improved convergence speed, test-time rendering speed, and visual quality from NeRF} \\
    \hline
    ours {\scriptsize ($M^{\text{(f)}}$=$160^3$)} & 31.95 & 0.957 & \makecell{0.053\textsuperscript{vgg}\\0.035\textsuperscript{alex}}  &  35.08 & 0.975 & \makecell{0.033\textsuperscript{vgg}\\0.019\textsuperscript{alex}}  &  28.02 & 0.922 & \makecell{0.101\textsuperscript{vgg}\\0.075\textsuperscript{alex}}  &  28.41 & 0.911 & \makecell{0.155\textsuperscript{vgg}\\0.148\textsuperscript{alex}} \\
    \cline{2-13}
    ours {\scriptsize ($M^{\text{(f)}}$=$256^3$)} & 32.80 & 0.961 & \makecell{0.045\textsuperscript{vgg}\\0.027\textsuperscript{alex}}  &  36.21 & 0.980 & \makecell{0.024\textsuperscript{vgg}\\0.012\textsuperscript{alex}}  &  28.64 & 0.933 & \makecell{0.081\textsuperscript{vgg}\\0.052\textsuperscript{alex}}  &  28.82 & 0.920 & \makecell{0.138\textsuperscript{vgg}\\0.124\textsuperscript{alex}} \\
    \hline
    \multicolumn{13}{p{\dimexpr\linewidth-2\tabcolsep-2\arrayrulewidth}}{\scriptsize{\textsuperscript{*}~The superscript denotes the pre-trained models used in LPIPS. The \textcolor{gray}{gray} numbers indicate that the code is unavailable or has a unconventional LPIPS implementation.}}
    \end{tabular}
    \egroup
    \caption{
    {\bf Quantitative comparisons for novel view synthesis.}
    Our method excels in convergence speed, \ie, 15 minutes per scene compared to many hours or days per scene using other methods.
    Besides, our rendering quality is better than the original NeRF~\cite{MildenhallSTBRN20} and the improved JaxNeRF~\cite{jaxnerf2020github} on the four datasets under all metrics.
    We also show comparable results to most of the recent methods.
    }
    \label{tab:results}
    \vspace{-1em}
\end{table*}

\paragraph{Quantitative evaluation on the synthesized novel view.}
We first quantitatively compare the novel view synthesis results in \cref{tab:results}.
PSNR, SSIM~\cite{WangBSS04}, and LPIPS~\cite{ZhangIESW18} are employed as evaluation metrics.
Our model with $M^{\text{(f)}}=160^3$ voxels already outperforms the original NeRF~\cite{MildenhallSTBRN20} and the improved JaxNeRF~\cite{jaxnerf2020github} re-implementation.
Besides, our results are also comparable to most of the recent methods, except JaxNeRF+~\cite{jaxnerf2020github} and Mip-NeRF~\cite{BarronMTHMS21}.
Moreover, our per-scene optimization only takes about 15 minutes, while all the methods after NeRF in \cref{tab:results} need quite a few hours per scene.
We also show our model with $M^{\text{(f)}}=256^3$ voxels, which significantly improves our results under all metrics and achieves more comparable results to JaxNeRF+ and Mip-NeRF.
We defer detail comparisons on the much simpler DeepVoxels~\cite{SitzmannTHNWZ19} dataset to supplementary material, where we achieve $45.83$ averaged PSNR and outperform NeRF's $40.15$ and IBRNet's $42.93$.

\paragraph{Training time comparisons.}
The key merit of our work is the significant improvement in convergence speed with NeRF-comparable quality.
In \cref{tab:train_speed}, we show a training time comparison.
We also show GPU specifications after each reported time as it is the main factor affecting run-time.

NeRF~\cite{MildenhallSTBRN20} with a more powerful GPU needs 1$\mbox{--}$2 days per scene to achieve $31.01$ PSNR, while our method achieves a superior $31.95$ and $32.80$ PSNR in about $15$ an $22$ minutes per scene respectively.
MVSNeRF~\cite{ChenXZZXYS21}, IBRNet~\cite{WangWGSZBMSF21}, and NeuRay~\cite{LiuPLWWTZW21} also show less per-scene training time than NeRF but with the additional cost to run a generalizable cross-scene pre-training.
MVSNeRF~\cite{ChenXZZXYS21}, after pre-training, optimizes a scene in 15 minutes as well, but the PSNR is degraded to $28.14$.
IBRNet~\cite{WangWGSZBMSF21} shows worse PSNR and longer training time than ours.
NeuRay~\cite{LiuPLWWTZW21} originally reports time in lower-resolution (NeuRay-Lo) setup, and we receive the training time of the high-resolution (NeuRay-Hi) setup from the authors.
NeuRay-Hi achieves $32.42$ PSNR and requires $23$ hours to train, while our method with $M^{\text{(f)}}=256^3$ voxels achieves superior $32.80$ in about $22$ minutes.
For the early-stopped NeuRay-Hi, unfortunately, only its training time is retained (early-stopped NeuRay-Lo achieves NeRF-similar PSNR).
NeuRay-Hi still needs $70$ minutes to train with early stopping, while we only need $15$ minutes to achieve NeRF-comparable quality and do not rely on generalizable pre-training or external depth information.
Mip-NeRF~\cite{BarronMTHMS21} has similar run-time to NeRF but with much better PSNRs, which also signifies using less training time to achieve NeRF's PSNR.
We train early-stopped Mip-NeRFs on our machine and show the averaged PSNR and training time.
The early-stopped Mip-NeRF achieves $30.85$ PSNR after $6$ hours of training, while we can achieve $31.95$ PSNR in just $15$ minutes.

\paragraph{Rendering speed comparisons.}
Improving test-time rendering speed is not the main focus of this work, but we still achieve $\sim 45\times$ speedups from NeRF---$0.64$ seconds versus $29$ seconds per $800\times 800$ image on our machine.

\begin{table}[tbp]
    \centering
    \bgroup
    \small
    \begin{tabular}{@{}l|c|c@{\hskip 3pt}l|c@{\hskip 3pt}l@{}}
    \hline
    Methods & PSNR$\uparrow$ & \multicolumn{2}{c|}{\makecell{generalizable\\pre-training}} & \multicolumn{2}{c}{\makecell{per-scene\\optimization}} \\
    \hline\hline
    NeRF~\cite{MildenhallSTBRN20} & 31.01 & \multicolumn{2}{c|}{no need} & 1$\mbox{--}$2 days & {\scriptsize (V100)} \\
    MVSNeRF~\cite{ChenXZZXYS21} & 27.21 & 30 hrs & {\scriptsize (2080Ti)} & 15 mins & {\scriptsize (2080Ti)} \\
    IBRNet~\cite{WangWGSZBMSF21} & 28.14 & 1 day & {\scriptsize (8xV100)} & 6 hrs & {\scriptsize (V100)} \\
    NeuRay~\cite{LiuPLWWTZW21}\textdagger & 32.42 & 2 days & {\scriptsize (2080Ti)} & 23 hrs & {\scriptsize (2080Ti)} \\
    Mip-NeRF~\cite{BarronMTHMS21}\textdaggerdbl & 30.85 & \multicolumn{2}{c|}{no need} & 6 hrs & {\scriptsize (2080Ti)} \\
    \hline
    ours {\scriptsize ($M^{\text{(f)}}$=$160^3$)} & 31.95 & \multicolumn{2}{c|}{no need} & 15 mins & {\scriptsize (2080Ti)} \\
    ours {\scriptsize ($M^{\text{(f)}}$=$256^3$)} & 32.80 & \multicolumn{2}{c|}{no need} & 22 mins & {\scriptsize (2080Ti)} \\
    \hline
    \multicolumn{6}{l}{\textdagger~\footnotesize{Use external depth information.}}\\
    \multicolumn{6}{l}{\textdaggerdbl~\footnotesize{Our reproduction with early stopping on our machine.}}
    \end{tabular}
    \egroup
    \caption{
    {\bf Training time comparisons.}
    We take the training time and GPU specifications reported in previous works directly.
    A V100 GPU can run faster and has more storage than a 2080Ti GPU.
    Our method achieves good PSNR in a significantly less per-scene optimization time.
    }
    \label{tab:train_speed}
    \vspace{-1.5em}
\end{table}

\begin{figure}
    \centering
    \begin{tabular}{@{}c@{\hskip 1.5pt}c@{\hskip 1.5pt}c@{\hskip 1.5pt}c@{\hskip 1.5pt}c@{\hskip 1.5pt}c@{}}
    {\footnotesize GT} & {\footnotesize Ours} & {\footnotesize PlenOctree} & {\footnotesize GT} & {\footnotesize Ours} & {\footnotesize PlenOctree} \\

    \includegraphics[width=.16\linewidth]{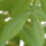} &
    \includegraphics[width=.16\linewidth]{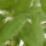} &
    \includegraphics[width=.16\linewidth]{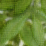} &
    \includegraphics[width=.16\linewidth]{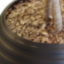} &
    \includegraphics[width=.16\linewidth]{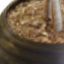} &
    \includegraphics[width=.16\linewidth]{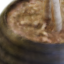} \\[-2pt]
    
    \includegraphics[width=.16\linewidth]{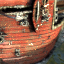} &
    \includegraphics[width=.16\linewidth]{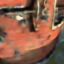} &
    \includegraphics[width=.16\linewidth]{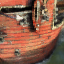} &
    \includegraphics[width=.16\linewidth]{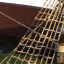} &
    \includegraphics[width=.16\linewidth]{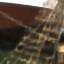} &
    \includegraphics[width=.16\linewidth]{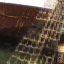} \\[-2pt]
    
    \includegraphics[width=.16\linewidth]{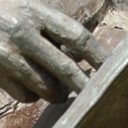} &
    \includegraphics[width=.16\linewidth]{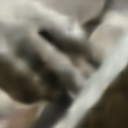} &
    \includegraphics[width=.16\linewidth]{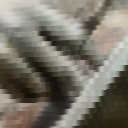} &
    \includegraphics[width=.16\linewidth]{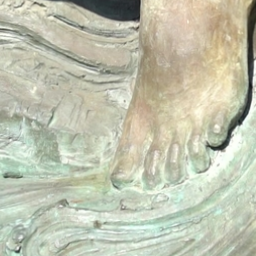} &
    \includegraphics[width=.16\linewidth]{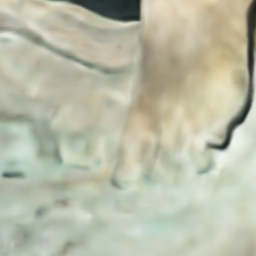} &
    \includegraphics[width=.16\linewidth]{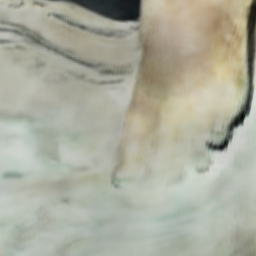} \\
    \end{tabular}
    \caption{
    {\bf Qualitative comparisons on the challenging parts.}
    {\it Top}: On {\it ficus} scene, we do not show blocking artifacts as PlenOctree and recover the pot better.
    {\it Middle}: We produce blurrier results on {\it ship}'s body and rigging, but we do not have the background artifacts.
    {\it Bottom}: On real-world captured {\it Ignatius}, we show better quality without blocking artifacts (left) and recover the color tone better (right).
    See supplementary material for more visualizations.
    }
    \label{fig:qual}
    \vspace{-1em}
\end{figure}

\paragraph{Qualitative comparison.}
\cref{fig:qual} shows our rendering results on the challenging parts and compare them with the results (better than NeRF's) provided by PlenOctrees~\cite{YuLTLNK2021}.

\subsection{Ablation studies} \label{ssec:ablation}
We mainly validate the effectiveness of the two proposed techniques---post-activation and the imposed priors---that enable voxel grids to model scene geometry with NeRF-comparable quality.
We subsample two scenes for each dataset.
See supplementary material for more detail and additional ablation studies on the number of voxels, point-sampling step size, progressive scaling, free space skipping, view-dependent colors modeling, and the losses.

\paragraph{Effectiveness of the post-activation.}
We show in \cref{sec:post_act_vox} that the proposed post-activated trilinear interpolation enables the discretized grid to model sharper surfaces.
In \cref{tab:interp}, we compare the effectiveness of post-activation in scene reconstruction for novel view synthesis.
Our grid in the fine stage consists of only $160^3$ voxels, where nearest-neighbor interpolation results in worse quality than trilinear interpolation.
The proposed post-activation can improve the results further compared to pre- and in-activation.
We find that we gain less in the real-world captured {\it BlendedMVS} and {\it Tanks and Temples} datasets.
The intuitive reason is that real-world data introduces more uncertainty (\eg, inconsistent lightning, SfM error), which results in multi-view inconsistent and blurrier surfaces.
Thus, the advantage is lessened for scene representations that can model sharper surfaces.
We speculate that resolving the uncertainty in future work can increase the gain of the proposed post-activation.

\paragraph{Effectiveness of the imposed priors.}
As discussed in \cref{ssec:coarse_learn}, it is crucial to initialize the voxel grid with low density to avoid suboptimal geometry.
The hyperparameter $\alpha^{\text{(init)(c)}}$ controls the initial activated alpha values via \cref{eq:low_density_init}.
In \cref{tab:priors}, we compare the quality with different $\alpha^{\text{(init)(c)}}$ and the view-count-based learning rate.
Without the low-density initialization, the quality drops severely for all the scenes.
When $\alpha^{\text{(init)(c)}}=10^{-7}$, we have to train the coarse stage of some scenes for more iterations.
The effective range of $\alpha^{\text{(init)(c)}}$ is scene-dependent.
We find $\alpha^{\text{(init)(c)}}=10^{-6}$ generally works well on all the scenes in this work.
Finally, using a view-count-based learning rate can further improve the results and allocate noiseless voxels in the coarse stage.

\begin{table}[tbp]
    \centering
    \bgroup
    \setlength{\tabcolsep}{3pt}
    \footnotesize
    \begin{tabular}{@{}cl|c@{\hskip 3pt}c|c@{\hskip 3pt}c|c@{\hskip 3pt}c|c@{\hskip 3pt}c@{}}
    \hline
    \multicolumn{2}{c|}{\multirow{2}{*}{Interp.}} &
    \multicolumn{2}{c|}{\it Syn.-NeRF} &
    \multicolumn{2}{c|}{\it Syn.-NSVF} &
    \multicolumn{2}{c|}{\it BlendedMVS} &
    \multicolumn{2}{c}{\it T\&T} \\
    & & {PSNR$\uparrow$} & {$\Delta$} & {PSNR$\uparrow$} & {$\Delta$} & {PSNR$\uparrow$} & {$\Delta$} & {PSNR$\uparrow$} & {$\Delta$} \\
    \hline\hline
    \multicolumn{2}{c|}{Nearest} & 28.61 & -2.77 & 28.86 & -6.22 & 25.49 & -2.48 & 26.39 & -1.27 \\
    \cline{1-2}
    \multirow{3}{*}{Tri.} & {pre-} & 30.84 & -0.55 & 32.66 & -2.41 & 27.39 & -0.58 & 27.44 & -0.21 \\
    \cline{2-2}
     & {in-} & 29.91 & -1.48 & 32.42 & -2.66 & 27.29 & -0.68 & 27.52 & -0.13 \\
    \cline{2-10}
     & {post-} & {\bf 31.39} & - & {\bf 35.08} & - & {\bf 27.97} & - & {\bf 27.66} & - \\
    \hline
    \end{tabular}
    \egroup
    \caption{
    {\bf Effectiveness of the post-activation.}
    Geometry modeling with density voxel grid can achieve better PSNRs by using the proposed post-activated trilinear interpolation.
    }
    \label{tab:interp}
    \vspace{-.5em}
\end{table}

\begin{table}[tbp]
    \centering
    \begin{subtable}[t]{\linewidth}
    \centering
    \bgroup
    \setlength{\tabcolsep}{3pt}
    \footnotesize
    \begin{tabular}{@{}c@{}c|c@{\hskip 2pt}c|c@{\hskip 2pt}c|c@{\hskip 2pt}c|c@{\hskip 2pt}c@{}}
    \hline
    \multirow{2}{*}{\scriptsize $\alpha^{\text{(init)(c)}}$} & \multirow{2}{*}{\scriptsize \makecell{View.\\lr.}} &
    \multicolumn{2}{c|}{\it Syn.-NeRF} &
    \multicolumn{2}{c|}{\it Syn.-NSVF} &
    \multicolumn{2}{c|}{\it BlendedMVS} &
    \multicolumn{2}{c}{\it T\&T} \\
    & & {PSNR$\uparrow$} & {$\Delta$} & {PSNR$\uparrow$} & {$\Delta$} & {PSNR$\uparrow$} & {$\Delta$} & {PSNR$\uparrow$} & {$\Delta$} \\
    \hline\hline
    -         & \checkmark & 28.88 & -2.51 & 25.12 & -9.96 & 22.17 & -5.79 & 25.33 & -2.33 \\
    $10^{-3}$ & \checkmark & 30.96 & -0.42 & 27.24 & -7.84 & 23.17 & -4.79 & 26.04 & -1.61 \\
    $10^{-4}$ & \checkmark & 31.29 & -0.09 & 31.05 & -4.03 & 26.09 & -1.88 & 27.60 & -0.05 \\
    $10^{-5}$ & \checkmark & 31.41 & +0.02  & 35.04 & -0.04 & 27.36 & -0.61 & 27.63 & -0.02 \\
    $10^{-6}$ &            & 31.40 & +0.01  & 35.03 & -0.04 & 27.37 & -0.60 & 27.59 & -0.07 \\
    $10^{-7}$ & \checkmark & 31.36 & -0.02 & 35.03 & -0.05 & 27.73 & -0.23 & 27.59 & -0.06 \\
    \hline
    $10^{-6}$ & \checkmark & {\bf 31.39} & - & {\bf 35.08} & - & {\bf 27.97} & - & {\bf 27.66} & - \\
    \hline
    \end{tabular}
    \egroup
    \end{subtable}
    \par\smallskip
    \begin{subtable}[t]{\linewidth}
    \centering
    \begin{tabular}{@{}c@{\hskip 1.5pt}c@{\hskip 1.5pt}c@{\hskip 1.5pt}c@{}}
        \includegraphics[width=.16\linewidth]{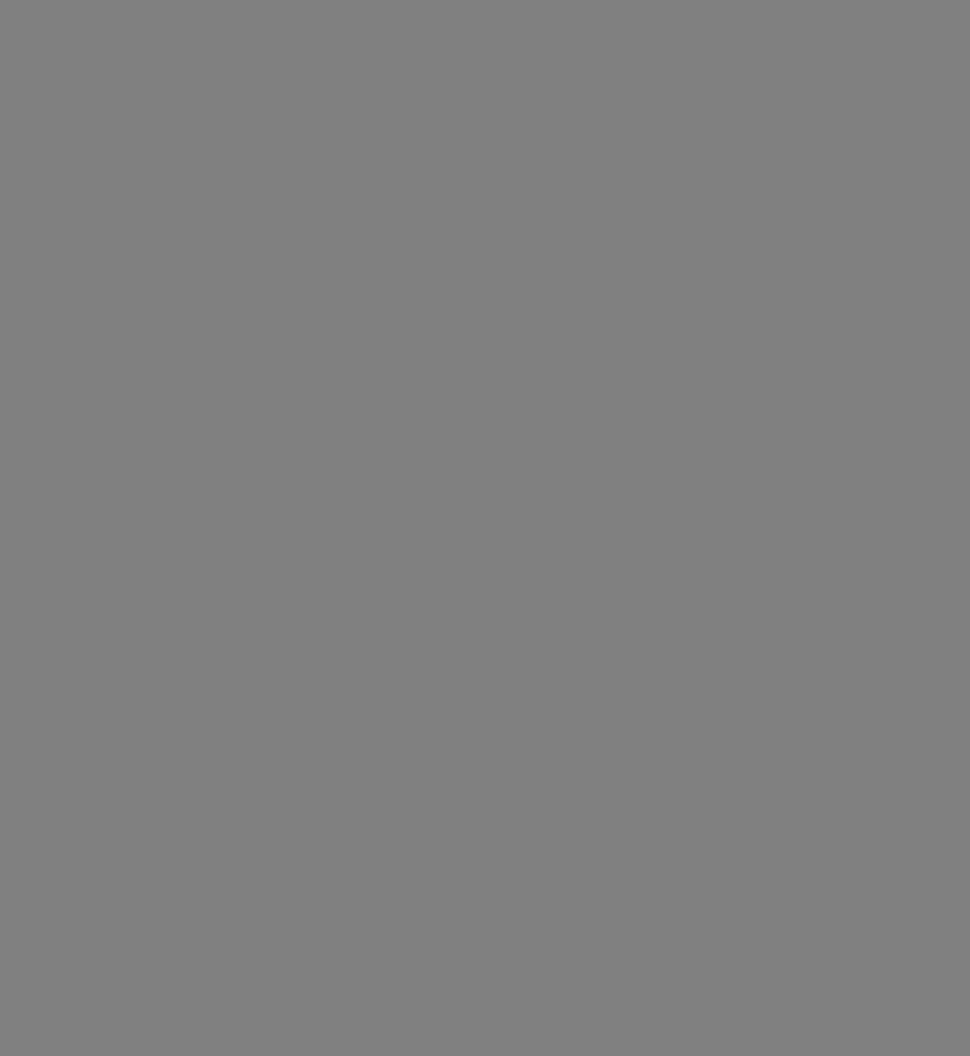} &
        \includegraphics[width=.16\linewidth]{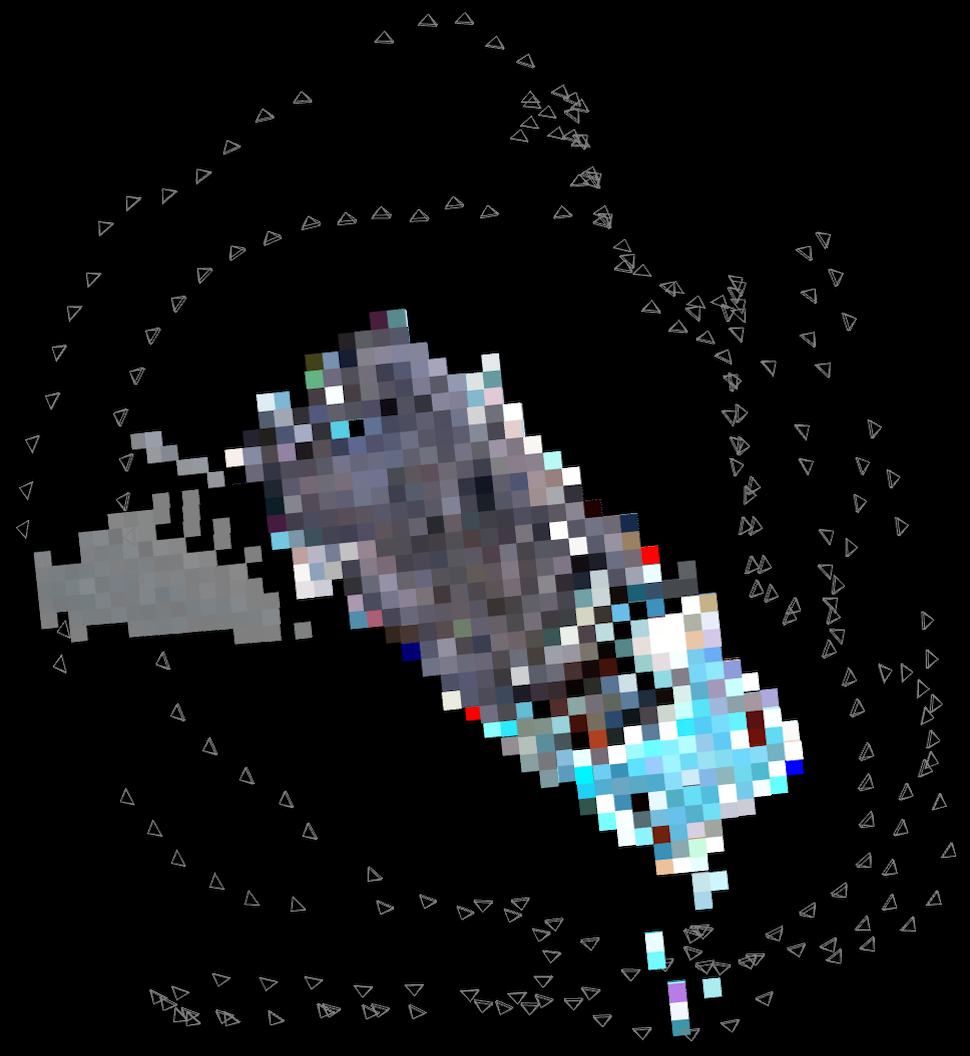} &
        \includegraphics[width=.16\linewidth]{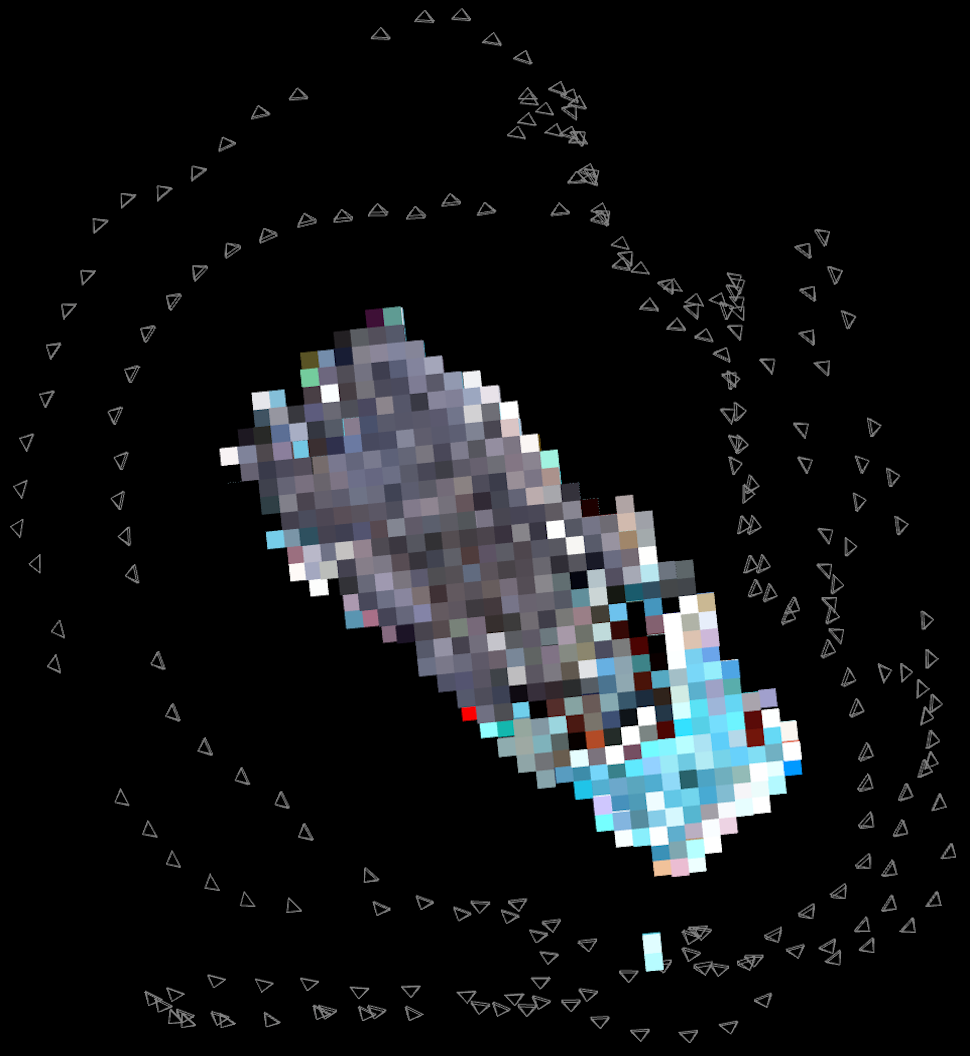} &
        \includegraphics[width=.16\linewidth]{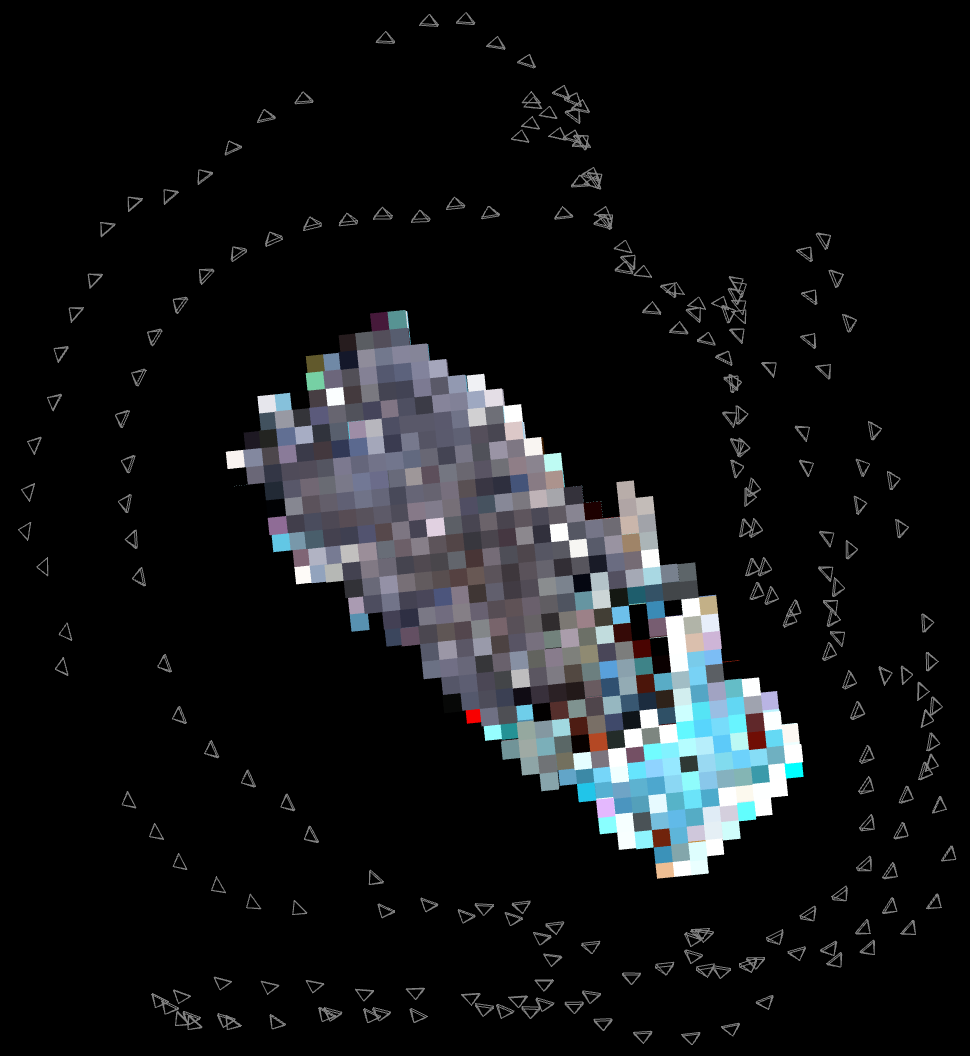} \\
        {\scriptsize - ~/~ \checkmark} &
        {\scriptsize $10^{-3}$ ~/~ \checkmark} &
        {\scriptsize $10^{-6}$ ~/~ -} &
        {\scriptsize $10^{-6}$ ~/~ \checkmark} \\
    \end{tabular}
    \end{subtable}
    \caption{
    {\bf Effectiveness of the imposed priors.}
    We compare our different settings in the coarse geometry search.
    {\it Top}: We show their impacts on the final PSNRs after the fine stage reconstruction.
    {\it Bottom}: We visualize the allocated voxels by coarse geometry search on the {\it Truck} scene.
    Overall, low-density initialization is essential; using $\alpha^{\text{(init)(c)}}=10^{-6}$ and view-count-based learning rate generally achieves cleaner voxels allocation in the coarse stage and better PSNR after the fine stage.
    }
    \label{tab:priors}
    \vspace{-1em}
\end{table}


\section{Conclusion}
Our method directly optimizes the voxel grid and achieves super-fast convergence in per-scene optimization with NeRF-comparable quality---reducing training time from many hours to 15 minutes.
However, we do not deal with the unbounded or forward-facing scenes, while we believe our method can be a stepping stone toward fast convergence in such scenarios.
We hope our method can boost the progress of NeRF-based scene reconstruction and its applications.

\noindent \textbf{Acknowledgements:}
This work was supported in part by the MOST grants 110-2634-F-001-009 and 110-2622-8-007-010-TE2 of Taiwan. We are grateful to National Center for High-performance Computing for providing computational resources and facilities.


\noindent\textbf{\Large Supplementary material}

\renewcommand\thesection{\Alph{section}}
\setcounter{table}{0}
\renewcommand\thetable{\Alph{section}\arabic{table}}
\setcounter{figure}{0}
\renewcommand\thefigure{\Alph{section}\arabic{figure}}

\vspace{1em}
In \cref{sec:supp_impl_detail}, we give more implementation details.
In \cref{sec:supp_more_abla}, we present additional ablation studies for the important hyperparameters that affect the quality and convergence speed.
In \cref{sec:supp_main_abla_detail}, we show detailed results of our main ablation studies presented in the main paper.
In \cref{sec:supp_train_time_detail}, we report our training time breakdown of the coarse and fine stage on each scene.
In \cref{sec:supp_per_scene_analysis}, we show the detailed quantitative and qualitative results on each scene.
Finally, we derive for the low-density initialization in \cref{sec:supp_derive_low_dense}, and we prove that the proposed post-activated trilinear interpolation can produce sharp linear surfaces in \cref{sec:supp_derive_post_act}.

\section{Additional implementation details} \label{sec:supp_impl_detail}
We choose the same hyperparameters generally for all scenes in the five datasets.

In the coarse stage, the expected number of voxels is  $M^{\text{(c)}}=100^3$.
The activated alpha values are initialized to be $\alpha^{\text{(init)(c)}}=10^{-6}$.

In the fine stage, we use $M^{\text{(f)}}=160^3$ voxels.
We use a higher $\alpha^{\text{(init)(f)}}=0.01$ as the query points in the known free space are skipped.
The shallow MLP layer comprises two hidden layers with $128$ channels.
The number of channels of the feature voxel grid $\bm{V}^{\text{(feat)(f)}}$ is set to $D=12$.
The number of frequencies of positional embedding for the query position $\bm{x}$ and the viewing direction $\bm{d}$ are $k_{\bm{x}}=5, k_{\bm{d}}=4$.
We progressively scale the fine-stage voxel grid at the checkpoint $\mathrm{pg\_ckpt}$ being chosen as the $1000$th, $2000$th, and $3000$th training steps.
We use threshold $\tau^{\text{(c)}}=10^{-3}$ to define the known free space and $\tau^{\text{(f)}}=10^{-4}$ to skip the low-density query points in unknown space.

We set the sampling step sizes to be half of the voxel sizes, \ie, $\delta^{\text{(c)}} = 0.5s^{\text{(c)}}$ and $\delta^{\text{(f)}} = 0.5s^{\text{(f)}}$.
During training, we randomly shift the query points on a ray by $\left(p \cdot \delta^{(\cdot)} \cdot \bm{d} / \|\bm{d}\|^2\right)$, where $p$ is sampled uniformly in $[0,1]$, and $\bm{d}$ is the viewing direction of the ray.

The loss weights are set as follows: $w_{\mathrm{photo}}^{\text{(c)}}=1$, $w_{\mathrm{pt\_rgb}}^{\text{(c)}}=10^{-1}$, $w_{\mathrm{bg}}^{\text{(c)}}=10^{-2}$, $w_{\mathrm{photo}}^{\text{(f)}}=1$, $w_{\mathrm{pt\_rgb}}^{\text{(f)}}=10^{-2}$, and $w_{\mathrm{bg}}^{\text{(f)}}=10^{-3}$.
We use the Adam optimizer with a batch size of $8{,}192$ rays to optimize the coarse and fine scene representations for $10$k and $20$k iterations.
The base learning rates are $0.1$ for all voxel grids and $10^{-3}$ for the MLP.
The exponential learning rate decay is applied such that the learning rates are downscaled by $0.1$ at $20$k iterations.

\setcounter{table}{0}
\setcounter{figure}{0}
\section{Additional ablation experiments} \label{sec:supp_more_abla}
We use the {\it Robot} scene to conduct the additional ablation experiments.
In all the additional ablation experiments, the fine stage are early stopped at $10$k iterations.
The reported training times are measured on our machine with a single RTX2080 Ti GPU.

\paragraph{Effect of the number of voxels.}
The hyperparameters $M^{\text{(c)}}$ and $M^{\text{(f)}}$ define the numbers of voxels in the coarse stage and the fine stage.
In \cref{tab:supp_num_vox}, we show the effect of different $M^{(\cdot)}$ setups on the final PSNRs and the training times.
We can use more voxels for better quality or use fewer voxels for faster convergence speed.
\begin{table}[htpb]
    \centering
    \begin{subtable}[t]{\linewidth}
        \centering
        \begin{tabular}{cc|cc}
        \hline
        $M^{\text{(c)}}$ & $M^{\text{(f)}}$ & PSNR$\uparrow$ & Training minutes$\downarrow$ \\
        \hline\hline
        $100^3$ & $100^3$ & 32.65 & {\bf 5.1} \\
        $100^3$ & $136^3$ & 34.57 & 7.0 \\
        $100^3$ & $160^3$ & {\bf 35.54} & 7.3 \\
        \hline
        \end{tabular}
        \caption{\footnotesize
        Using fewer voxels in the fine stage results in inferior qualities but faster training speed.
        }
    \end{subtable}
    \par\medskip
    \begin{subtable}[t]{\linewidth}
        \centering
        \begin{tabular}{cc|cc}
        \hline
        $M^{\text{(c)}}$ & $M^{\text{(f)}}$ & PSNR$\uparrow$ & Training minutes$\downarrow$ \\
        \hline\hline
        $64^3$ & $160^3$ & 34.91 & {\bf 6.1} \\
        $100^3$ & $160^3$ & 35.54 & 7.3 \\
        $136^3$ & $160^3$ & {\bf 35.84} & 9.3 \\
        \hline
        \end{tabular}
        \caption{\footnotesize
        The number of voxels in the coarse stage can also affect final results as the fine stage relies on the coarse geometry optimized in the coarse stage.
        }
        \end{subtable}
    \caption{
    Effect of the number of voxels in the coarse stage ($M^{\text{(c)}}$) and the fine stage ($M^{\text{(f)}}$).
    Using more voxels can improve quality with the cost of more computation and training time.
    }
    \label{tab:supp_num_vox}
    \vspace{-1em}
\end{table}

\paragraph{Effect of the point sampling step size.}
In our design, the voxel sizes $s^{(\cdot)}$ are automatically derived from the hyperparameters of the number of voxels $M^{(\cdot)}$.
We set the step-size-to-voxel-size ratio instead of directly assigning the step size for points sampling on rays, \ie, $\delta^{(\cdot)}=0.5s^{(\cdot)}$ means that the step size is half a voxel size.
We show in \cref{tab:supp_stepsize} that the step sizes are also important hyperparameters for the speed-quality tradeoff.
We can also improve the rendering quality of a trained model by using a finer step size.
For the sake of simplicity, we leave future work to explore the effect of different hierarchical point sampling strategies~\cite{MildenhallSTBRN20,BarronMTHMS21,OechslePG21}.

\begin{table}[h]
    \centering
    \begin{subtable}{\linewidth}
        \centering
        \begin{tabular}{@{}cc|cc@{}}
        \hline
        \multicolumn{2}{c|}{Step size} & \multirow{2}{*}{PSNR$\uparrow$} & \multirow{2}{*}{sec/frame$\downarrow$} \\
        Training & Testing & & \\
        \hline\hline
        $\delta^{(\cdot)}=0.50s^{(\cdot)}$ & $\delta^{(\cdot)}=0.50s^{(\cdot)}$ & 35.54 & {\bf 0.64} \\
        $\delta^{(\cdot)}=0.50s^{(\cdot)}$ & $\delta^{(\cdot)}=0.25s^{(\cdot)}$ & {\bf 35.72} & 1.15 \\
        \hline
        \end{tabular}
        \caption{\footnotesize
        We can improve quality by using a finer step size in test-time with the cost of slower rendering speed.
        The two results share a trained model and are only different in the testing step size.
        }
    \end{subtable}
    \par\medskip
    \begin{subtable}{\linewidth}
        \centering
        \begin{tabular}{c|cc}
        \hline
        Step size & PSNR$\uparrow$ & Training minutes$\downarrow$ \\
        \hline\hline
        $\delta^{(\cdot)}=1.00s^{(\cdot)}$ & 34.17 & {\bf 5.2} \\
        $\delta^{(\cdot)}=0.75s^{(\cdot)}$ & 34.96 & 5.9 \\
        $\delta^{(\cdot)}=0.50s^{(\cdot)}$ & 35.54 & 7.3 \\
        $\delta^{(\cdot)}=0.25s^{(\cdot)}$ & {\bf 36.01} & 11.5 \\
        \hline
        \end{tabular}
        \caption{\footnotesize
        Training with finer step size can improve results with the cost of more computation and training time.
        }
    \end{subtable}
    \caption{
    Effect of the step size in points sampling.
    The step size $\delta^{(\cdot)}$ is relative to the voxel size $s^{(\cdot)}$, and we apply the same step-size-to-voxel-size ratio in our coarse and fine stage.
    We can achieve better quality by using a smaller (finer) step size with the cost of more computation.
    }
    \label{tab:supp_stepsize}
\end{table}

\paragraph{Effect of the progressive scaling in the fine stage.}
In the fine-stage training, we scale the resolution of our voxel grids progressively.
We show in \cref{tab:supp_pg} that such a strategy can improve the training efficiency with a slightly better rendering quality.
\begin{table}[htpb]
    \centering
    \begin{tabular}{c|cc}
    \hline
    Progressive scaling & PSNR$\uparrow$ & Training minutes$\downarrow$ \\
    \hline\hline
    & 35.50 & 11.7 \\
    \checkmark & {\bf 35.54} & {\bf 7.3} \\
    \hline
    \end{tabular}
    \caption{Scaling the grid resolutions progressively in the fine stage can improve training efficiency with a slightly better quality.}
    \label{tab:supp_pg}
\end{table}

\paragraph{Effect of the free space skipping in the fine stage.}
There are three models in the fine stage---{\it i)} the optimized and frozen coarse density voxel grid $\bm{V}^{\text{(density)(c)}}$, {\it ii)} the finer density voxel grid $\bm{V}^{\text{(density)(f)}}$, and {\it iii)} the model for the view-dependent color emission.
Querying the optimized $\bm{V}^{\text{(density)(c)}}$ is more efficient than querying the finer density $\bm{V}^{\text{(density)(f)}}$; querying the view-dependent color emission is the slowest and computationally expensive.
For each query point, we first query $\bm{V}^{\text{(density)(c)}}$ and ignore the query point if the activated alpha is below a threshold $\tau^{\text{(c)}}$ (\ie, the point is in the known free space).
As the training progresses, we can filter out more points that are in the free space defined by the ``sculpted" $\bm{V}^{\text{(density)(f)}}$.
Specifically, we ignore a query point if the activated alpha from $\bm{V}^{\text{(density)(f)}}$ is below a threshold $\tau^{\text{(f)}}$.
Thus, we only query for the view-dependent color if the query point passes the two criteria.

In \cref{tab:supp_skip}, we show the effect of the free space skipping strategy.
In the case of $\tau^{\text{(c)}}=0$, the coarse voxel grid $\bm{V}^{\text{(density)(c)}}$ is only used to determine a tighten BBox enclosing the scene.
We finally run out of memory (OOM) at the fine stage if no skipping rule is applied ($\tau^{\text{(c)}}=0, \tau^{\text{(f)}}=0$).
When only $\tau^{\text{(f)}}$ is applied, we can still sculpt out many irrelevant query points during the progressive scaling and save the memory before our voxel grids are scaled to the finest resolution.
The rendering quality is degraded when $\tau^{\text{(c)}}=0$, which suggests that the imposed priors in the coarse stage are helpful to the final quality (we cannot apply the low-density initialization and the free space skipping at the same time).
Finally, we achieve the best rendering quality and convergence speed when both $\tau^{\text{(c)}}$ and $\tau^{\text{(f)}}$ are applied.
\begin{table}[htpb]
    \centering
    \begin{tabular}{cc|cc}
    \hline
    \multicolumn{2}{c|}{Free space skipping} & \multirow{2}{*}{PSNR$\uparrow$} & \multirow{2}{*}{Training minutes$\downarrow$} \\
    $\tau^{\text{(c)}}$ & $\tau^{\text{(f)}}$ & & \\
    \hline\hline
    $0$ & $0$ & - & OOM \\
    $0$ & $10^{-4}$ & 34.89 & 7.8 \\
    $10^{-3}$ & $0$ & {\bf 35.54} & 15.6 \\
    $10^{-3}$ & $10^{-4}$ & {\bf 35.54} & {\bf 7.3} \\
    \hline
    \end{tabular}
    \caption{Effect of the free space skipping strategy in the fine stage training.}
    \label{tab:supp_skip}
\end{table}

\paragraph{Ablation studies for the view-dependent color modeling.}
Our hybrid representation for the view-dependent color emissions in the fine stage comprises a feature voxel grid and a shallow MLP. 
In \cref{tab:supp_rgb}, we show different strategies and hyperparameters to model view-dependent colors.
Applying only the shallow MLP results in worse outputs as our MLP is much ``shallower" than the MLP in NeRF---our two layers with 128 channels versus NeRF's eight layers with 256 channels.
Using only the voxel grid to model view-invariant diffused colors can converge in much less time, but the rendering quality is degraded.
In NeRF-based scene reconstruction, spherical harmonic representation~\cite{YuLTLNK2021,LiuPLWWTZW21}, learnable basis~\cite{WizadwongsaPYS21}, or deferred rendering~\cite{HedmanSMBD2021} are also shown to be effective, but we leave the explorations for future work.
We also compare the number of dimensions $D$ of our feature grid, the number of layers, and the number of channels of the shallow MLP.
Using a larger model leads to better rendering quality with the cost of more computation.
\begin{table}[htpb]
    \centering
    \begin{tabular}{ccc|cc}
    \hline
    \multicolumn{3}{l|}{Colors rep. in fine stage} & PSNR$\uparrow$ & Training minutes$\downarrow$ \\
    \hline\hline
    \multicolumn{3}{l|}{implicit {\footnotesize (shallow MLP)}} & 26.57 & 6.2 \\
    \multicolumn{3}{l|}{explicit {\footnotesize (diffused only)}} & 32.15 & {\bf 2.9} \\
    \cline{1-3}
    \multicolumn{3}{l|}{hybrid} &  \\
    $D$ & {\scriptsize MLP layers} & {\scriptsize MLP ch.} & & \\
    \hline
    12 & 1 & 128 & 35.11 & 7.0 \\
    12 & 2 & 128 & 35.54 & 7.3 \\
    12 & 3 & 128 & 35.70 & 7.8 \\
    12 & 2 &  64 & 35.35 & 7.0 \\
    12 & 2 & 192 & 35.71 & 7.7 \\
    24 & 2 & 128 & {\bf 35.90} & 8.8 \\
    \hline
    \end{tabular}
    \caption{Different representations and hyperparameters for modeling the view-dependent color emissions.}
    \label{tab:supp_rgb}
\end{table}

\paragraph{Effect of the auxiliary losses.}
We show in \cref{tab:supp_weight} that incorporating the per-point rgb loss and the background entropy loss added to the main photometric loss can improve our results.
The per-point rgb loss supervises the color emission of each query points directly instead of the accumulated color.
The background entropy loss enforces the rendered background probability to concentrate on the background or foreground.
We achieve the best results when adding the two auxiliary losses.
\begin{table}[htpb]
    \centering
    \begin{tabular}{cc|c}
    \hline
    Per-point rgb loss & Background entropy loss & PSNR$\uparrow$ \\
    \hline\hline
     & & 34.84 \\
    \checkmark & & 35.37 \\
     & \checkmark & 35.53 \\
    \checkmark & \checkmark & {\bf 35.54} \\
    \hline
    \end{tabular}
    \caption{Effect of the auxiliary losses.}
    \label{tab:supp_weight}
\end{table}

\setcounter{table}{0}
\setcounter{figure}{0}
\section{Main ablation studies details} \label{sec:supp_main_abla_detail}
In \cref{tab:supp_main_abla_detail}, we show detailed results for the ablation experiments presented in the main paper.
We subsample two scenes for each dataset to conduct the main ablation experiments---{\it Materials} and {\it Mic} from Synthetic-NeRF dataset; {\it Robot} and {\it Lifestyle} from Synthetic-NSVF dataset; {\it Character} and {\it Statues} from BlendedMVS dataset, and {\it Ignatius} and {\it Truck} from Tanks and Temples dataset.

We show that the proposed post-activation significantly improves our quality.
The low-density initialization is shown to be essential to our method and is an important hyperparameter.
The view-count-based learning rate can slightly improve the results, and the PSNR without the view-count prior can degrade up to $-1.22$ in the worst case.

\begin{table*}[htb]
    \centering
    \bgroup
    \setlength{\tabcolsep}{1.25pt}
    \scriptsize
    \begin{tabular}{lcc|cccc||cccc|cccc|cccc|cccc}
    \hline
    \multirow{3}{*}{Interp.} & \multirow{3}{*}{$\alpha^{\text{(init)(c)}}$} & \multirow{3}{*}{\makecell{View.\\lr.}} & \multicolumn{4}{c||}{\multirow{2}{*}{Overall}} & \multicolumn{4}{c|}{Synthetic-NeRF} & \multicolumn{4}{c|}{Synthetic-NSVF} & \multicolumn{4}{c|}{BlendedMVS} & \multicolumn{4}{c}{Tanks and Temples} \\
    & & & & & & & \multicolumn{2}{c}{\it Materials} & \multicolumn{2}{c|}{\it Mic} & \multicolumn{2}{c}{\it Robot} & \multicolumn{2}{c|}{\it Lifestyle} & \multicolumn{2}{c}{\it Character} & \multicolumn{2}{c|}{\it Statues} & \multicolumn{2}{c}{\it Ignatius} & \multicolumn{2}{c}{\it Truck} \\
    & & & PSNR$\uparrow$ & $\Delta_{\text{avg}}$ & $\Delta_{\text{worst}}$ & $\Delta_{\text{best}}$ & PSNR$\uparrow$ & $\Delta$ & PSNR$\uparrow$ & $\Delta$ & PSNR$\uparrow$ & $\Delta$ & PSNR$\uparrow$ & $\Delta$ & PSNR$\uparrow$ & $\Delta$ & PSNR$\uparrow$ & $\Delta$ & PSNR$\uparrow$ & $\Delta$ & PSNR$\uparrow$ & $\Delta$ \\
    \hline\hline
post-act.                  & $10^{-6}$ & \checkmark & \gold{30.52} &  -    &  -    &  -    & \silver{29.57} & - & 33.20 & - & \silver{36.36} & - & \gold{33.79} & - & 30.31 & - & \gold{25.62} & - & \brown{28.16} & - & \gold{27.15} \\
\hline
nearest                    & $10^{-6}$ & \checkmark & 27.33  & -3.19 & -7.50 & -0.34 & 28.17 & -1.40 & 29.05 & -4.15 & 28.86 & -7.50 & 28.85 & -4.94 & 27.28 & -3.03 & 23.69 & -1.93 & 27.82 & -0.34 & 24.95 & -2.20 \\
pre-act.                   & $10^{-6}$ & \checkmark & 29.58  & -0.94 & -2.88 & -0.14 & 29.24 & -0.33 & 32.43 & -0.77 & 33.48 & -2.88 & 31.85 & -1.94 & 29.73 & -0.58 & \brown{25.04} & -0.58 & 28.02 & -0.14 & 26.86 & -0.29 \\
in-act.                    & $10^{-6}$ & \checkmark & 29.28  & -1.24 & -3.30 &  0.00 & 27.34 & -2.23 & 32.47 & -0.73 & 33.06 & -3.30 & 31.77 & -2.02 & 29.74 & -0.57 & 24.84 & -0.78 & 28.16 &  0.00 & 26.89 & -0.26 \\
\hline
\multirow{6}{*}{post-act.} & -         & \checkmark & 25.37  & -5.15 & -9.98 & -1.61 & 27.40 & -2.17 & 30.35 & -2.85 & 26.43 & -9.93 & 23.81 & -9.98 & 24.70 & -5.61 & 19.65 & -5.97 & 26.55 & -1.61 & 24.10 & -3.05 \\
                           & $10^{-3}$ & \checkmark & 26.85  & -3.67 & -8.40 & -0.23 & 28.96 & -0.61 & 32.97 & -0.23 & 29.09 & -7.27 & 25.39 & -8.40 & 25.12 & -5.19 & 21.22 & -4.40 & 27.23 & -0.93 & 24.86 & -2.29 \\
                           & $10^{-4}$ & \checkmark & 29.01  & -1.51 & -4.30 & +0.05 & 29.35 & -0.22 & \gold{33.23} & +0.03 & 32.06 & -4.30 & 30.04 & -3.75 & 28.47 & -1.84 & 23.70 & -1.92 & \gold{28.21} & +0.05 & 26.99 & -0.16 \\
                           & $10^{-5}$ & \checkmark & \brown{30.36} & -0.16 & -1.22 & +0.03 & \gold{29.60} & +0.03 & \silver{33.22} & +0.02 & 36.32 & -0.04 & \silver{33.75} & -0.04 & \brown{30.32} & +0.01 & 24.40 & -1.22 & \silver{28.17} & +0.01 & \brown{27.10} & -0.05 \\
                           & $10^{-6}$ &            & 30.35  & -0.17 & -1.22 & +0.02 & \silver{29.57} &  0.00 & \silver{33.22} & +0.02 & \brown{36.33} & -0.03 & \brown{33.74} & -0.05 & \silver{30.33} & +0.02 & 24.40 & -1.22 & 28.06 & -0.10 & \silver{27.12} & -0.03 \\
                           & $10^{-7}$ & \checkmark & \silver{30.43} & -0.09 & -0.56 & +0.20 & \silver{29.57} &  0.00 & 33.15 & -0.05 & \gold{36.56} & +0.20 & 33.50 & -0.29 & \gold{30.40} & +0.09 & \silver{25.06} & -0.56 & 28.12 & -0.04 & 27.07 & -0.08 \\
    \hline
    \end{tabular}
    \egroup
    \caption{
        We detail the per-scene results of different ablation setups presented in the main paper.
        We also show their difference ($\Delta$) to our final setting (the first row).
        The $\Delta_{\text{worst}}$ highlights the worst-case degradation overall scenes, where most of the settings lead to more than 1 dB degradation in the worst case.
        The $\Delta_{\text{best}}$ shows the best case difference, where some settings can slightly improve the results on some scenes.
        We highlight the top 3 results of each column in \gold{gold}, \silver{silver}, and \brown{bronze}.
    }
    \label{tab:supp_main_abla_detail}
\end{table*}

\setcounter{table}{0}
\setcounter{figure}{0}
\section{Additional training time details} \label{sec:supp_train_time_detail}
\subsection{Training time comparisons}
Mip-NeRF~\cite{BarronMTHMS21} requires similar run-time as NeRF~\cite{MildenhallSTBRN20} but achieves much better quality.
As a side benefit, Mip-NeRF can attain NeRF's PSNR in less time.
To compare the convergence speed with Mip-NeRF, we use the code open-sourced by the authors (\url{https://github.com/google/mipnerf}).
We re-train early-stopped Mip-NeRFs on our machine with only a single RTX2080 Ti GPU.
The comparison is shown in \cref{tab:supp_training_time}.
The early-stopped Mip-NeRF that is trained for 6 hours achieves an average PSNR of $30.85$.
On the other hand, our method is only optimized for about $15$ minutes and achieves an average PSNR of $31.95$.

\begin{table*}[htp]
    \centering
    \begin{tabular}{c@{\hskip 4pt}c|c|@{\hskip 8pt}c@{\hskip 8pt}c@{\hskip 8pt}c@{\hskip 8pt}c@{\hskip 8pt}c@{\hskip 8pt}c@{\hskip 8pt}c@{\hskip 8pt}c}
    \hline
    \multicolumn{2}{c|}{Methods} & Avg. & {\it Chair} & {\it Drums} & {\it Ficus} & {\it Hotdog} & {\it Lego} & {\it Materials} & {\it Mic} & {\it Ship} \\
    \hline\hline
    \multirow{2}{*}{Mip-NeRF~\cite{BarronMTHMS21} {\scriptsize (early-stopped)}} & PSNR$\uparrow$ & 30.85 & 32.25 & 24.95 & 30.66 & 35.56 & 32.92 & 29.28 & 32.61 & 28.56 \\
     & {\footnotesize training minutes$\downarrow$} &  361.2 & 361.1 & 363.0 & 360.1 & 359.0 & 363.7 & 360.7 & 359.3 & 363.0 \\
    \hline
    \multirow{2}{*}{Ours} & PSNR$\uparrow$ & {\bf 31.95} & {\bf 34.09} & {\bf 25.44} & {\bf 32.78} & {\bf 36.74} & {\bf 34.64} & {\bf 29.57} & {\bf 33.20} & {\bf 29.13} \\
     & {\footnotesize training minutes$\downarrow$} & {\bf 14.2} & {\bf 12.5} & {\bf 12.0} & {\bf 13.8} & {\bf 15.5} & {\bf 13.2} & {\bf 15.4} & {\bf 11.0} & {\bf 20.2} \\
    \hline
    \end{tabular}
    \caption{
    We compare the training time with the early-stopped Mip-NeRF, which is a stronger baseline than NeRF.
    The training time is measured on our machine with only a single RTX2080 Ti GPU.
    Our method with about 15 minutes of training time outperforms the early-stopped Mip-NeRF with 6 hours of training by a large margin.
    }
    \label{tab:supp_training_time}
\end{table*}

\subsection{The detailed training time of our model}
We report our training times in detail on each scene and each stage in \cref{tab:supp_our_traintime}.
The coarse stage optimization accounts for about $15\mathrm{-}20\%$ of the overall training time.
The per-scene training times are about less than 15 minutes, except for the {\it Tanks\&Temples}~\cite{KnapitschPZK17} dataset, which takes just a few more minutes.
\begin{table*}[htpb]
    \centering
    \begin{subtable}{\linewidth}
        \centering
        \begin{tabular}{l|c|@{\hskip 8pt}c@{\hskip 8pt}c@{\hskip 8pt}c@{\hskip 8pt}c@{\hskip 8pt}c@{\hskip 8pt}c@{\hskip 8pt}c@{\hskip 8pt}c}
        \hline
        Stage & Avg. & {\it Chair} & {\it Drums} & {\it Ficus} & {\it Hotdog} & {\it Lego} & {\it Materials} & {\it Mic} & {\it Ship} \\
        \hline\hline
        Coarse & 2.3 & 2.5 & 2.4 & 2.0 & 2.3 & 2.3 & 2.1 & 2.3 & 2.3 \\
        Fine & 11.9 & 10.1 & 9.7 & 11.7 & 13.2 & 10.9 & 13.3 & 8.7 & 17.9 \\
        All & 14.2 & 12.5 & 12.0 & 13.8 & 15.5 & 13.2 & 15.4 & 11.0 & 20.2 \\
        \hline
        \end{tabular}
        \caption{\footnotesize
        Training times (minutes) on the eight scenes of {\bf Synthetic-NeRF}~\cite{MildenhallSTBRN20} dataset.
        }
    \end{subtable}
    \par\medskip
    \begin{subtable}{\linewidth}
        \centering
        \begin{tabular}{l|c|@{\hskip 8pt}c@{\hskip 8pt}c@{\hskip 8pt}c@{\hskip 8pt}c@{\hskip 8pt}c@{\hskip 8pt}c@{\hskip 8pt}c@{\hskip 8pt}c}
        \hline
        Stage & Avg. & {\it Wineholder} & {\it Steamtrain} & {\it Toad} & {\it Robot} & {\it Bike} & {\it Palace} & {\it Spaceship} & {\it Lifestyle} \\
        \hline\hline
        Coarse & 2.5 & 2.5 & 2.5 & 2.5 & 2.5 & 2.5 & 2.5 & 2.5 & 2.5 \\
        Fine & 10.7 & 9.6 & 12.8 & 9.8 & 9.5 & 10.1 & 11.2 & 12.1 & 10.1 \\
        All & 13.2 & 12.1 & 15.3 & 12.3 & 12.0 & 12.6 & 13.7 & 14.6 & 12.6 \\
        \hline
        \end{tabular}
        \caption{\footnotesize
        Training times (minutes) on the eight scenes of {\bf Synthetic-NSVF}~\cite{LiuGLCT20} dataset.
        }
    \end{subtable}
    \par\medskip
    \begin{subtable}{0.49\linewidth}
        \centering
        \begin{tabular}{l|c|@{\hskip 8pt}c@{\hskip 8pt}c@{\hskip 8pt}c@{\hskip 8pt}c}
        \hline
        Stage & Avg. & {\it Jade} & {\it Fountain} & {\it Character} & {\it Statues} \\
        \hline\hline
        Coarse & 2.6 & 2.0 & 2.7 & 3.2 & 2.3 \\
        Fine & 11.2 & 11.8 & 10.6 & 11.5 & 11.0 \\
        All & 13.8 & 13.8 & 13.3 & 14.7 & 13.2 \\
        \hline
        \end{tabular}
        \caption{\footnotesize
        Training times (minutes) on the four scenes of {\bf BlendedMVS}~\cite{YaoLLZRZFQ20} dataset.
        }
    \end{subtable}
    \hfill
    \begin{subtable}{0.50\linewidth}
        \centering
        \begin{tabular}{l|c|@{\hskip 8pt}c@{\hskip 8pt}c@{\hskip 8pt}c@{\hskip 8pt}c@{\hskip 8pt}c}
        \hline
        Stage & Avg. & {\it Ignatius} & {\it Truck} & {\it Barn} & {\it Cate.} & {\it Family} \\
        \hline\hline
        Coarse & 3.6 & 3.4 & 3.6 & 3.8 & 3.6 & 3.4 \\
        Fine & 14.2 & 12.4 & 14.3 & 18.2 & 14.1 & 11.9 \\
        All & 17.7 & 15.7 & 17.8 & 22.0 & 17.7 & 15.3 \\
        \hline
        \end{tabular}
        \caption{\footnotesize
        Training times (minutes) on the five scenes of {\bf Tanks\&Temples}~\cite{KnapitschPZK17} dataset.
        }
    \end{subtable}
    \par\medskip
    \begin{subtable}{\linewidth}
        \centering
        \begin{tabular}{l|c|@{\hskip 8pt}c@{\hskip 8pt}c@{\hskip 8pt}c@{\hskip 8pt}c}
        \hline
        Stage & Avg. & {\it Chair} & {\it Pedestal} & {\it Cube} & {\it Vase} \\
        \hline\hline
        Coarse & 2.4 & 2.5 & 2.5 & 2.1 & 2.5 \\
        Fine & 11.9 & 12.8 & 11.6 & 11.1 & 12.2 \\
        All & 14.3 & 15.2 & 14.1 & 13.2 & 14.7 \\
        \hline
        \end{tabular}
        \caption{\footnotesize
        Training times (minutes) on the four scenes of {\bf DeepVoxels}~\cite{SitzmannTHNWZ19} dataset.
        }
    \end{subtable}
    \caption{We report the detail of our per-scene optimization time in minutes measured on our machine with a single RTX2080 Ti GPU.}
    \label{tab:supp_our_traintime}
\end{table*}

\setcounter{table}{0}
\setcounter{figure}{0}
\section{Per-scene analysis} \label{sec:supp_per_scene_analysis}

\subsection{Per-scene quantitative results}
We report the per-scene quantitative comparisons in
\cref{tab:supp_breakdown_nerf} for {\bf Synthetic-NeRF}~\cite{MildenhallSTBRN20} dataset, \cref{tab:supp_breakdown_nsvf} for {\bf Synthetic-NSVF}~\cite{LiuGLCT20} dataset, \cref{tab:supp_breakdown_blendedmvs} for {\bf BlendedMVS}~\cite{YaoLLZRZFQ20} dataset, \cref{tab:supp_breakdown_tanksandtemples} for {\bf Tanks\&Temples}~\cite{KnapitschPZK17} dataset, and \cref{tab:supp_breakdown_deepvoxels} for {\bf DeepVoxels}~\cite{SitzmannTHNWZ19} dataset.
Our method achieves comparable results to most of the recent methods, except the Mip-NeRF~\cite{BarronMTHMS21} and JaxNeRF+~\cite{jaxnerf2020github}.
Moreover, all the methods after NeRF on the tables take quite a few hours to train for each scene, while our method only takes about 15 minutes per-scene optimization time as reported in \cref{tab:supp_our_traintime}.

\subsection{Per-scene qualitative results}
\begin{figure*}[b]
    \centering
    \includegraphics[width=\linewidth, trim=0 120 0 0, clip]{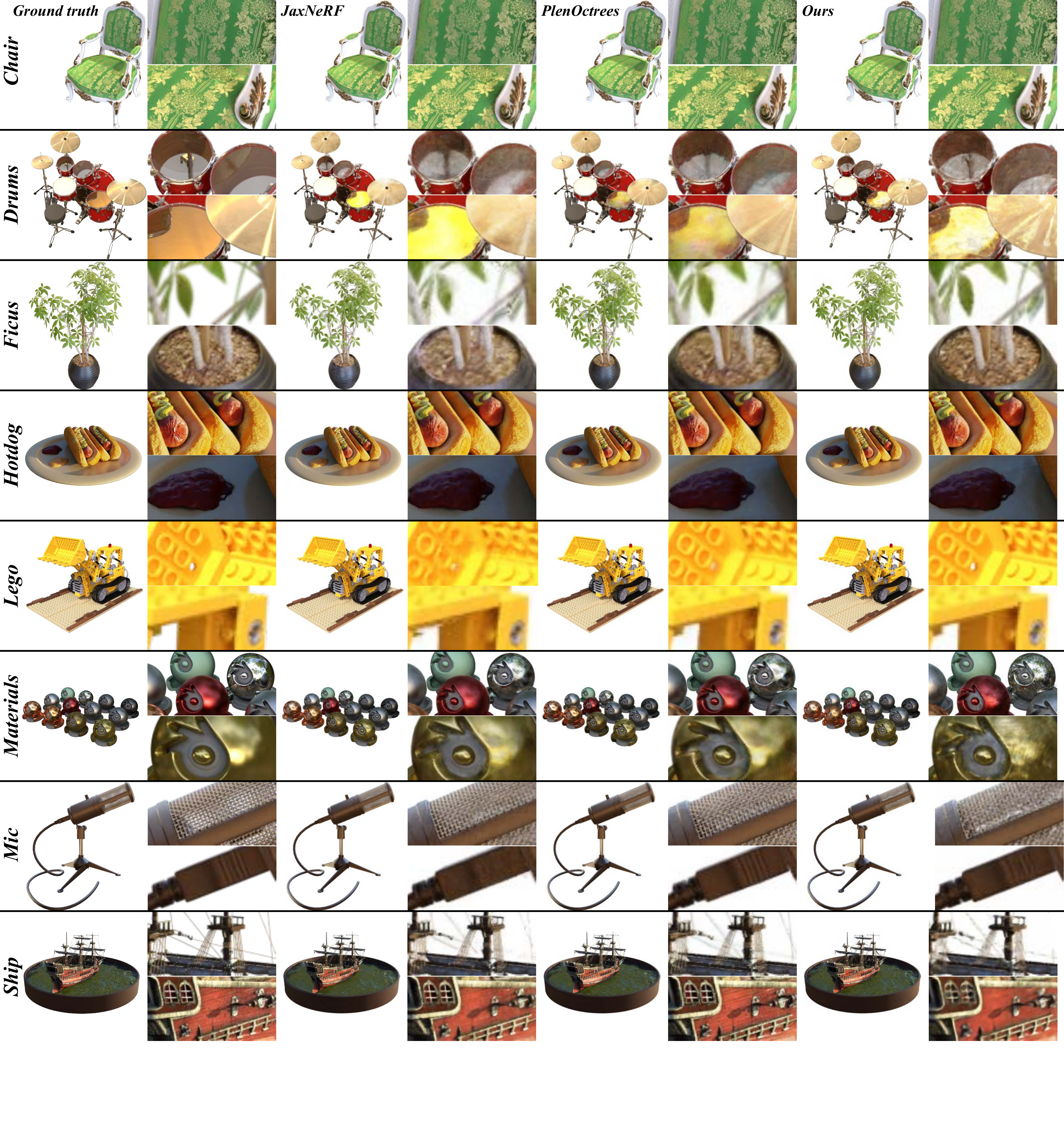}
    \caption{
    Qualitative comparisons on the on the {\bf Synthetic-NeRF}~\cite{MildenhallSTBRN20} dataset. 
    We manually resize, crop, and compress the images.
    JaxNeRF and PlenOctrees are both stronger baselines than NeRF.
    No method produces consistently better fine detail on all scenes.
    JaxNeRF recovers the shadow better in {\it Hotdog} and no blocking artifact in {\it Materials}; PlenOctrees shows better fine texture on {\it Mic} and {\it Ship}; our method reconstructs the fine detail of {\it Ficus} and {\it Lego} better.
    }
    \label{fig:our_on_nerfdataset}
\end{figure*}

\begin{figure*}[hbpt]
    \centering
    \includegraphics[width=\linewidth, trim=100 180 220 80, clip]{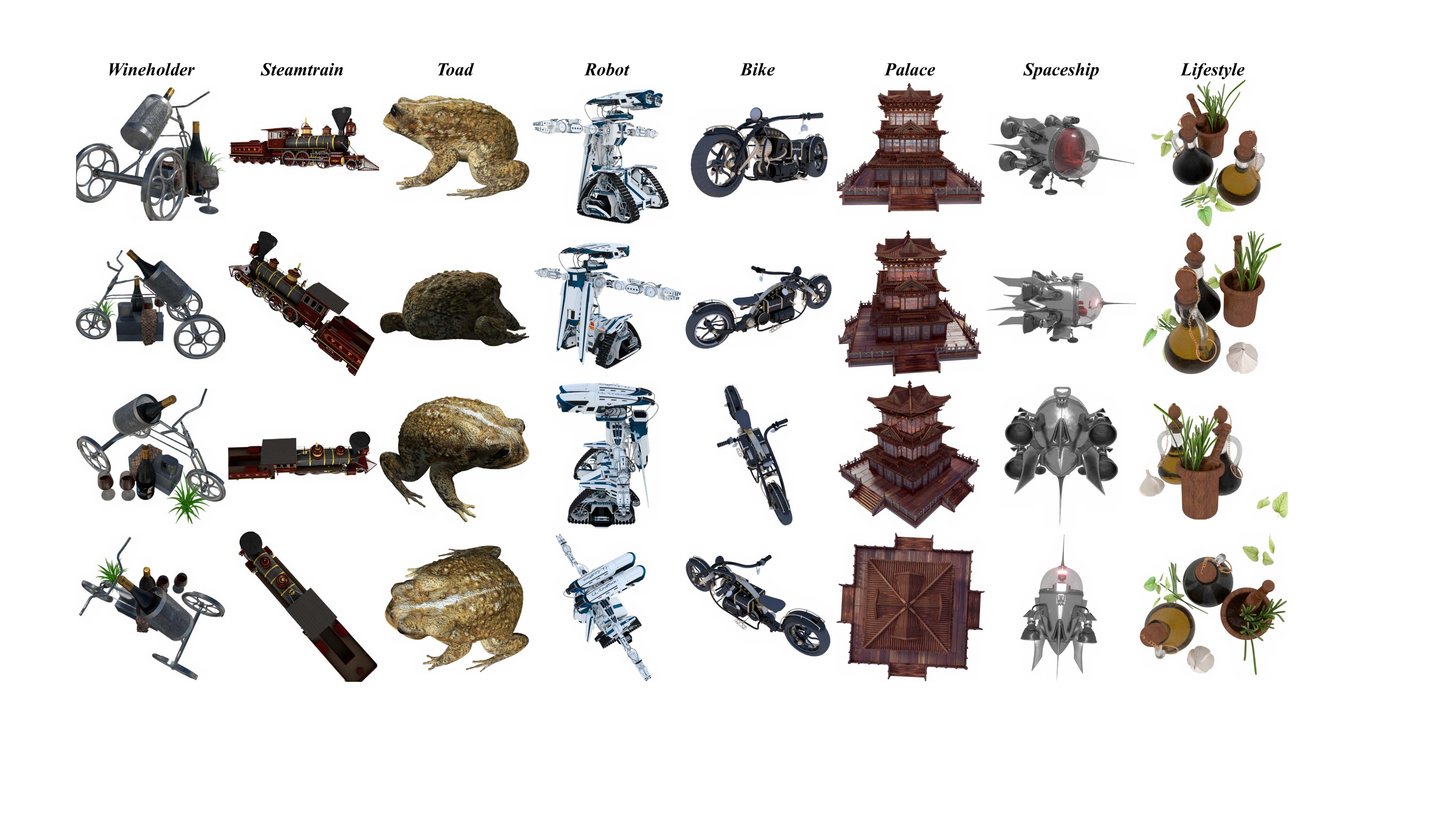}
    \caption{
    The synthesized novel views by our method on the {\bf Synthetic-NSVF}~\cite{LiuGLCT20} dataset.
    {\footnotesize We manually resize, crop, and compress the images.}
    }
    \label{fig:our_on_nsvfdataset}
\end{figure*}

\begin{figure*}[hbpt]
    \centering
    \includegraphics[width=\linewidth, trim=100 180 220 80, clip]{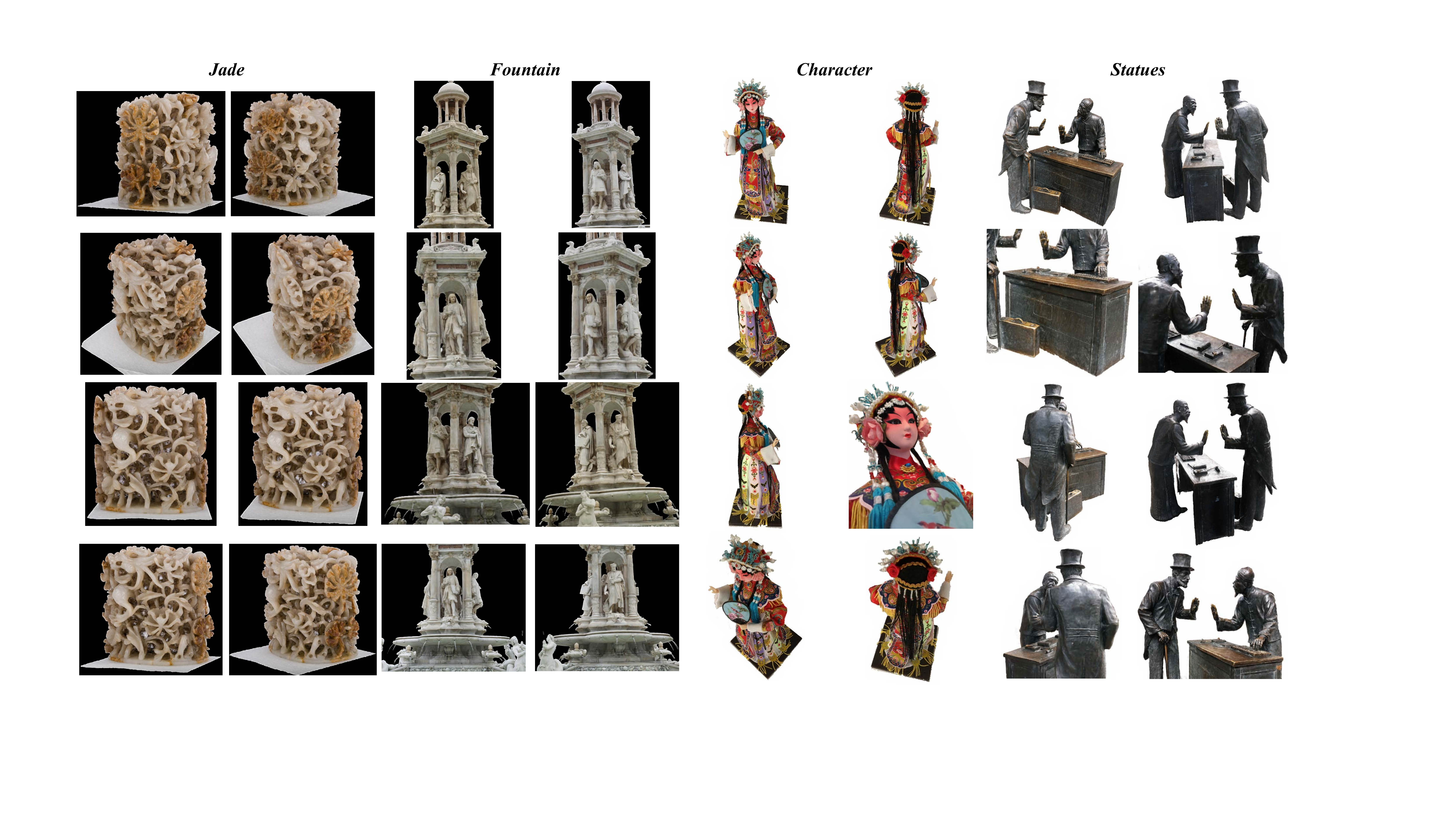}
    \caption{
    The synthesized novel views by our method on the {\bf BlendedMVS}~\cite{YaoLLZRZFQ20} dataset.
    {\footnotesize We manually resize, crop, and compress the images.}
    }
    \label{fig:our_on_blendedmvsdataset}
\end{figure*}

\begin{figure*}[hbpt]
    \centering
    \includegraphics[width=.9\linewidth, trim=0 384 220 80, clip]{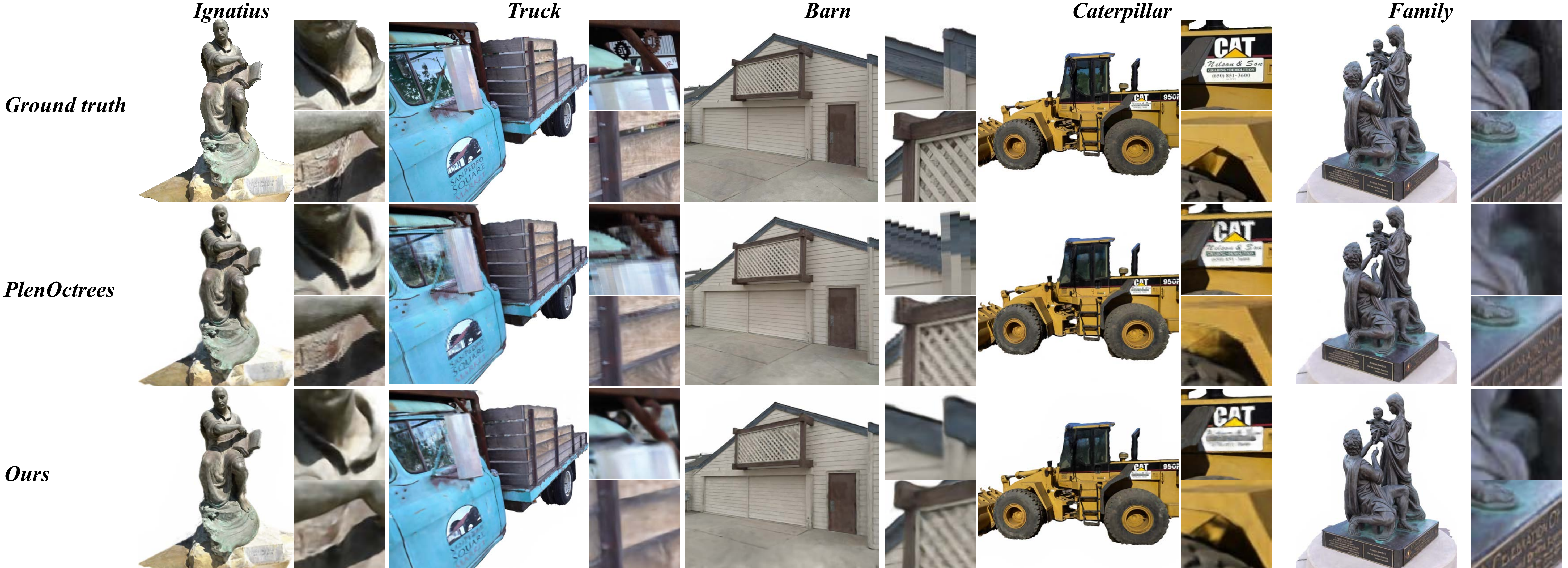}
    \caption{
    Qualitative comparisons on the {\bf Tanks\&Temples}~\cite{KnapitschPZK17} dataset. 
    We manually resize, crop, and compress the images.
    Our quality is comparable to PlenOctrees, and no method produces consistent finer detail.
    Besides, we do not show blocking artifacts.
    }
    \label{fig:our_on_tanksandtempledataset}
\end{figure*}

\begin{figure*}[hbpt]
    \centering
    \includegraphics[width=\linewidth, trim=100 580 220 80, clip]{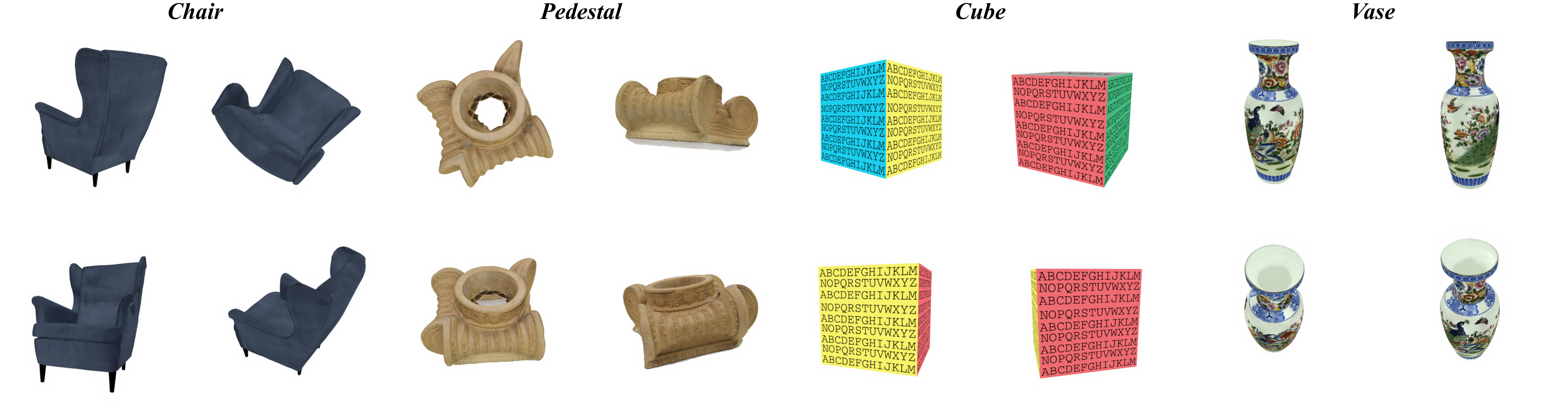}
    \caption{
    The synthesized novel views by our method on the {\bf DeepVoxels}~\cite{SitzmannTHNWZ19} dataset.
    }
    \label{fig:our_on_deepvoxelsdataset}
\end{figure*}

We provide more qualitative results in \cref{fig:our_on_nsvfdataset} for {\bf Synthetic-NSVF}~\cite{LiuGLCT20} dataset, \cref{fig:our_on_blendedmvsdataset} for {\bf BlendedMVS}~\cite{YaoLLZRZFQ20} dataset, and \cref{fig:our_on_deepvoxelsdataset} for {\bf DeepVoxels}~\cite{SitzmannTHNWZ19} dataset.
In \cref{fig:our_on_nerfdataset}, we compare our results with the results provided by JaxNeRF~\cite{jaxnerf2020github} and PlenOctrees~\cite{YuLTLNK2021} on {\bf Synthetic-NeRF}~\cite{MildenhallSTBRN20} dataset.
Both JaxNeRF and PlenOctrees are stronger baselines than the original NeRF~\cite{MildenhallSTBRN20}.
In \cref{fig:our_on_tanksandtempledataset}, we also compare our results with PlenOctrees on the {\bf Tanks\&Temples}~\cite{KnapitschPZK17} dataset.
In the qualitative comparisons, there is no consistent superior method across all the scenes, which matches the results in quantitative comparisons.

\begin{table*}[htpb]
    \vspace{2em}
    \centering
    \begin{tabular}{l|c|cccccccc}
    \hline
    Methods & Avg. & {\it Chair} & {\it Drums} & {\it Ficus} & {\it Hotdog} & {\it Lego} & {\it Materials} & {\it Mic} & {\it Ship} \\
    \hline\hline
    \multicolumn{9}{@{}l}{\rule{0pt}{3ex}\bf PSNR$\uparrow$} \\
    \hline
    SRN~\cite{SitzmannZW19} & 22.26 & 26.96 & 17.18 & 20.73 & 26.81 & 20.85 & 18.09 & 26.85 & 20.60 \\
    NV~\cite{LombardiSSSLS19} & 26.05 & 28.33 & 22.58 & 24.79 & 30.71 & 26.08 & 24.22 & 27.78 & 23.93 \\
    NeRF~\cite{MildenhallSTBRN20} & 31.01 & 33.00 & 25.01 & 30.13 & 36.18 & 32.54 & 29.62 & 32.91 & 28.65 \\
    JaxNeRF~\cite{jaxnerf2020github} & 31.69 & 34.08 & 25.03 & 30.43 & 36.92 & 33.28 & 29.91 & \brown{34.53} & 29.36 \\
    JaxNeRF+~\cite{jaxnerf2020github} & \silver{33.00} & \gold{35.35} & \gold{25.65} & \brown{32.77} & \gold{37.58} & \silver{35.35} & \brown{30.29} & \gold{36.52} & \gold{30.48} \\
    Mip-NeRF~\cite{BarronMTHMS21} & \gold{33.09} & \silver{35.14} & \silver{25.48} & \gold{33.29} & \silver{37.48} & \gold{35.70} & \silver{30.71} & \silver{36.51} & \silver{30.41} \\
    AutoInt~\cite{LindellMW2021} & 25.55 & 25.60 & 20.78 & 22.47 & 32.33 & 25.09 & 25.90 & 28.10 & 24.15 \\
    FastNeRF~\cite{GarbinKJSV2021} & 29.97 & 32.32 & 23.75 & 27.79 & 34.72 & 32.28 & 28.89 & 31.77 & 27.69 \\
    SNeRG~\cite{HedmanSMBD2021} & 30.38 & 33.24 & 24.57 & 29.32 & 34.33 & 33.82 & 27.21 & 32.60 & 27.97 \\
    KiloNeRF~\cite{ReiserPLG2021} & 31.00 & - & - & - & - & - & - & - & - \\
    PlenOctrees~\cite{YuLTLNK2021} & 31.71 & \brown{34.66} & 25.31 & 30.79 & 36.79 & 32.95 & 29.76 & 33.97 & \brown{29.42} \\
    NSVF~\cite{LiuGLCT20} & 31.75 & 33.19 & 25.18 & 31.23 & \brown{37.14} & 32.29 & \gold{32.68} & 34.27 & 27.93 \\
    Ours & \brown{31.95} & 34.09 & \brown{25.44} & \silver{32.78} & 36.74 & \brown{34.64} & 29.57 & 33.20 & 29.13 \\
    \hline

    \multicolumn{9}{@{}l}{\rule{0pt}{3ex}\bf SSIM$\uparrow$} \\
    \hline
    SRN~\cite{SitzmannZW19} & 0.846 & 0.910 & 0.766 & 0.849 & 0.923 & 0.809 & 0.808 & 0.947 & 0.757 \\
    NV~\cite{LombardiSSSLS19} & 0.893 & 0.916 & 0.873 & 0.910 & 0.944 & 0.880 & 0.888 & 0.946 & 0.784 \\
    NeRF~\cite{MildenhallSTBRN20} & 0.947 & 0.967 & 0.925 & 0.964 & 0.974 & 0.961 & 0.949 & 0.980 & 0.856 \\
    JaxNeRF~\cite{jaxnerf2020github} & 0.953 & 0.975 & 0.925 & 0.967 & 0.979 & 0.968 & 0.952 & \brown{0.987} & 0.868 \\
    JaxNeRF+~\cite{jaxnerf2020github} & \gold{0.962} & \gold{0.982} & \gold{0.936} & \gold{0.980} & \gold{0.983} & \gold{0.979} & \silver{0.956} & \gold{0.991} & \gold{0.887} \\
    Mip-NeRF~\cite{BarronMTHMS21} & \silver{0.961} & \silver{0.981} & \brown{0.932} & \gold{0.980} & \silver{0.982} & \silver{0.978} & \gold{0.959} & \gold{0.991} & \brown{0.882} \\
    AutoInt~\cite{LindellMW2021} & 0.911 & 0.928 & 0.861 & 0.898 & 0.974 & 0.900 & 0.930 & 0.948 & 0.852 \\
    FastNeRF~\cite{GarbinKJSV2021} & 0.941 & 0.966 & 0.913 & 0.954 & 0.973 & 0.964 & 0.947 & 0.977 & 0.805 \\
    SNeRG~\cite{HedmanSMBD2021} & 0.950 & 0.975 & 0.929 & 0.967 & 0.971 & 0.973 & 0.938 & 0.982 & 0.865 \\
    KiloNeRF~\cite{ReiserPLG2021} & 0.95 & - & - & - & - & - & - & - & - \\
    PlenOctrees~\cite{YuLTLNK2021} & \brown{0.958} & \silver{0.981} & \silver{0.933} & 0.970 & \silver{0.982} & 0.971 & \brown{0.955} & \brown{0.987} & \silver{0.884} \\
    NSVF~\cite{LiuGLCT20} & 0.953 & 0.968 & 0.931 & 0.973 & 0.980 & 0.960 & 0.973 & \brown{0.987} & 0.854 \\
    Ours & 0.957 & 0.977 & 0.930 & \brown{0.978} & 0.980 & \brown{0.976} & 0.951 & 0.983 & 0.879 \\
    \hline


    \multicolumn{9}{@{}l}{\rule{0pt}{3ex}\bf LPIPS$\downarrow$ {\footnotesize (Vgg)}} \\
    \hline
    SRN~\cite{SitzmannZW19} & 0.170 & 0.106 & 0.267 & 0.149 & 0.100 & 0.200 & 0.174 & 0.063 & 0.299 \\
    NV~\cite{LombardiSSSLS19} & 0.160 & 0.109 & 0.214 & 0.162 & 0.109 & 0.175 & 0.130 & 0.107 & 0.276 \\
    NeRF~\cite{MildenhallSTBRN20} & 0.081 & 0.046 & 0.091 & 0.044 & 0.121 & 0.050 & 0.063 & 0.028 & 0.206 \\
    JaxNeRF~\cite{jaxnerf2020github} & 0.068 & 0.035 & 0.085 & \brown{0.038} & 0.079 & 0.040 & 0.060 & 0.019 & 0.185 \\
    Mip-NeRF~\cite{BarronMTHMS21} & \gold{0.043} & \gold{0.021} & \gold{0.065} & \gold{0.020} & \gold{0.027} & \gold{0.021} & \gold{0.040} & \gold{0.009} & \gold{0.138} \\
    PlenOctrees~\cite{YuLTLNK2021} & \silver{0.053} & \silver{0.022} & \silver{0.076} & \brown{0.038} & \silver{0.032} & \brown{0.034} & \brown{0.059} & \silver{0.017} & \silver{0.144} \\
    Ours & \silver{0.053} & \brown{0.027} & \brown{0.077} & \silver{0.024} & \brown{0.034} & \silver{0.028} & \silver{0.058} & \silver{0.017} & \brown{0.161} \\
    \hline

    \multicolumn{9}{@{}l}{\rule{0pt}{3ex}\bf LPIPS$\downarrow$ {\footnotesize (Alex)}} \\
    \hline
    NSVF~\cite{LiuGLCT20} & \silver{0.047} & \silver{0.043} & \silver{0.069} & \silver{0.017} & \silver{0.025} & \silver{0.029} & \gold{0.021} & \gold{0.010} & \silver{0.162} \\
    Ours & \gold{0.035} & \gold{0.016} & \gold{0.061} & \gold{0.015} & \gold{0.017} & \gold{0.014} & \silver{0.026} & \silver{0.014} & \gold{0.118} \\
    \hline

    \end{tabular}
    \caption{Quantitative results on each scene from the {\bf Synthetic-NeRF}~\cite{MildenhallSTBRN20} dataset. We highlight the top 3 results of each column under each metric in \gold{gold}, \silver{silver}, and \brown{bronze}.}
    \label{tab:supp_breakdown_nerf}
    \vspace{2em}
\end{table*}

\begin{table*}[htpb]
    \centering
    \begin{tabular}{l|c|cccccccc}
    \hline
    Methods & Avg. & {\it Wineholder} & {\it Steamtrain} & {\it Toad} & {\it Robot} & {\it Bike} & {\it Palace} & {\it Spaceship} & {\it Lifestyle} \\
    \hline\hline
    \multicolumn{9}{@{}l}{\rule{0pt}{3ex}\bf PSNR$\uparrow$} \\
    \hline
    SRN~\cite{SitzmannZW19} & 24.33 & 20.74 & 25.49 & 25.36 & 22.27 & 23.76 & 24.45 & 27.99 & 24.58 \\
    NV~\cite{LombardiSSSLS19} & 25.83 & 21.32 & 25.31 & 24.63 & 24.74 & 26.65 & 26.38 & 29.90 & 27.68 \\
    NeRF~\cite{MildenhallSTBRN20} & 30.81 & \brown{28.23} & \brown{30.84} & \brown{29.42} & \brown{28.69} & \brown{31.77} & \brown{31.76} & \brown{34.66} & \brown{31.08} \\
    KiloNeRF~\cite{ReiserPLG2021} & \brown{33.37} & - & - & - & - & - & - & - & - \\
    NSVF~\cite{LiuGLCT20} & \gold{35.13} & \gold{32.04} & \silver{35.13} & \gold{33.25} & \silver{35.24} & \silver{37.75} & \silver{34.05} & \gold{39.00} & \gold{34.60} \\
    Ours & \silver{35.08} & \silver{30.26} & \gold{36.56} & \silver{33.10} & \gold{36.36} & \gold{38.33} & \gold{34.49} & \silver{37.71} & \silver{33.79} \\
    \hline

    \multicolumn{9}{@{}l}{\rule{0pt}{3ex}\bf SSIM$\uparrow$} \\
    \hline
    SRN~\cite{SitzmannZW19} & 0.882 & 0.850 & 0.923 & 0.822 & 0.904 & 0.926 & 0.792 & 0.945 & 0.892 \\
    NV~\cite{LombardiSSSLS19} & 0.892 & 0.828 & 0.900 & 0.813 & 0.927 & 0.943 & 0.826 & 0.956 & 0.941 \\
    NeRF~\cite{MildenhallSTBRN20} & 0.952 & \brown{0.920} & \brown{0.966} & \brown{0.920} & \brown{0.960} & \brown{0.970} & \brown{0.950} & \brown{0.980} & \brown{0.946} \\
    KiloNeRF~\cite{ReiserPLG2021} & \brown{0.97} & - & - & - & - & - & - & - & - \\
    NSVF~\cite{LiuGLCT20} & \gold{0.979} & \gold{0.965} & \silver{0.986} & \gold{0.968} & \silver{0.988} & \gold{0.991} & \gold{0.969} & \gold{0.991} & \gold{0.971} \\
    Ours & \silver{0.975} & \silver{0.949} & \gold{0.989} & \silver{0.966} & \gold{0.992} & \gold{0.991} & \silver{0.962} & \silver{0.988} & \silver{0.965} \\
    \hline

    \multicolumn{9}{@{}l}{\rule{0pt}{3ex}\bf LPIPS$\downarrow$ {\footnotesize (Vgg)}} \\
    \hline
    Ours & 0.033 & 0.055 & 0.019 & 0.047 & 0.013 & 0.011 & 0.043 & 0.019 & 0.054 \\
    \hline

    \multicolumn{9}{@{}l}{\rule{0pt}{3ex}\bf LPIPS$\downarrow$ {\footnotesize (Alex)}} \\
    \hline
    SRN~\cite{SitzmannZW19} & 0.141 & 0.224 & 0.082 & 0.204 & 0.120 & 0.075 & 0.240 & 0.061 & 0.120 \\
    NV~\cite{LombardiSSSLS19} & 0.124 & 0.204 & 0.121 & 0.192 & 0.096 & 0.067 & 0.173 & 0.056 & 0.088 \\
    NeRF~\cite{MildenhallSTBRN20} & \brown{0.043} & \brown{0.096} & \brown{0.031} & \brown{0.069} & \brown{0.038} & \brown{0.019} & \brown{0.031} & \brown{0.016} & \brown{0.047} \\
    NSVF~\cite{LiuGLCT20} & \gold{0.015} & \gold{0.020} & \gold{0.010} & \silver{0.032} & \silver{0.007} & \gold{0.004} & \gold{0.018} & \gold{0.006} & \gold{0.020} \\
    Ours & \silver{0.019} & \silver{0.038} & \gold{0.010} & \gold{0.030} & \gold{0.005} & \gold{0.004} & \silver{0.027} & \silver{0.009} & \silver{0.027} \\
    \hline

    \end{tabular}
    \caption{Quantitative results on each scene from the {\bf Synthetic-NSVF}~\cite{LiuGLCT20} dataset. We highlight the top 3 results of each column under each metric in \gold{gold}, \silver{silver}, and \brown{bronze}.}
    \label{tab:supp_breakdown_nsvf}
\end{table*}

\begin{table*}[htpb]
    \centering
    \begin{tabular}{l|c|cccc}
    \hline
    Methods & Avg. & {\it Jade} & {\it Fountain} & {\it Character} & {\it Statues} \\
    \hline\hline
    \multicolumn{6}{@{}l}{\rule{0pt}{3ex}\bf PSNR$\uparrow$} \\
    \hline
    SRN~\cite{SitzmannZW19} & 20.51 & 18.57 & 21.04 & 21.98 & 20.46 \\
    NV~\cite{LombardiSSSLS19} & 23.03 & 22.08 & 22.71 & 24.10 & 23.22 \\
    NeRF~\cite{MildenhallSTBRN20} & 24.15 & \brown{21.65} & \brown{25.59} & \brown{25.87} & \brown{23.48} \\
    KiloNeRF~\cite{ReiserPLG2021} & \silver{27.39} & - & - & - & - \\
    NSVF~\cite{LiuGLCT20} & \brown{26.89} & \silver{26.96} & \silver{27.73} & \silver{27.95} & \silver{24.97} \\
    Ours & \gold{28.02} & \gold{27.68} & \gold{28.48} & \gold{30.31} & \gold{25.62} \\
    \hline

    \multicolumn{6}{@{}l}{\rule{0pt}{3ex}\bf SSIM$\uparrow$} \\
    \hline
    SRN~\cite{SitzmannZW19} & 0.770 & 0.715 & 0.717 & 0.853 & 0.794 \\
    NV~\cite{LombardiSSSLS19} & 0.793 & \brown{0.750} & 0.762 & 0.876 & 0.785 \\
    NeRF~\cite{MildenhallSTBRN20} & 0.828 & \brown{0.750} & \brown{0.860} & \brown{0.900} & \brown{0.800} \\
    KiloNeRF~\cite{ReiserPLG2021} & \silver{0.92} & - & - & - \\
    NSVF~\cite{LiuGLCT20} & \brown{0.898} & \silver{0.901} & \silver{0.913} & \silver{0.921} & \silver{0.858} \\
    Ours & \gold{0.922} & \gold{0.914} & \gold{0.926} & \gold{0.963} & \gold{0.883} \\
    \hline

    \multicolumn{6}{@{}l}{\rule{0pt}{3ex}\bf LPIPS$\downarrow$ {\footnotesize (Vgg)}} \\
    \hline
    Ours & 0.101 & 0.108 & 0.115 & 0.045 & 0.137 \\
    \hline

    \multicolumn{6}{@{}l}{\rule{0pt}{3ex}\bf LPIPS$\downarrow$ {\footnotesize (Alex)}} \\
    \hline
    SRN~\cite{SitzmannZW19} & 0.294 & 0.323 & 0.291 & 0.208 & 0.354 \\
    NV~\cite{LombardiSSSLS19} & 0.243 & 0.292 & 0.263 & \brown{0.140} & 0.277 \\
    NeRF~\cite{MildenhallSTBRN20} & \brown{0.192} & \brown{0.264} & \brown{0.149} & 0.149 & \brown{0.206} \\
    NSVF~\cite{LiuGLCT20} & \silver{0.114} & \silver{0.094} & \silver{0.113} & \silver{0.074} & \silver{0.171} \\
    Ours & \gold{0.075} & \gold{0.075} & \gold{0.086} & \gold{0.029} & \gold{0.110} \\
    \hline

    \end{tabular}
    \caption{Quantitative results on each scene from the {\bf BlendedMVS}~\cite{YaoLLZRZFQ20} dataset. We highlight the top 3 results of each column under each metric in \gold{gold}, \silver{silver}, and \brown{bronze}.}
    \label{tab:supp_breakdown_blendedmvs}
\end{table*}

\begin{table*}[htpb]
    \centering
    \begin{tabular}{l|c|ccccc}
    \hline
    Methods & Avg. & {\it Ignatius} & {\it Truck} & {\it Barn} & {\it Caterpillar} & {\it Family} \\
    \hline\hline
    \multicolumn{7}{@{}l}{\rule{0pt}{3ex}\bf PSNR$\uparrow$} \\
    \hline
    SRN~\cite{SitzmannZW19} & 24.10 & 26.70 & 22.62 & 22.44 & 21.14 & 27.57 \\
    NV~\cite{LombardiSSSLS19} & 23.70 & 26.54 & 21.71 & 20.82 & 20.71 & 28.72 \\
    NeRF~\cite{MildenhallSTBRN20} & 25.78 & 25.43 & 25.36 & 24.05 & 23.75 & 30.29 \\
    JaxNeRF~\cite{jaxnerf2020github} & 27.94 & \brown{27.95} & 26.66 & \gold{27.39} & 25.24 & 32.47 \\
    KiloNeRF~\cite{ReiserPLG2021} & \silver{28.41} & - & - & - & - & - \\
    PlenOctrees~\cite{YuLTLNK2021} & 27.99 & \gold{28.19} & \brown{26.83} & 26.80 & \brown{25.29} & \brown{32.85} \\
    NSVF~\cite{LiuGLCT20} & \gold{28.48} & 27.91 & \silver{26.92} & \silver{27.16} & \gold{26.44} & \silver{33.58} \\
    Ours & \silver{28.41} & \silver{28.16} & \gold{27.15} & \brown{27.01} & \silver{26.00} & \gold{33.75} \\
    \hline

    \multicolumn{7}{@{}l}{\rule{0pt}{3ex}\bf SSIM$\uparrow$} \\
    \hline
    SRN~\cite{SitzmannZW19} & 0.847 & 0.920 & 0.832 & 0.741 & 0.834 & 0.908 \\
    NV~\cite{LombardiSSSLS19} & 0.834 & 0.922 & 0.793 & 0.721 & 0.819 & 0.916 \\
    NeRF~\cite{MildenhallSTBRN20} & 0.864 & 0.920 & 0.860 & 0.750 & 0.860 & 0.932 \\
    JaxNeRF~\cite{jaxnerf2020github} & 0.904 & \brown{0.940} & \brown{0.896} & \silver{0.842} & 0.892 & 0.951 \\
    KiloNeRF~\cite{ReiserPLG2021} & \brown{0.91} & - & - & - & - & - \\
    PlenOctrees~\cite{YuLTLNK2021} & \gold{0.917} & \gold{0.948} & \gold{0.914} & \gold{0.856} & \gold{0.907} & \gold{0.962} \\
    NSVF~\cite{LiuGLCT20} & 0.901 & 0.930 & 0.895 & 0.823 & \brown{0.900} & \brown{0.954} \\
    Ours & \silver{0.911} & \silver{0.944} & \silver{0.906} & \brown{0.838} & \silver{0.906} & \gold{0.962} \\
    \hline

    \multicolumn{7}{@{}l}{\rule{0pt}{3ex}\bf LPIPS$\downarrow$ {\footnotesize (Vgg)}} \\
    \hline
    JaxNeRF~\cite{jaxnerf2020github} & \brown{0.168} & \brown{0.102} & \brown{0.173} & \silver{0.286} & \brown{0.189} & \brown{0.092} \\
    PlenOctrees~\cite{YuLTLNK2021} & \gold{0.131} & \gold{0.080} & \gold{0.130} & \gold{0.226} & \gold{0.148} & \gold{0.069} \\
    Ours & \silver{0.155} & \silver{0.083} & \silver{0.160} & \brown{0.294} & \silver{0.167} & \silver{0.070} \\
    \hline

    \multicolumn{7}{@{}l}{\rule{0pt}{3ex}\bf LPIPS$\downarrow$ {\footnotesize (Alex)}} \\
    \hline
    SRN~\cite{SitzmannZW19} & 0.251 & 0.128 & 0.266 & 0.448 & 0.278 & 0.134 \\ 
    NV~\cite{LombardiSSSLS19} & 0.260 & 0.117 & 0.312 & 0.479 & 0.280 & 0.111 \\
    NeRF~\cite{MildenhallSTBRN20} & \brown{0.198} & \brown{0.111} & \brown{0.192} & \brown{0.395} & \brown{0.196} & \brown{0.098} \\
    NSVF~\cite{LiuGLCT20} & \silver{0.155} & \silver{0.106} & \silver{0.148} & \silver{0.307} & \gold{0.141} & \gold{0.063} \\
    Ours & \gold{0.148} & \gold{0.090} & \gold{0.145} & \gold{0.290} & \silver{0.152} & \silver{0.064} \\
    \hline

    \end{tabular}
    \caption{Quantitative results on each scene from the {\bf Tanks\&Temples}~\cite{KnapitschPZK17} dataset dataset. We highlight the top 3 results of each column under each metric in \gold{gold}, \silver{silver}, and \brown{bronze}.}
    \label{tab:supp_breakdown_tanksandtemples}
\end{table*}

\begin{table*}[htpb]
    \centering
    \begin{tabular}{l|c|cccc}
    \hline
    Methods & Avg. & {\it Chair} & {\it Pedestal} & {\it Cube} & {\it Vase} \\
    \hline\hline
    \multicolumn{6}{@{}l}{\rule{0pt}{3ex}\bf PSNR$\uparrow$} \\
    \hline
    Nearest Neighbor & 20.94 & 20.69 & 21.49 & 18.32 & 23.26 \\
    NV~\cite{LombardiSSSLS19} & 29.62 & 35.15 & 36.47 & 26.48 & 20.39 \\
    DeepVoxels~\cite{SitzmannTHNWZ19} & 30.55 & 33.45 & 32.35 & 28.42 & 27.99 \\
    DeepVoxels++~\cite{HeCJS20} & 37.31 & \brown{40.87} & \brown{38.93} & \brown{36.51} & 32.91 \\
    SRN~\cite{SitzmannZW19} & 33.20 & 36.67 & 35.91 & 28.74 & 31.46 \\
    LLFF~\cite{MildenhallSCKRN19} & 34.38 & 36.11 & 35.87 & 32.58 & \brown{32.97} \\
    NeRF~\cite{MildenhallSTBRN20} & \brown{40.15} & \silver{42.65} & \silver{41.44} & \silver{39.19} & \silver{37.32} \\
    IBRNet~\cite{WangWGSZBMSF21} & \silver{42.93} & - & - & - & - \\
    Ours & \gold{45.83} & \gold{48.48} & \gold{48.51} & \gold{43.77} & \gold{42.54} \\
    \hline

    \multicolumn{6}{@{}l}{\rule{0pt}{3ex}\bf SSIM$\uparrow$} \\
    \hline
    Nearest Neighbor & 0.89 & 0.94 & 0.87 & 0.83 & 0.92 \\
    NV~\cite{LombardiSSSLS19} & 0.929 & 0.980 & 0.963 & 0.916 & 0.857 \\
    DeepVoxels~\cite{SitzmannTHNWZ19} & 0.97 & 0.99 & 0.97 & 0.97 & 0.96 \\
    DeepVoxels++~\cite{HeCJS20} & 0.99 & 0.99 & 0.98 & 0.99 & 0.98 \\
    SRN~\cite{SitzmannZW19} & 0.963 & 0.982 & 0.957 & 0.944 & 0.969 \\
    LLFF~\cite{MildenhallSCKRN19} & 0.985 & \silver{0.992} & \brown{0.983} & \brown{0.983} & \brown{0.983} \\
    NeRF~\cite{MildenhallSTBRN20} & \brown{0.991} & \brown{0.991} & \silver{0.986} & \silver{0.996} & \silver{0.992} \\
    IBRNet~\cite{WangWGSZBMSF21} & \silver{0.997} & - & - & - & - \\
    Ours & \gold{0.998} & \gold{0.998} & \gold{0.998} & \gold{0.998} & \gold{0.998} \\
    \hline

    \multicolumn{6}{@{}l}{\rule{0pt}{3ex}\bf LPIPS$\downarrow$ {\footnotesize (Vgg)}} \\
    \hline
    SRN~\cite{SitzmannZW19} & 0.073 & 0.093 & 0.081 & 0.074 & 0.044 \\
    NV~\cite{LombardiSSSLS19} & 0.099 & 0.096 & 0.069 & 0.113 & 0.117 \\
    LLFF~\cite{MildenhallSCKRN19} & 0.048 & \brown{0.051} & \brown{0.039} & \brown{0.064} & \brown{0.039} \\
    NeRF~\cite{MildenhallSTBRN20} & \brown{0.023} & \silver{0.047} & \silver{0.024} & \silver{0.006} & \silver{0.017} \\
    IBRNet~\cite{WangWGSZBMSF21} & \silver{0.009} & - & - & - & - \\
    Ours & \gold{0.006} & \gold{0.015} & \gold{0.003} & \gold{0.002} & \gold{0.004} \\
    \hline

    \multicolumn{6}{@{}l}{\rule{0pt}{3ex}\bf LPIPS$\downarrow$ {\footnotesize (Alex)}} \\
    \hline
    Ours & 0.002 & 0.005 & 0.001 & 0.001 & 0.002 \\
    \hline

    \end{tabular}
    \caption{Quantitative results on each scene from the {\bf DeepVoxels}~\cite{SitzmannTHNWZ19} dataset. We highlight the top 3 results of each column under each metric in \gold{gold}, \silver{silver}, and \brown{bronze}.}
    \label{tab:supp_breakdown_deepvoxels}
\end{table*}

\clearpage\clearpage  

\setcounter{table}{0}
\setcounter{figure}{0}
\section{Derivation of low-density initialization} \label{sec:supp_derive_low_dense}

In this section, we derive the bias term $b$ in the low-density initialization to make the activated alpha value very close to zero (\ie, $\alpha^{\text{(init)(c)}}\approx 0$) at the beginning of our coarse stage optimization.
More specifically, $\alpha^{\text{(init)(c)}}$ is a hyperparameter, and $\left(1-\alpha^{\text{(init)(c)}}\right)$ means the decay factor of the accumulated transmittance for a ray tracing forward a distance of the voxel size $s^{\text{(c)}}$. 

Below we omit the stage-specific superscript for simplicity.
Let the accumulated distance of $s$ be divided into $N$ segments, each with a distance of $\delta_i$, \ie, 
\[
    \sum_{i=1}^N \delta_i = s ~.
\]
Recall that, in practice, we initialize all raw values $\ddot{\sigma}$ in $\bm{V}^{\text{(density)(c)}}$ to $0$ and thus the initial activated densities $\sigma_i$ are all equal to $\log\left(1 + \exp(b)\right)$.
Then, the decay factor of the accumulated transmittance can be written as
\begin{equation*}
\begin{split}
    \prod_{i=1}^N (1 - \alpha_i)
    &= \prod_{i=1}^N \left(\left(1 - \left(1 - \exp(-\sigma_i \delta_i)\right)\right)\right)\\
    &= \prod_{i=1}^N \exp(-\sigma_i \delta_i) \\
    &= \prod_{i=1}^N \exp\Big(\log\left(1 + \exp(b)\right) \cdot (-\delta_i)\Big) \\
     &= \exp\left( \sum_{i=1}^N \Big( \log\left(1 + \exp(b)\right) \cdot (-\delta_i) \Big) \right) \\
    &= \exp\left( \log\left(1 + \exp(b)\right) \cdot \sum_{i=1}^N (-\delta_i) \right) \\
    &= \exp\Big( \log\left(1 + \exp(b)\right) \cdot (-s) \Big) \\
    &= \left(1 + \exp(b)\right)^{-s} ~.
\end{split}
\end{equation*}
We want to find the $b$ such that the decay factor is $(1-\alpha^{\text{(init)}})$ for passing through a distance of the voxel size $s$.
Therefore, we have
\[
b = \log\left(\left(1 - \alpha^{\text{(init)}}\right)^{-\frac{1}{s}} - 1\right) ~,
\]
which is the equation presented in our main paper to impose the prior of low-density initialization.

\setcounter{table}{0}
\setcounter{figure}{0}
\section{Derivation of post-activation} \label{sec:supp_derive_post_act}
By expanding the post-activated trilinear interpolation, we have 
\begin{equation} \label{eq:supp_post_activation}
\begin{split}
    \alpha^{\text{(post)}}
    &= 1 - \exp\left(-\delta\log\left(1+\exp\left( \operatorname{interp}\left(\bm{x}, \bm{V}\right) \right)\right)\right) \\
    &= 1 - \left(1 + \exp\left(\operatorname{interp}\left(\bm{x}, \bm{V}\right)\right)\right)^{-\delta} ~ ,
\end{split}
\end{equation}
where $\delta$ is a volume rendering related value and is treated as a constant in later derivation, and $\operatorname{interp}$ is the interpolation operation of a spatial point $\bm{x}$ in grid $\bm{V}$.
The bias term $b$ is omitted here for simplicity.
The overall activation maps an interpolant ($\operatorname{interp}\left(\bm{x}, \bm{V}\right) \in \mathbb{R}$) into an alpha value ($\alpha^{\text{(post)}} \in [0, 1]$).

We want to prove that such a post-activation can produce sharp linear surface (decision boundary) in a single grid cell.
Let first prove in the simplest 1D grid and then we generalize it to 2D grid.
The case in 3D grid then can be proven easily using the derivations in 1D and 2D grid.

\subsection{Derivation for a 1D grid cell} \label{ssec:proof_1d}
\begin{figure}[h]
    \centering
    \begin{subfigure}[t]{0.495\linewidth}
        \centering
        \includegraphics[width=\linewidth]{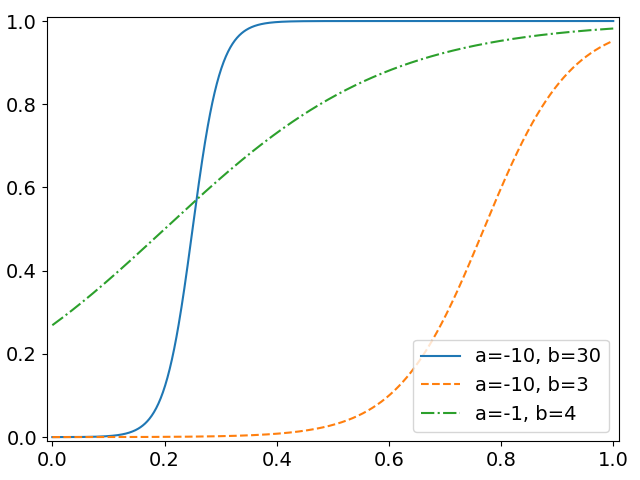}
        \caption{Three examples of $S(\bm{x};~a,b)$ with $\delta=1$}
    \end{subfigure}
    \hfill
    \begin{subfigure}[t]{0.495\linewidth}
        \centering
        \includegraphics[width=\linewidth]{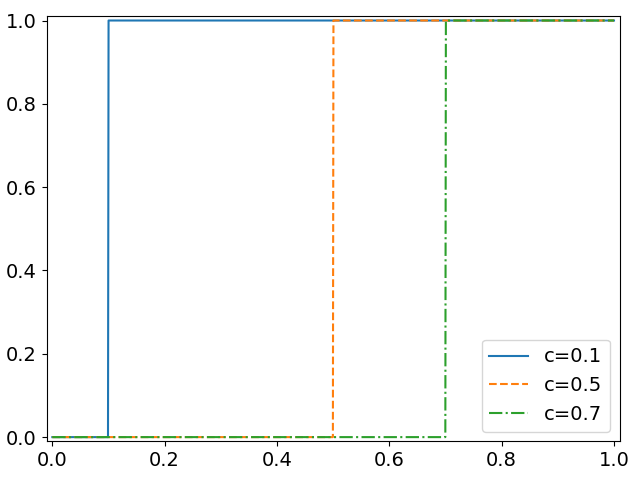}
        \caption{Three examples of $T(\bm{x};~c)$}
        \label{fig:vis_T}
    \end{subfigure}
    \caption{Visualization for some examples of $S$ and $T$ with different parameter settings.}
    \label{fig:vis_S_T}
\end{figure}

In 1D grid, $\bm{x}$ is a scalar.
Without loss of generality, we assume $\bm{x}=0$ and $\bm{x}=1$ are the left and right bound of the 1D grid cell, respectively.
Let $a, b \in \mathbb{R}$ be the grid values stored at the cell's left and right bound.
Then \cref{eq:supp_post_activation} with 1D linear interpolation of the cell can be re-written for this specific case:
\begin{equation} \label{eq:1d_proof_linear_interp}
    S(\bm{x};~a,b) = 1 - \left(1 + \exp( a(1-\bm{x}) + b\bm{x} )\right)^{-\delta} ~ .
\end{equation}
Consider a target function $T(\bm{x};~c)$ in the form of a shifted unit step function,
\begin{equation} \label{eq:1d_target}
\begin{split}
    T(\bm{x};~c)
    &= \mathbbm{1}(\bm{x}-c)\\
    &= \left\{\begin{matrix}
        1\,, & \bm{x}>c\,, \\ 
        0\,, & \bm{x}\leq c\,,
        \end{matrix}\right.
\end{split} ~ 
\end{equation}
where $0<c<1$ is the position of the target linear surface (decision boundary) in the 1D grid cell.
We only derive for the occupancy at right-hand side as the opposite direction can be trivially generalized.
Visualization for $S$ and $T$ with some specific parameters is shown in \cref{fig:vis_S_T}.

We show that $S(\bm{x};~a,b)$ can be made arbitrarily close to $T(\bm{x};~c)$. Specifically,
given any $\epsilon$ and $\Delta$ satisfying $0<\epsilon<1$ and $0<\Delta<\min(c,1-c)$, 
we can find the grid values $a, b$ such that
\[
    |S(\bm{x};~a,b) - T(\bm{x};~c)| \leq \epsilon,~ \forall |\bm{x}-c| \geq \Delta ~.
\]
As the function $S$ is a monotonically increasing function bounded in $[0, 1]$, the criterion can then be simplified into
\begin{subequations} \label{eq:1d_proof_c}
\begin{align}
    1-S(c+\Delta;~a,b) &\leq \epsilon ~, \label{eq:1d_proof_c1} \\ 
    S(c-\Delta;~a,b)   &\leq \epsilon ~. \label{eq:1d_proof_c2}
\end{align}
\end{subequations}
By expanding the inequality (\ref{eq:1d_proof_c1}), we get
\begin{align*}
    &1-S(c+\Delta;~a,b) \\
    &= 1 - \left(1 - \left(1 + \exp( a(1-(c+\Delta)) + b(c+\Delta) \right)^{-\delta}\right) \\
    &= \left(1 + \exp( a(1-(c+\Delta)) + b(c+\Delta) \right)^{-\delta} \leq \epsilon ~.
\end{align*}
We solve the inequality:
\begin{align*}
    \left(\frac{1}{1 + \exp( a(1-(c+\Delta)) + b(c+\Delta) )}\right)^{\delta} &\leq \epsilon \\
    \frac{1}{1 + \exp( a(1-(c+\Delta)) + b(c+\Delta) )} &\leq \epsilon^{1/\delta} \\
    1 + \exp( a(1-(c+\Delta)) + b(c+\Delta) ) &\geq \frac{1}{\epsilon^{1/\delta}} \\
    \exp( a(1-(c+\Delta)) + b(c+\Delta) ) &\geq \frac{1}{\epsilon^{1/\delta}}-1 ~.
\end{align*}
Finally, we obtain
\begin{equation} \label{eq:1d_proof_ineq1}
    a(1-(c+\Delta)) + b(c+\Delta) \geq \log\left(\frac{1}{\epsilon^{1/\delta}}-1\right) ~.
\end{equation}
By applying the same process to the inequality (\ref{eq:1d_proof_c2}), we get another inequality:
\begin{equation} \label{eq:1d_proof_ineq2}
    a(1-(c-\Delta)) + b(c-\Delta) \leq \log\left( \frac{1}{(1 - \epsilon)^{1/\delta}} - 1 \right) ~.
\end{equation}
There are infinite numbers of solutions satisfying the inequalities (\ref{eq:1d_proof_ineq1}) and (\ref{eq:1d_proof_ineq2}).
Here, we introduce one more constraint to make the derivation and later extensions simpler:
\begin{equation} \label{eq:1d_extra_constraint}
    S(c;~a,b) = 0.5 ~.
\end{equation}
From \cref{eq:1d_proof_linear_interp} with this constraint we can then derive the linear relation between $a, b$:
\begin{equation} \label{eq:1d_b_linear_to_a}
    b = a\frac{c-1}{c} + \frac{\log(2^{\frac{1}{\delta}} - 1)}{c} ~.
\end{equation}
Substitute $b$ in the inequality (\ref{eq:1d_proof_ineq1}), we get
\begin{align*}
    &a(1-(c+\Delta)) + b(c+\Delta) \\
    &= a(1-(c+\Delta)) + \left(a\frac{c-1}{c} + \frac{\log(2^{\frac{1}{\delta}} - 1)}{c}\right)(c+\Delta) \\
    &= a\frac{-\Delta}{c} + \log(2^{\frac{1}{\delta}} - 1) \frac{c+\Delta}{c}
    \geq \log\left(\frac{1}{\epsilon^{1/\delta}}-1\right) ~.
\end{align*}
Similarly, substitute $b$ in the inequality (\ref{eq:1d_proof_ineq2}), we get
\begin{align*}
    &a(1-(c-\Delta)) + b(c-\Delta) \\
    &= a(1-(c-\Delta)) + \left(a\frac{c-1}{c} + \frac{\log(2^{\frac{1}{\delta}} - 1)}{c}\right)(c-\Delta) \\
    &= a\frac{\Delta}{c} + \log(2^{\frac{1}{\delta}} - 1)\frac{c-\Delta}{c}
    \leq \log\left( \frac{1}{(1 - \epsilon)^{1/\delta}} - 1 \right) ~.
\end{align*}
In summary, by adding the extra constraint (\ref{eq:1d_extra_constraint}), the inequality (\ref{eq:1d_proof_ineq1}) and (\ref{eq:1d_proof_ineq2}) become
\begin{align} \label{eq:1d_a_upper_bound_loose}
    \left\{\begin{matrix}
    a \leq& \log(2^{\frac{1}{\delta}} - 1) \frac{c+\Delta}{\Delta} - \log\left(\frac{1}{\epsilon^{1/\delta}}-1\right)\frac{c}{\Delta} \\
    a \leq& \log\left( \frac{1}{(1 - \epsilon)^{1/\delta}} - 1 \right)\frac{c}{\Delta} - \log(2^{\frac{1}{\delta}} - 1)\frac{c-\Delta}{\Delta} ~,
    \end{matrix}\right.
\end{align}
which defines the upper bound of $a$ such that the conditions on the function $S$ in (\ref{eq:1d_proof_c1}), (\ref{eq:1d_proof_c2}), and (\ref{eq:1d_extra_constraint}) can all be satisfied.
The tighter upper bound is
\begin{multline} \label{eq:1d_a_upper_bound}
    a^{\text{(upper bound)}} = \\
    \begin{cases}
        \log(2^{\frac{1}{\delta}} - 1) \frac{c+\Delta}{\Delta} - \log\left(\frac{1}{\epsilon^{1/\delta}}-1\right)\frac{c}{\Delta}, & \text{if } \delta < 1 \\
        \log\left( \frac{1}{(1 - \epsilon)^{1/\delta}} - 1 \right)\frac{c}{\Delta} - \log(2^{\frac{1}{\delta}} - 1)\frac{c-\Delta}{\Delta}, & \text{otherwise}
    \end{cases} ~,
\end{multline}
where the $\delta > 0$ is the pre-defined step size in volume rendering and we skip the detailed derivation of \cref{eq:1d_a_upper_bound} here.

\begin{figure}[h]
    \centering
    \begin{subtable}{\linewidth}
        \centering
        \begin{tabular}{c|c|c}
        \hline
        $T(\bm{x};~c)$ & By \cref{eq:1d_a_upper_bound} & By \cref{eq:1d_b_linear_to_a} \\
        \hline\hline
        $c=0.1$ & $a \approx -172.1$ & $b \approx 1560.1$ \\
        $c=0.5$ & $a \approx -865.0$ & $b \approx 867.2$ \\
        $c=0.7$ & $a \approx -1211.4$ & $b \approx 520.8$ \\
        \hline
        \end{tabular}
    \end{subtable}
    \par\medskip
    \begin{subfigure}[t]{0.49\linewidth}
        \centering
        \includegraphics[width=\linewidth]{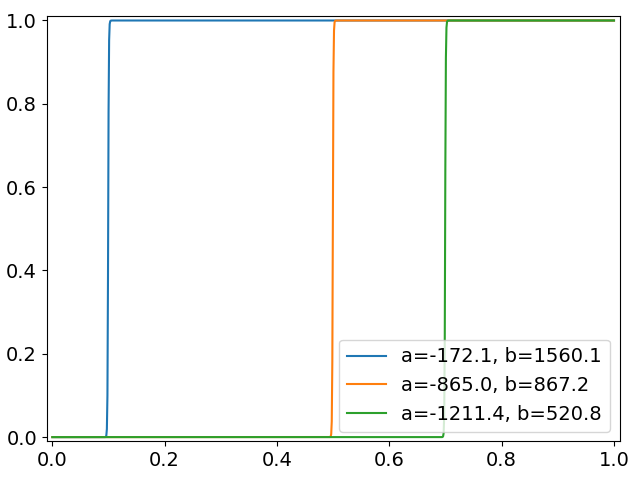}
        \caption{The derived $S(\bm{x};~a,b)$.}
    \end{subfigure}
    \hfill
    \begin{subfigure}[t]{0.49\linewidth}
        \centering
        \includegraphics[width=\linewidth]{figs/proof/proof_example_T.png}
        \caption{Some specific $T(\bm{x};~c)$}
    \end{subfigure}
    \caption{Using the derived upper bound (\ref{eq:1d_a_upper_bound}) and \cref{eq:1d_b_linear_to_a} to directly find the grid values $a, b$ that fit the target function $T(\bm{x};~c)$.}
    \label{fig:verify_1d_plot}
\end{figure}

As a verification, we re-use the examples in \cref{fig:vis_T} as the target functions and set the tolerance to $\epsilon=10^{-4}, \Delta=10^{-2}$ and the volume rendering related value to $\delta=0.5$.
We can directly find the grid values $a, b$ using the derivations given above.
We show the derived numbers and the resulting plot visualization in \cref{fig:verify_1d_plot}. It can be seen that the derived $S(\bm{x};~a,b)$ can faithfully resemble the target $T(\bm{x};~c)$.

\subsection{Derivation for a 2D grid cell} \label{ssec:proof_2d}
\begin{figure}[h]
    \centering
    \begin{subfigure}[t]{0.45\linewidth}
        \includegraphics[width=\linewidth, trim=0 0 7cm 0, clip]{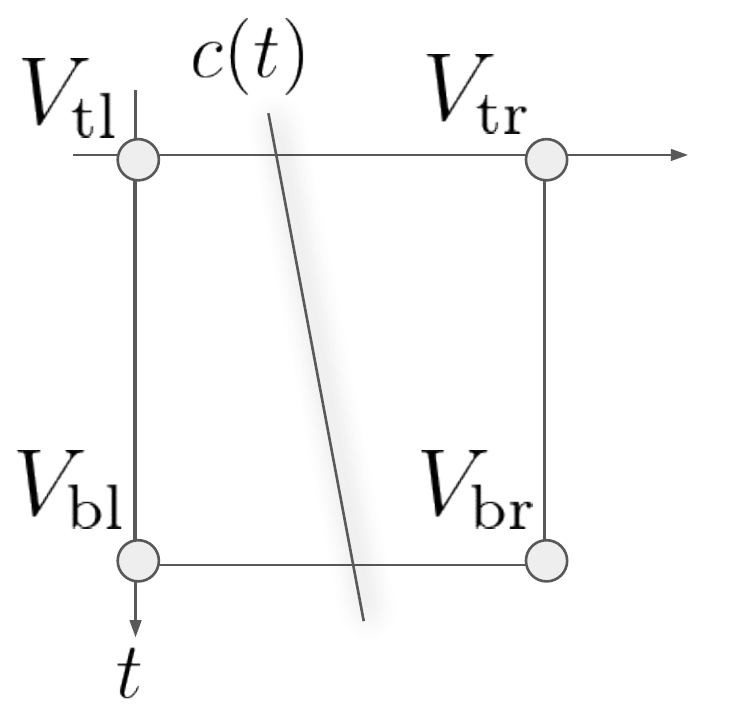}
        \caption{\footnotesize
            {\bf Case I:} The boundary intersects the top and bottom edge of the grid.
        }
        \label{fig:2d_setup_case1}
    \end{subfigure}
    \hfill
    \begin{subfigure}[t]{0.45\linewidth}
        \includegraphics[width=\linewidth, trim=7cm 0 0 0, clip]{figs/proof/2d_setup.pdf}
        \caption{\footnotesize
            {\bf Case II:} The boundary intersects the top and right edge of the grid.
        }
        \label{fig:2d_setup_case2}
    \end{subfigure}
    \caption{
        Two cases of a linear boundary $c(t)$ crossing a 2D grid cell, where $V_{\text{lt}}, V_{\text{rt}}, V_{\text{lb}}, V_{\text{rb}}$ are the grid values of the cell.
        Without loss of generality, we assume the linear boundary always intersects the top edge of the grid cell, \ie, $0 < c(0) < 1$.
    }
    \label{fig:2d_setup}
\end{figure}

We illustrate two situations that a linear boundary crossing a 2D grid cell in \cref{fig:2d_setup}, where $V_{\text{tl}}, V_{\text{tr}}, V_{\text{bl}}, V_{\text{br}}$ are the top-left, top-right, bottom-left, and bottom-right values of a 2D grid cell and $c(t)$ is the linear boundary parameterized by the vertical position $t$.
Let the coordinates of the top-left, top-right, bottom-left, and bottom-right corners be $(0,0)$, $(1,0)$, $(0,1)$, and $(1,1)$, so the top and bottom edges of the grid cell are on the horizontal lines $t=0$ and $t=1$, respectively.
Without loss of generality, we assume the linear boundary always intersects the top edge of the 2D grid cell (\ie, $0 < c(0) < 1$) and the left-hand-side $[0, c(0))$ of the boundary is free space.
We can generalize to other cases via rotation and flipping, or by negating the grid values.
We consider the 2D target function $T$ with respect to the decision boundary inside the grid cell as a 2D unit step function, and parameterize it by the vertical coordinate $t$ so that on each horizontal scan line the target function can be expressed as
\begin{equation} \label{eq:2d_target}
\begin{split}
    T(\bm{x}(t);~c(t))
    &= \mathbbm{1}(\bm{x}(t)-c(t))\\
    &= \left\{\begin{matrix}
        1\,, & \bm{x}(t)>c(t) \,, \\ 
        0\,, & \bm{x}(t)\leq c(t) \,.
        \end{matrix}\right.
\end{split} 
\end{equation}
The goal here is to show that, by choosing suitable values for $V_{\text{tl}}, V_{\text{tr}}, V_{\text{bl}}, V_{\text{br}}$, we are able to approximate the target function closely enough within the required tolerance using our post-activation scheme.

\paragraph{Case I: $0 < c(0) < 1$ and $0 < c(1) < 1$.}
An example is illustrated in \cref{fig:2d_setup_case1}, where the linear boundary intersects the top edge and the bottom edge of a grid cell.
The linear boundary $c$ is a linear function of $t$ defined by $c(t) = (1-t) \cdot c(0) + t \cdot c(1)$. Based on the results we have derived for 1D,
to prove this 2D case we only need to show that \cref{eq:1d_b_linear_to_a} and \cref{eq:1d_a_upper_bound} are linear function of $c$, so that once we use the two equations to determine $V_{\text{tl}}, V_{\text{tr}}$ for approximating $c(0)$ and $V_{\text{bl}}, V_{\text{br}}$ for approximating $c(1)$, we can readily recover $c(t)$ from $c(0)$ and $c(1)$ on all horizontal scan lines $0 \leq t \leq 1$. That is, the approximation criteria in \cref{eq:1d_proof_c1} and \cref{eq:1d_proof_c2}, where the target surface position is $c=c(t)$, are automatically satisfied by
\begin{align*}
   a = a(t) &= V_{\text{tl}}(1-t) + V_{\text{bl}}t ~,\\
   b = b(t) &= V_{\text{tr}}(1-t) + V_{\text{br}}t ~.
\end{align*}

\cref{eq:1d_a_upper_bound} is trivially a linear function of $c$ given $\delta$.
By substituting $a$ in \cref{eq:1d_b_linear_to_a} with $a^{\text{(upper bound)}}$ from \cref{eq:1d_a_upper_bound}, we get
\[
    b = \log(2^{\frac{1}{\delta}} - 1) \frac{c+\Delta-1}{\Delta} - \log\left(\frac{1}{\epsilon^{1/\delta}}-1\right)\frac{c-1}{\Delta}
\]
when $0 < \delta < 1$, and
\[
    b = \log\left( \frac{1}{(1 - \epsilon)^{1/\delta}} - 1 \right)\frac{c-1}{\Delta} - \log(2^{\frac{1}{\delta}} - 1)\frac{c-\Delta-1}{\Delta}
\]
when $1 \leq \delta$.
Thus, $b$ is also a linear function of $c$ given $\delta$.

\begin{figure}[h]
    \centering
    \begin{subtable}{\linewidth}
        \centering
        \begin{tabular}{c|c|c}
        \hline
        $T(\bm{x};~c)$ & By \cref{eq:1d_a_upper_bound} & By \cref{eq:1d_b_linear_to_a} \\
        \hline\hline
        $c=-0.6$ & $a \approx 1040.4$ & $b \approx 2772.6$ \\
        $c=1.3$ & $a \approx -2250.8$ & $b \approx -518.6$ \\
        $c=1.5$ & $a \approx -2597.2$ & $b \approx -865.0$ \\
        \hline
        \end{tabular}
    \end{subtable}
    \par\medskip
    \begin{subfigure}[t]{0.49\linewidth}
        \centering
        \includegraphics[width=\linewidth]{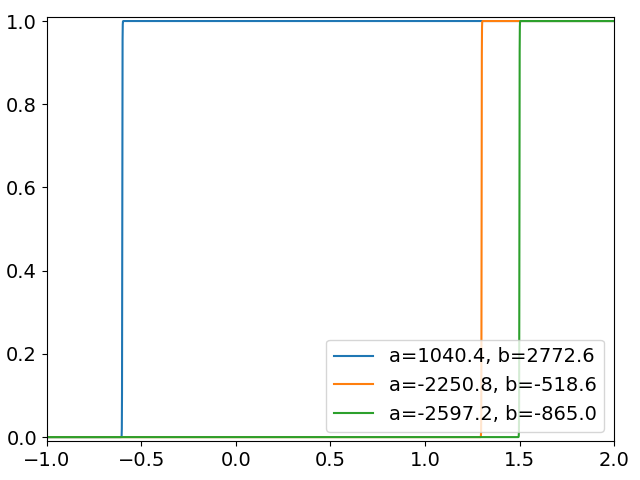}
        \caption{The derived $S(\bm{x};~a,b)$.}
    \end{subfigure}
    \hfill
    \begin{subfigure}[t]{0.49\linewidth}
        \centering
        \includegraphics[width=\linewidth]{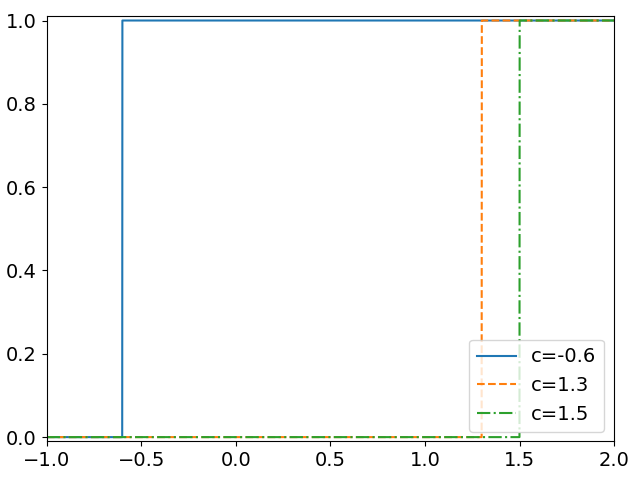}
        \caption{Some specific $T(\bm{x};~c)$.}
    \end{subfigure}
    \caption{The derivation in \cref{ssec:proof_1d} is also applicable to extrapolation ($c<0$ or $c>1$) with minor modifications.}
    \label{fig:verify_1d_plot_extrapolation}
\end{figure}

\paragraph{Case II: $0 < c(0) < 1$ and $1 < c(1)$.}
We assume $0<c<1$ in the target function in \cref{eq:1d_target}, but it finally turns out that the derivations in \cref{ssec:proof_1d} can trivially generalize to $1<c$.
Actually, the derivations also work for $c < 0$ as long as we modify the signs in inequality (\ref{eq:1d_a_upper_bound_loose}) to `$\geq$', and the upper bound in \cref{eq:1d_a_upper_bound} becomes lower bound.
We verify the results with some examples shown in \cref{fig:verify_1d_plot_extrapolation}.

In summary, the derivation in \cref{ssec:proof_1d} also works for extrapolation with minor modifications and the derivation in {\bf Case I} is directly applicable for {\bf Case II}.

\begin{figure}
    \centering
    \begin{subfigure}[t]{.25\linewidth}
        \centering
        \includegraphics[width=\linewidth]{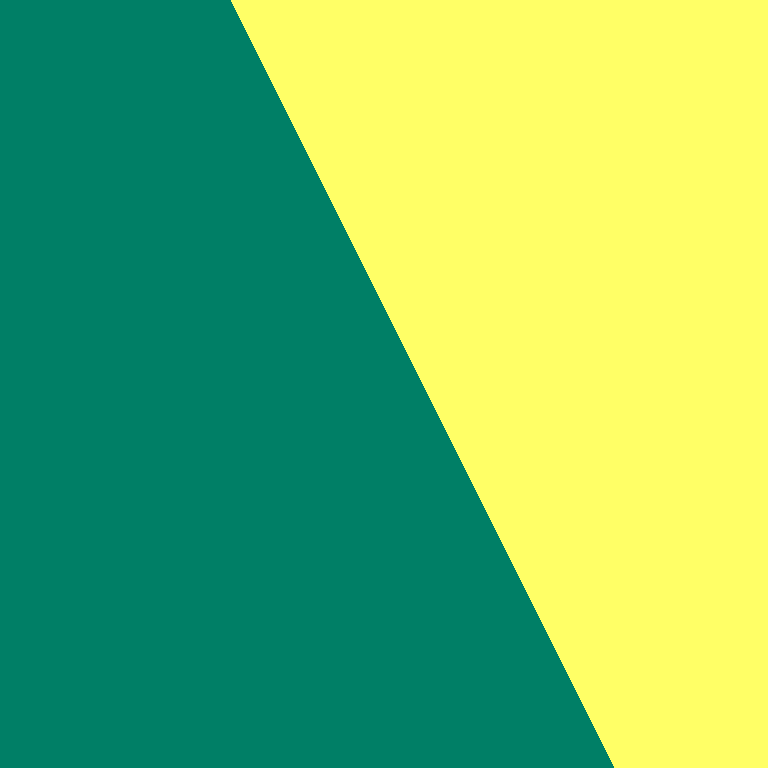}
        \caption{Target boundary $c(t) = 0.5\cdot t + 0.3$.}
    \end{subfigure}
    \hfill
    \begin{subtable}[t]{.7\linewidth}
        \centering
        \begin{tabular}{@{}c|c|c@{}}
        $\epsilon$ & $10^{-4}$ & $10^{-4}$ \\
        $\Delta$ & $0.20$ & $0.05$ \\
        \hline
        $V_{\text{tl}}$ & $-24.9$ & $-518.6$ \\
        $V_{\text{tr}}$ & $61.7$ & $1213.6$ \\
        $V_{\text{bl}}$ & $-68.2$ & $-1384.7$ \\
        $V_{\text{br}}$ & $18.4$ & $347.5$ \\
        \hline
        & \includegraphics[width=.38\linewidth]{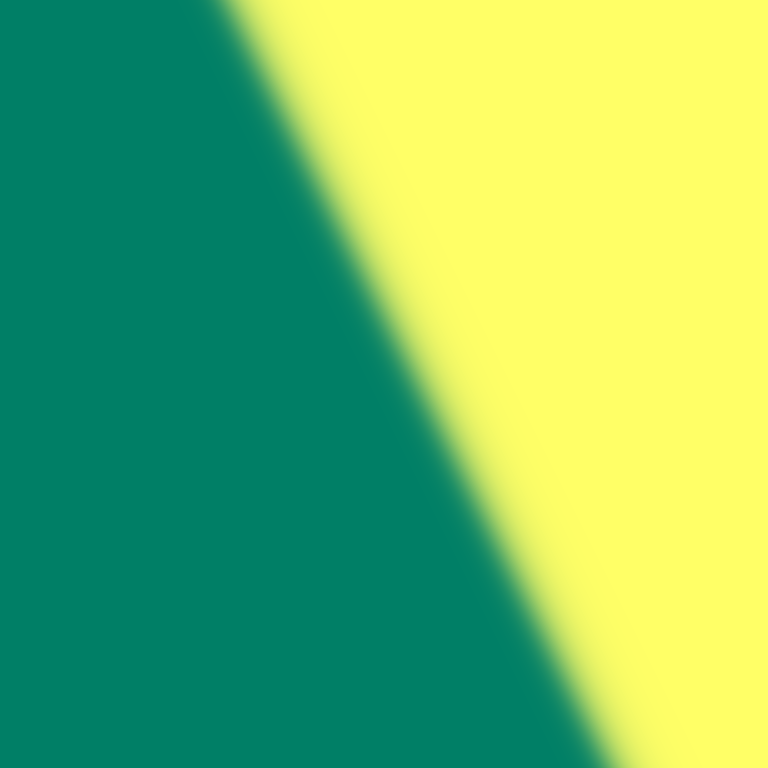} & \includegraphics[width=.38\linewidth]{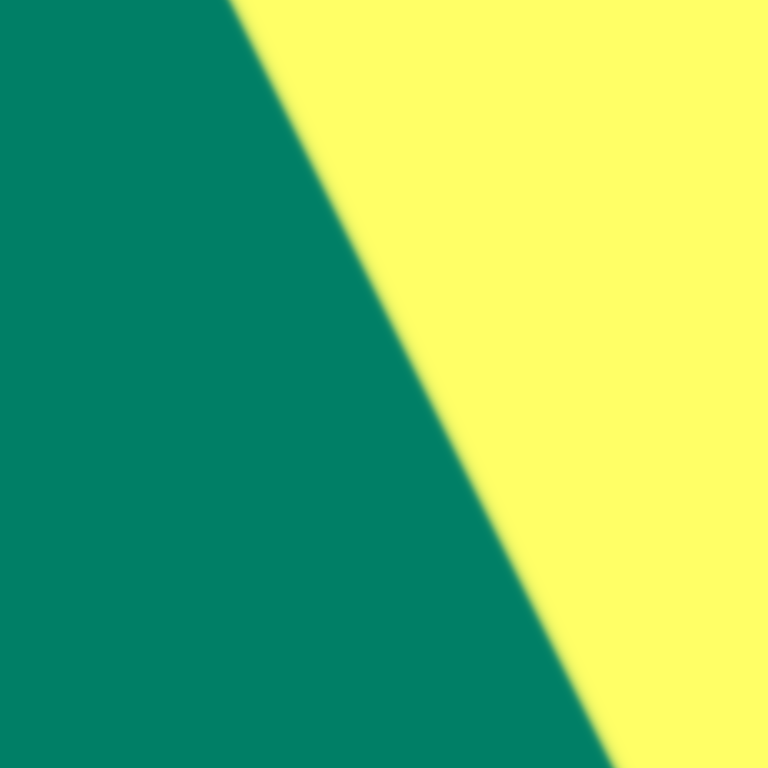} \\
        \end{tabular}
    \end{subtable}
    \par\bigskip
    \begin{subfigure}[t]{.25\linewidth}
        \centering
        \includegraphics[width=\linewidth]{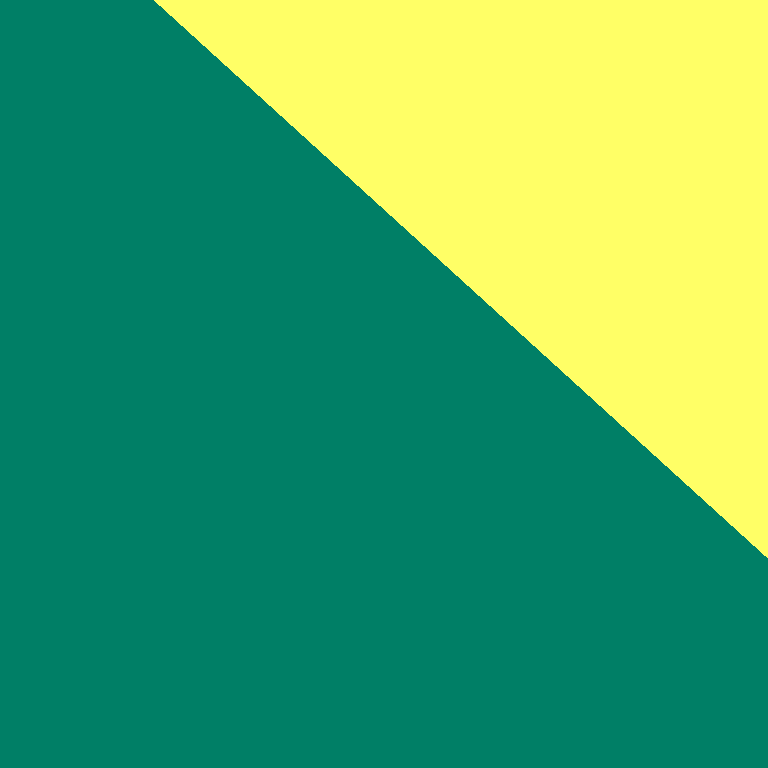}
        \caption{Target boundary $c(t) = 1.1\cdot t + 0.2$.}
    \end{subfigure}
    \hfill
    \begin{subtable}[t]{.7\linewidth}
        \centering
        \begin{tabular}{@{}c|c|c@{}}
        $\epsilon$ & $10^{-4}$ & $10^{-4}$ \\
        $\Delta$ & $0.20$ & $0.05$ \\
        \hline
        $V_{\text{tl}}$ & $-16.2$ & $-345.3$ \\
        $V_{\text{tr}}$ & $70.4$ & $1386.9$ \\
        $V_{\text{bl}}$ & $-111.5$ & $-2250.8$ \\
        $V_{\text{br}}$ & $-24.9$ & $-518.6$ \\
        \hline
        & \includegraphics[width=.38\linewidth]{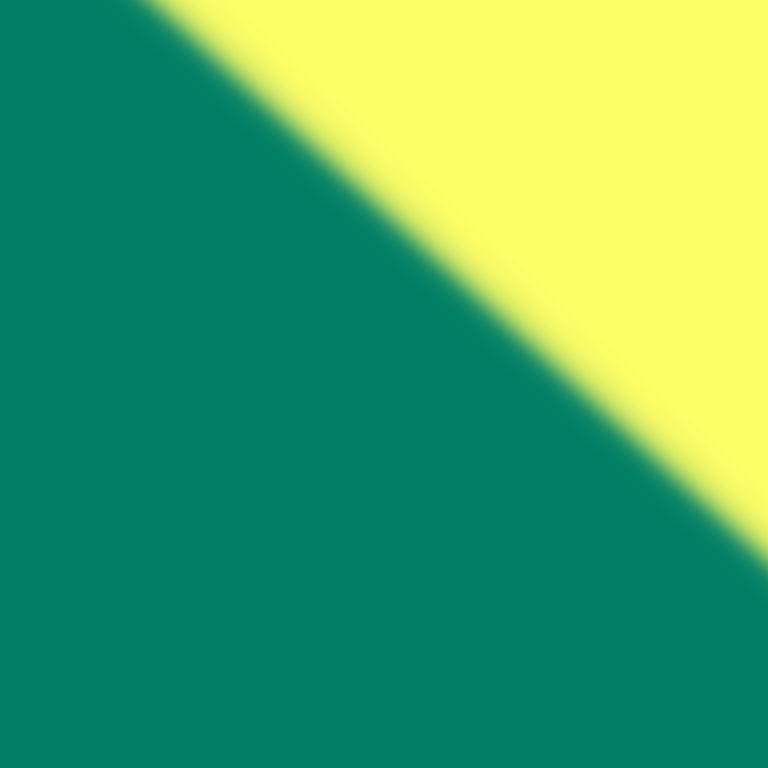} & \includegraphics[width=.38\linewidth]{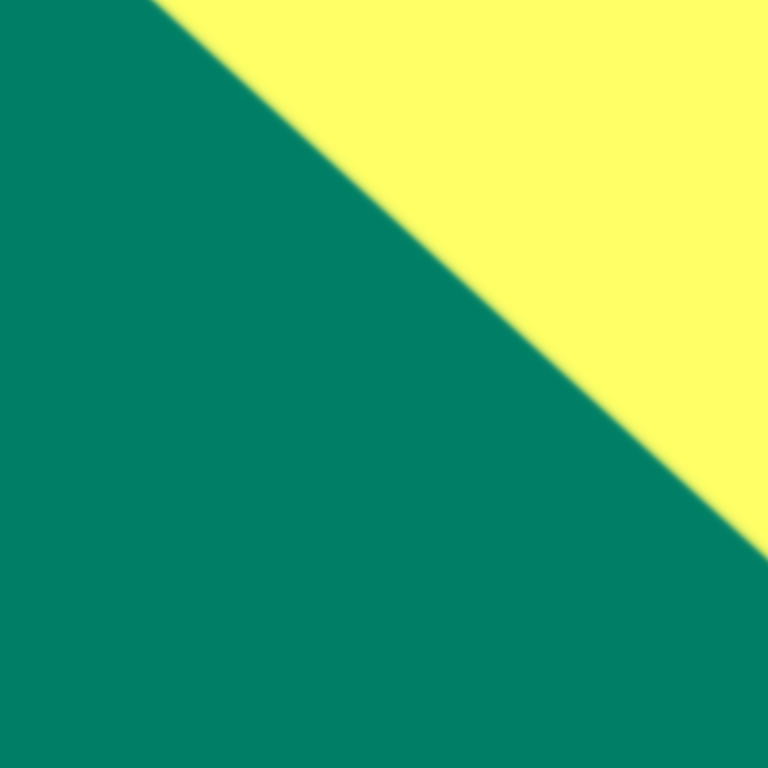} \\
        \end{tabular}
    \end{subtable}
    \caption{Using the extended 2D derivation to directly find the grid values $V_{\text{tl}}, V_{\text{tr}}, V_{\text{bl}}, V_{\text{br}}$ that fit the target boundary $c(t)$ under the required error bound $\epsilon$ and tolerance $\Delta$.}
    \label{fig:2d_examples_task}
\end{figure}

As a verification, we construct two examples in \cref{fig:2d_examples_task}.
The volume rendering step size is set to $\delta=0.5$, the error bound is set to $\epsilon=10^{-4}$, and we show two results with tolerance $\Delta=0.20$ and $\Delta=0.05$.
For each target boundary, we first evaluate $c(0)$ and $c(1)$.
The values of $V_{\text{tl}}, V_{\text{bl}}$ can then be derived from \cref{eq:1d_a_upper_bound} and the values of $V_{\text{tr}}, V_{\text{br}}$ can be derived from \cref{eq:1d_b_linear_to_a}.

\subsection{Derivation for a 3D grid cell}
\begin{figure}
    \centering
    \includegraphics[width=.95\linewidth]{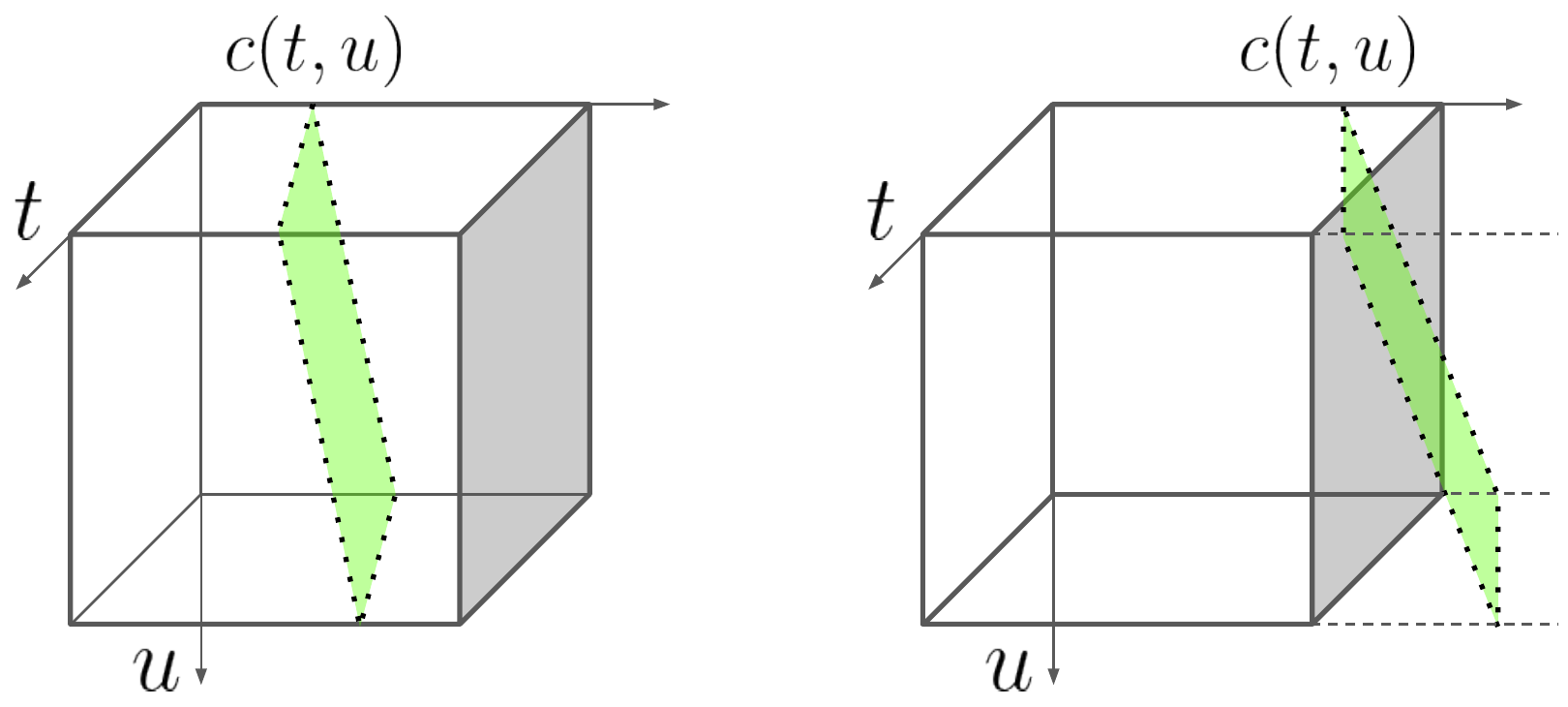}
    \caption{A linear surface $c(t,u)$ crossing a 3D grid cells. Without loss of generality, we assume the surface intersects with two horizontals planes, $u=0$ and $u=1$, which the top face and the bottom face of a 3D grid cell aligned with.}
    \label{fig:3d_setup}
\end{figure}

Similar to the proof for 2D in \cref{ssec:proof_2d}, we can assume that the 3D linear surface $c(t,u)$ intersects the top face of a 3D grid cell, as illustrated in \cref{fig:3d_setup}.
Without loss of generality, we assume the surface $c(t,u)$ intersects the horizontal planes $u=0$ and $u=1$, and the left-hand-side of the surface,
\[\left\{[t, u, v] \mid 0\leq t\leq 1, 0\leq u\leq 1, 0\leq v\leq s(t,u)\right\} ~,\]
is free space.
The linear surface is $c(t,u) = (1-u)\cdot c(t,0) + u\cdot c(t,1)$.
We can use the results in \cref{ssec:proof_2d} to determine the grid values of the top four corners with $c(t,0)$ and the values of the bottom four corners with $c(t,1)$.
The linear boundary at every horizontal slice $0 \leq u \leq 1$ can be automatically satisfied, as we have shown in \cref{ssec:proof_2d}.

\subsection{Future extensions}

\begin{figure}
    \centering
    \begin{tabular}{@{}c|c|c@{}}
    \hline
    \includegraphics[width=0.2\linewidth]{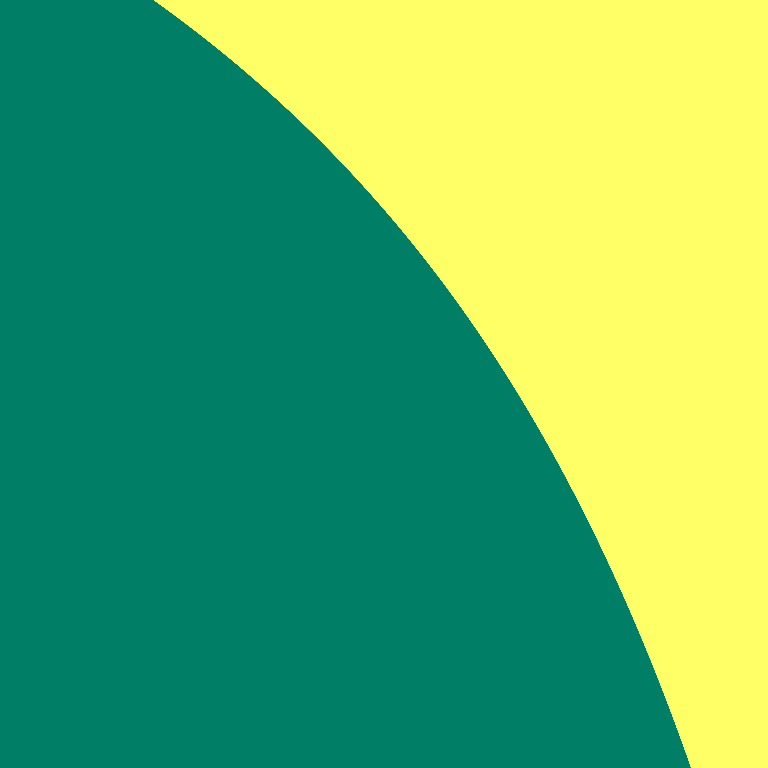} &
    \includegraphics[width=0.2\linewidth]{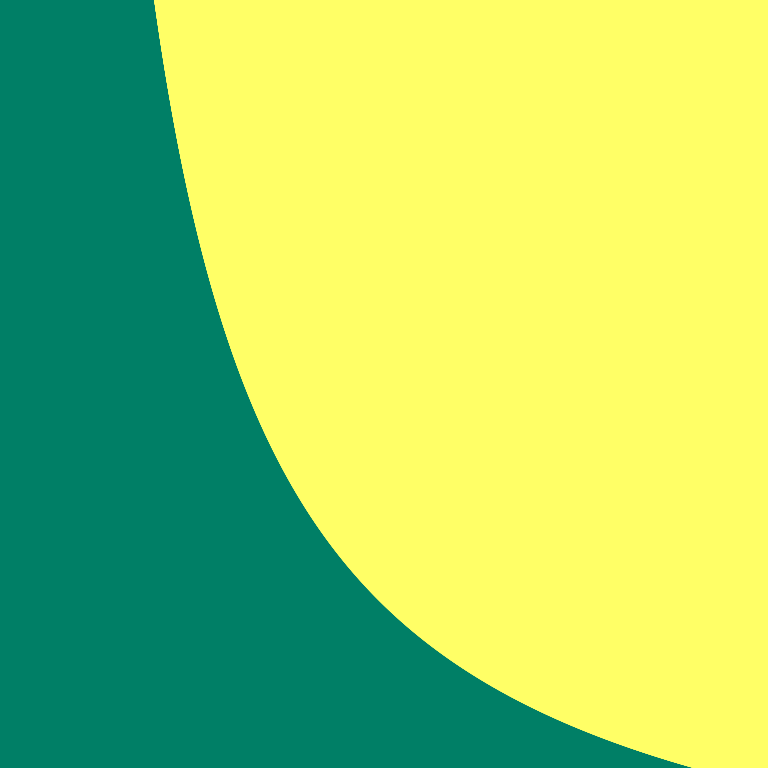} &
    \includegraphics[width=0.2\linewidth]{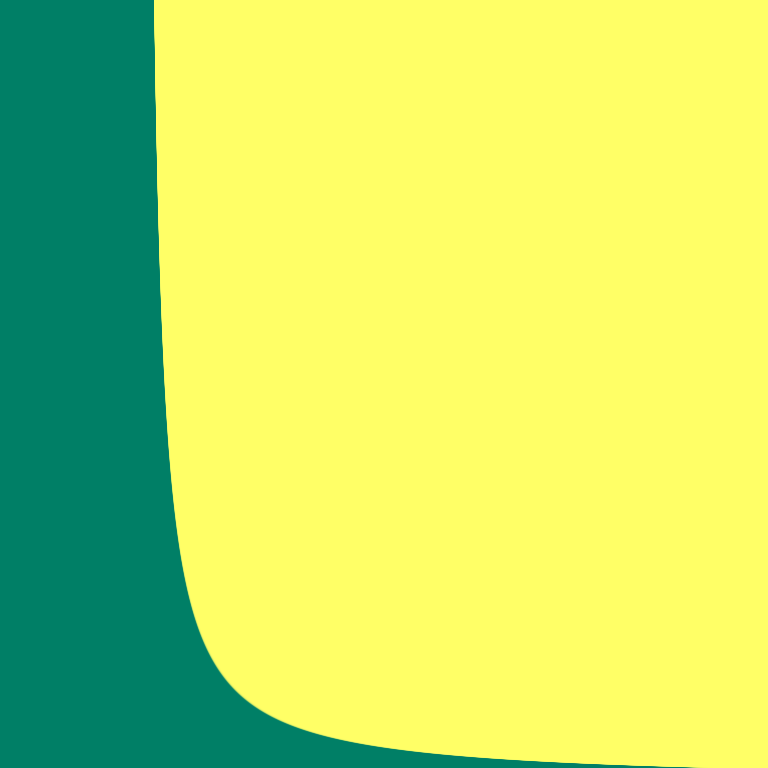} \\
    {\scriptsize $\Delta^{\text{(t)}}=1\mathrm{e}{-3}$} &
    {\scriptsize $\Delta^{\text{(t)}}=1\mathrm{e}{-3}$} &
    {\scriptsize $\Delta^{\text{(t)}}=1\mathrm{e}{-3}$} \\
    {\scriptsize $\Delta^{\text{(b)}}=5\mathrm{e}{-4}$} & 
    {\scriptsize $\Delta^{\text{(b)}}=5\mathrm{e}{-3}$} & 
    {\scriptsize $\Delta^{\text{(b)}}=5\mathrm{e}{-2}$} \\
    \hline
    \includegraphics[width=0.2\linewidth]{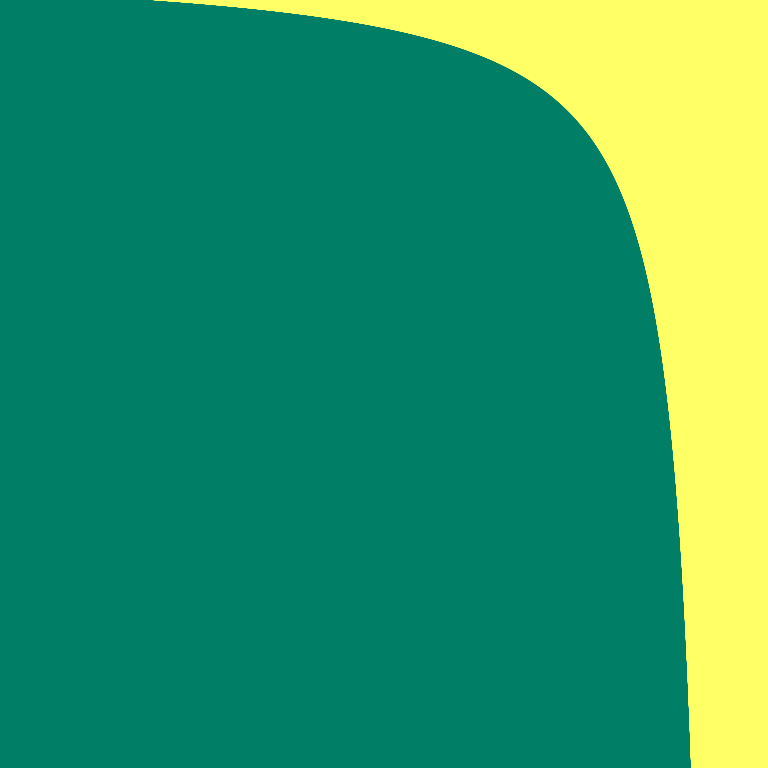} &
    \includegraphics[width=0.2\linewidth]{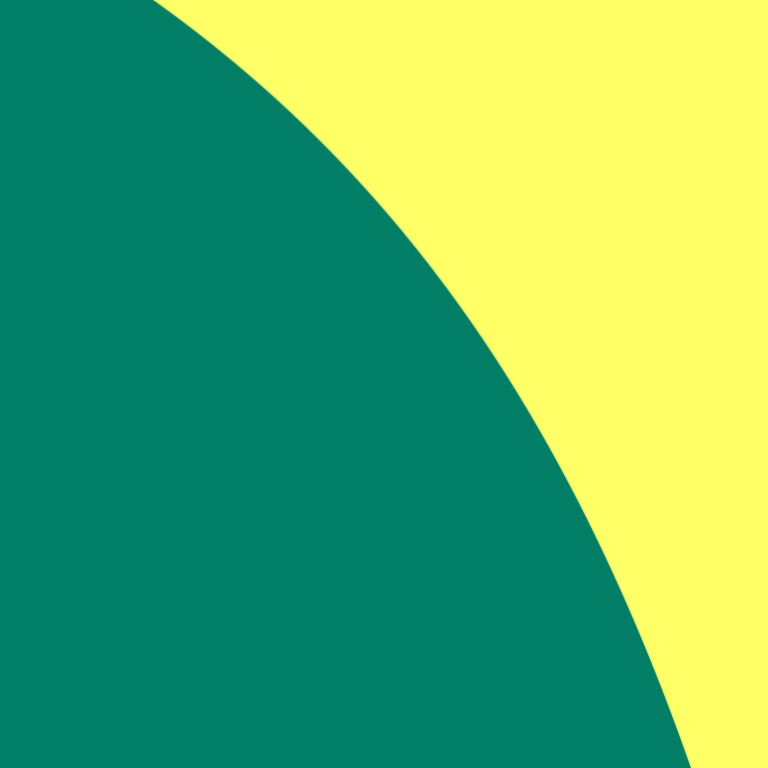} &
    \includegraphics[width=0.2\linewidth]{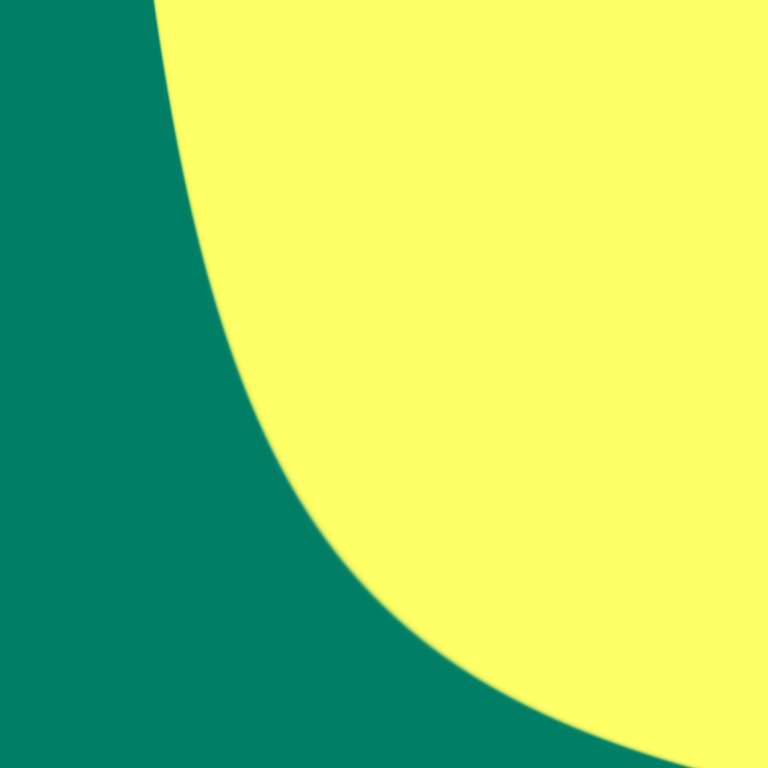} \\
    {\scriptsize $\Delta^{\text{(t)}}=1\mathrm{e}{-2}$} &
    {\scriptsize $\Delta^{\text{(t)}}=1\mathrm{e}{-2}$} &
    {\scriptsize $\Delta^{\text{(t)}}=1\mathrm{e}{-2}$} \\
    {\scriptsize $\Delta^{\text{(b)}}=5\mathrm{e}{-4}$} & 
    {\scriptsize $\Delta^{\text{(b)}}=5\mathrm{e}{-3}$} & 
    {\scriptsize $\Delta^{\text{(b)}}=5\mathrm{e}{-2}$} \\
    \hline
    \includegraphics[width=0.2\linewidth]{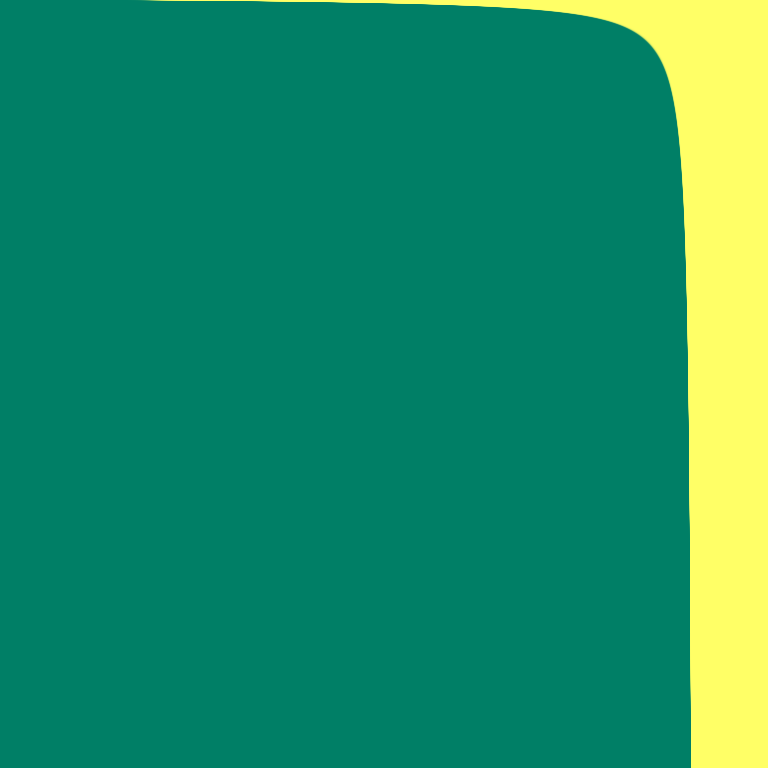} &
    \includegraphics[width=0.2\linewidth]{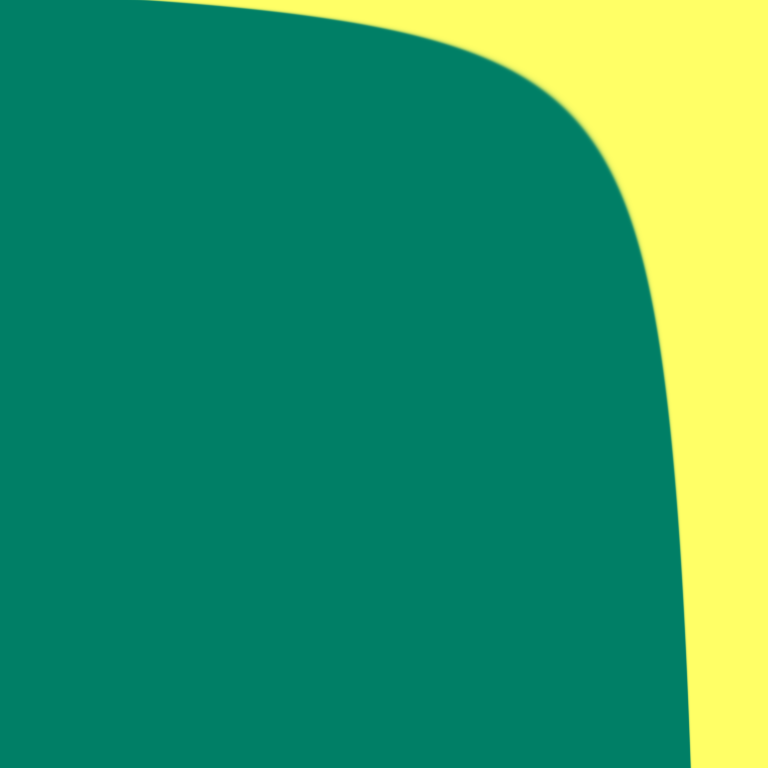} &
    \includegraphics[width=0.2\linewidth]{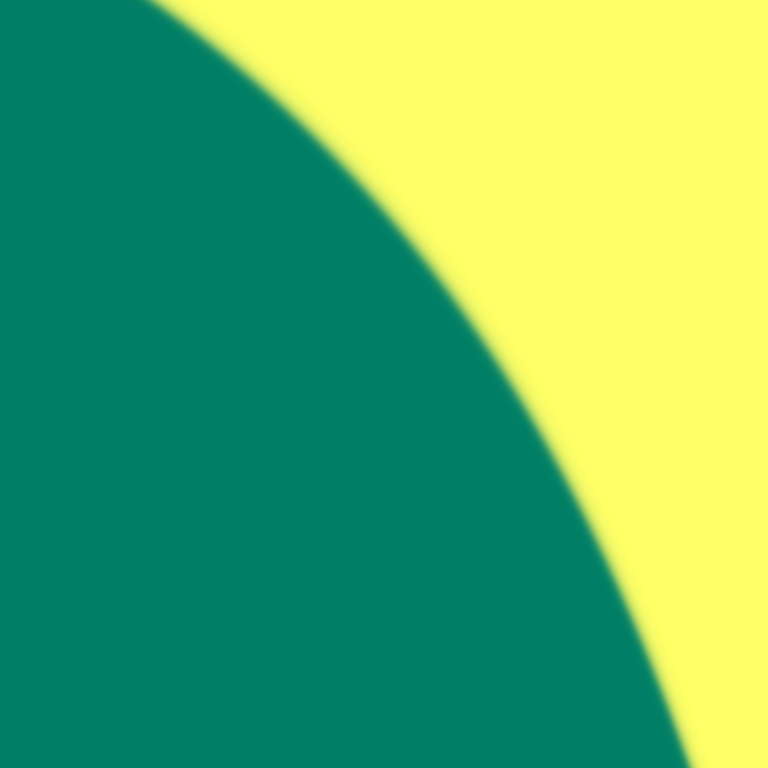} \\
    {\scriptsize $\Delta^{\text{(t)}}=1\mathrm{e}{-1}$} &
    {\scriptsize $\Delta^{\text{(t)}}=1\mathrm{e}{-1}$} &
    {\scriptsize $\Delta^{\text{(t)}}=1\mathrm{e}{-1}$} \\
    {\scriptsize $\Delta^{\text{(b)}}=5\mathrm{e}{-4}$} & 
    {\scriptsize $\Delta^{\text{(b)}}=5\mathrm{e}{-3}$} & 
    {\scriptsize $\Delta^{\text{(b)}}=5\mathrm{e}{-2}$} \\
    \hline
    \end{tabular}
    \caption{
        We fix $c(0)=0.2, c(1)=0.9, \delta=0.5, \epsilon=10^{-4}$ but use different tolerances for the top edge ($\Delta^{\text{(t)}}$) and the bottom edge ($\Delta^{\text{(b)}}$).
        Following the same procedure as in \cref{ssec:proof_2d} to determine the grid values, we can obtain a sharp non-linear surface.
    }
    \label{fig:beyond_linear}
\end{figure}

\paragraph{Beyond linear surface (decision boundary).}
We only prove that the alpha field by post-activated interpolation can be arbitrarily close to a linear surface.
We show in \cref{fig:beyond_linear} that we can further tune the tolerances at the top edge and bottom edge of a grid to produce sharp and non-linear surfaces.
Extending the proof or the capability of post-activation in future work could be helpful to geometry modeling.

\paragraph{Closed form solution when 3D available.}
In this work, we only consider the same input setup as NeRF, where only 2D observations and camera poses are available.
In cases that the 3D model of the scene is available, an algorithm to convert the 3D model to our post-activated density voxel grid can be helpful.
Our representation is directly compatible with gradient-based optimization and volume rendering to support follow-up applications, while 3D in other formats like mesh or point cloud may need more effort.
We believe our derivations are useful for future work to develop a closed-form solution to convert the 3D in other formats into our representation.

{\small
\bibliographystyle{ieee_fullname}
\bibliography{neural_rendering}
}

\end{document}